\documentclass[journal]{IEEEtran} 

\IEEEoverridecommandlockouts                              

\usepackage{amsmath, amssymb, amsfonts, mathtools}
\usepackage{times} 
\usepackage{acronym}
\usepackage{units}
\usepackage{tabularx}
\usepackage{booktabs}
\usepackage{cite}
\usepackage{url}

\usepackage[table]{xcolor}
\usepackage[short]{optidef}
\usepackage{tcolorbox}
\usepackage{tikz}
\usepackage{algorithm}
\usepackage{colortbl}   
\usepackage{algorithmic}
\usetikzlibrary{shapes.geometric,shapes.misc}
\usepackage[
    colorlinks=true,
    linkcolor=black,
    citecolor=black,
    urlcolor=blue,
    implicit=false
]{hyperref}
\newcommand{\grayrow}{\rowcolor[gray]{0.9}}

\newtheorem{assumption}{Assumption}

\newtheorem{proof}{Proof}

\definecolor{myred}{RGB}{214,39,40}
\definecolor{mygreen}{RGB}{44,160,44}

\definecolor{taborange}{rgb}{1.0, 0.8, 0.6}  
\definecolor{tabpurple}{rgb}{0.82, 0.75, 0.95} 
\definecolor{tabgrey}{gray}{0.9}                 
\definecolor{darkred}{rgb}{0.5,0.0,0.0}
\definecolor{darkgreen}{rgb}{0.0,0.4,0.0}

\newcommand{\reb}[2][\relax]{%
{#2}%
}
\newcommand{\rev}[1]{#1}

\def\suppresschangemarks{} 
\newif\ifmarkchanges
\markchangestrue
\ifdefined\suppresschangemarks\markchangesfalse\fi
\newcommand{\restored}[1]{{\ifmarkchanges\color{red}\hypersetup{citecolor=red}\fi#1}}
\makeatletter
\AtBeginDocument{%
\ifmarkchanges
\let\orig@bibitem\bibitem
\renewcommand{\bibitem}[1]{%
	\normalcolor
	\ifnum\pdfstrcmp{#1}{delpia2017Mixed}=0\color{red}\fi
	\ifnum\pdfstrcmp{#1}{koeppe2012Complexity}=0\color{red}\fi
	\ifnum\pdfstrcmp{#1}{burer2012nonconvex}=0\color{red}\fi
	\orig@bibitem{#1}}%
\fi}
\makeatother

\newtcolorbox{algbox}[2][]{%
	colframe=gray!15, 
	colback=gray!05, 
	coltitle=black, 
	fonttitle=\bfseries,
	colupper=black, 
	title=#2, 
	arc=1mm,
	boxrule=0.0mm, 
	left=0pt,
	right=0pt,
	top=-5pt,
	bottom=0pt,
	#1 
}

\definecolor{somegray}{rgb}{0.5, 0.5, 0.5}
\newcommand{\darkgrayed}[1]{\textcolor{somegray}{#1}}
\makeatletter
\newcommand*\titleheader[1]{\gdef\@titleheader{#1}}
\AtBeginDocument{%
	\let\st@red@title\@title
	\def\@title{%
		\vskip-0.75em
		\bgroup\normalfont\large\centering\@titleheader\par\egroup
		\vskip0.15em\st@red@title}
}
\makeatother
\titleheader{\darkgrayed{\footnotesize This paper has been accepted for publication in IEEE Transactions on Robotics.}}

\title{\LARGE \bf
\reb{Temporal Cascading of Planning and Control for Quadrotor MPC}
}

 \author{Rudolf Reiter$^{1}$, Chao Qin$^{2}$, Leonard Bauersfeld$^{1}$ and Davide Scaramuzza$^{1}$
     \thanks{\rev{$^{1}$Authors are with the Robotics and Perception Group, Department of Informatics, University of Zurich, Andreasstrasse 15, 8050 Zurich, Switzerland (\protect\url{https://rpg.ifi.uzh.ch}). Rudolf Reiter is the corresponding author (e-mail: rreiter@ifi.uzh.ch).}
     This work was supported by the European Union’s Horizon Europe Research and Innovation Programme under grant agreement No. 101120732 (AUTOASSESS) and the European Research Council (ERC) under grant agreement No. 864042 (AGILEFLIGHT).}
     \thanks{\rev{$^{2}$Institute for Aerospace Studies, University of Toronto, 4925 Dufferin Street, Toronto, ON M3H 5T6, Canada.}}%
 }

\newcommand{\ub}[1]{\overline{#1}}
\newcommand{\lb}[1]{\underline{#1}}
\newcommand{\hf}[1]{#1}
\newcommand{\lf}[1]{{#1}^{\mathrm{lf}}}

\newcommand{\uHf}{u^\mathrm{hf}}

\newcommand{\xPos}{p}
\newcommand{\xQuat}{q}
\newcommand{\xVel}{v}
\newcommand{\xBr}{\omega}
\newcommand{\xRot}{\Omega}
\newcommand{\xF}{f_z}

\newcommand{\uDf}{\dot{f}_{z}}
\newcommand{\xAcc}{a}
\newcommand{\uJerk}{j}
\newcommand{\bfr}[0]{\ensuremath{\mathcal{B}}} 

\newcommand{\R}{\mathbb{R}}
\newcommand{\Z}{\mathbb{N}}

\newcommand{\T}{\top}

\DeclareMathSymbol{\shortminus}{\mathbin}{AMSa}{"39}

\newacro{QP}{quadratic program}
\newacro{LP}{linear program}
\newacro{CP}{convex program}
\newacro{NLP}{nonlinear program}
\newacro{IOC}{inverse optimal control}
\newacro{NMPC}{nonlinear model predictive control}
\newacro{SQP}{sequential quadratic programming}
\newacro{OCP}{optimal control problem}
\newacro{BB}{branch-and-bound}
\newacro{MPC}{model predictive control}
\newacro{FONC}{first order necessary condition}
\newacro{MDP}{Markov decision process}
\newacro{IP}{interior point}
\acrodefplural{MDP}[MDPs]{Markov decision processes}
\newacro{IVP}{initial value problem}
\newacro{RTI}{real time iteration}
\newacro{KKT}{Karush-Kuhn-Tucker}
\newacro{BFGS}{Broyden–Fletcher–Goldfarb–Shanno}
\newacro{MPPI}{model predictive path integral}
\newacro{DAE}{differential algebraic equation}
\newacro{SOSC}{second-order sufficient condition}

\newacro{PPO}{proximal policy optimization}
\newacro{SAC}{soft actor critic}
\newacro{FF}{feed forward}
\newacro{RL}{reinforcement learning}
\newacro{AI}{artificial intelligence}
\newacro{NN}{neural network}
\newacro{LSTM}{Long Short-Term Memory}
\newacro{RNN}{recurrent neural network}
\newacro{DP}{dynamic programming}
\newacro{DAgger}{dataset aggregation}

\newacro{CG}{center of gravity}

\newacro{ODE}{ordinary differential equation}
\newacro{PCC}{Pearson correlation coefficient}
\newacro{MIQP}{mixed-integer quadratic program}
\newacro{MINLP}{mixed-integer nonlinear program}

\newlength{\dhatheight}

\begin{document}

\markboth{IEEE Transactions on Robotics}%
{Reiter \MakeLowercase{\textit{et al.}}: Temporal Cascading of Planning and Control for Quadrotor MPC}

\maketitle

\begin{abstract}
	\reb[4.9]{Many aerial tasks involving quadrotors demand both instant reactivity and long-horizon planning for obstacle avoidance, energy efficiency, or trajectory tracking.
	High-fidelity models enable accurate control but are too slow for long horizons. Low-fidelity planners scale but cannot control the system directly, necessitating cascaded architectures.
	Prevailing hierarchical approaches plan with a simplified model and use a high-fidelity controller for tracking, yet this decomposition is inherently suboptimal. The controller is limited by the coarse plan, and conventional MPC alternatives shorten the horizon to stay real-time feasible.
	We present \textsc{Unique}, an MPC architecture that replaces this hierarchical stacking with temporal cascading. The planning problem is formulated as the second tail horizon of a single multi-phase MPC, rather than solved separately. We align costs across horizons, derive feasibility constraints for the point-mass planning model, and introduce transition constraints that convert high-fidelity states into meaningful low-fidelity states. Parallel point-mass and mixed-integer solvers address nonconvexities while incorporating progressive 3D obstacle smoothing over the planning horizon.
	In simulation and real flights, under equal computational budgets, \textsc{Unique} improves closed-loop tracking by up to~$75\%$ compared with standard MPC and hierarchical baselines. Ablations and Pareto analyses confirm performance gains across variations in horizon, constraint approximations, and smoothing schedules.}


\end{abstract}

\begin{IEEEkeywords}
\rev{Aerial systems, model predictive control, motion planning, multi-fidelity optimization, quadrotors.}
\end{IEEEkeywords}

\section*{Supplementary Material}
Video available at 
\url{https://youtu.be/CefIqpld9N8}.
\rev{Supplementary derivations and parameters are in the Appendix.}

\section{Introduction}
Fast and safe autonomous quadrotor flight depends on two competing requirements: \emph{fast feedback} to disturbances and dynamics, and \emph{planning} to reason over long horizons for obstacle avoidance, efficiency, and mission objectives. Optimizing a high-fidelity quadrotor model over such horizons is typically intractable in real time, causing the long-standing separation between planning and control for quadrotors~\cite{romero_2022,romero_2022a,krinner_2024}. 

\begin{figure}
	\centering
	\includegraphics[width=\linewidth, trim=0cm 0.0cm 0cm 0.0cm, clip]{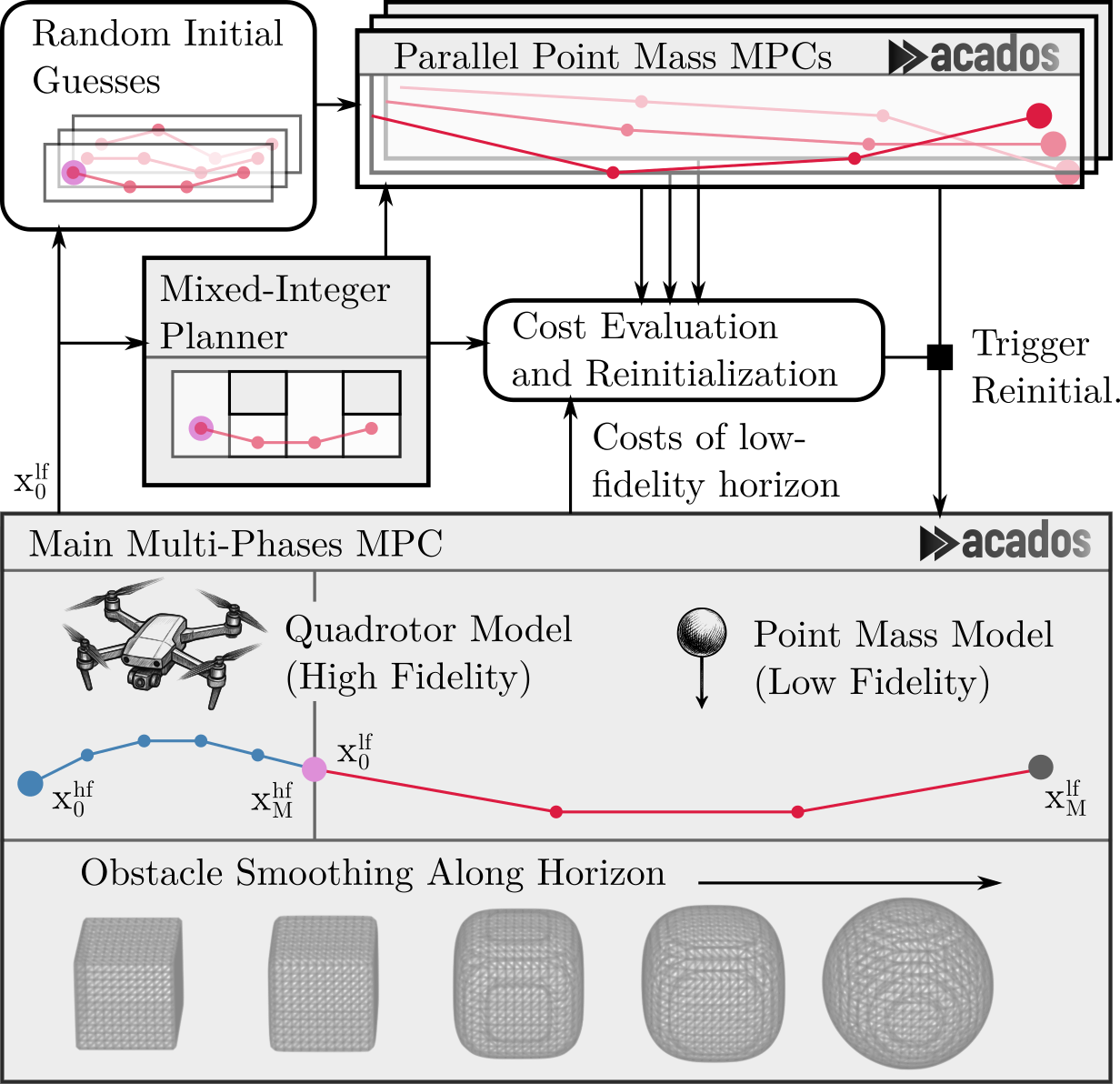}
	\caption{\reb[4.9]{Conceptual sketch of \textsc{Unique}. We employ a multi-phase MPC with two horizons: a high-fidelity control horizon and a long horizon with a lower-fidelity point-mass model, enabling long-horizon planning. Local minima of the gradient-based optimizer are avoided by (i) smoothing obstacle shapes along the horizon, and (ii) running parallel planners solely for the computationally cheap low-fidelity model on the second horizon. Whenever the cost on the low-fidelity planning horizon of such a parallel MPC is lower, the second horizon of the multi-phase MPC is reinitialized by the corresponding optimization variables.}}
	\label{fig:sketch}
\end{figure}

Hierarchical architectures compute a long-horizon plan using a simplified model, such as a point mass or a polynomial model~\cite{hehn_2015,romero_2022}, and track it with an MPC. \reb[4.1]{This hierarchical decomposition is inherently suboptimal. The asynchronously computed low-fidelity plan cannot account for the true vehicle dynamics, leading to} (i) performance limitations imposed by the coarse plan, (ii) redundant optimization in the initial horizon, and (iii) conflicting inner or outer loops that may create infeasibilities or unstable behavior. Pure MPC alternatives retain the high-fidelity model but shorten the horizon to remain real-time feasible, relying on a terminal cost and set that are difficult to obtain for complex scenes involving moving obstacles or nonlinear constraints and are prone to local minima in nonsmooth environments. Remarkably, the planning problem remains NP-hard~\cite{lavalle2006planning}, yet heuristics and tailored algorithms often make it computationally tractable.

\reb[4.1]{Our pivotal idea is to replace the conventional hierarchical planner-controller decomposition with a \emph{temporal cascading} of planning and control inside a single MPC.} \reb[4.1]{Rather than computing a plan separately and tracking it, we embed the planning problem as the second tail horizon of a multi-phase MPC, as sketched in Fig.~\ref{fig:sketch}.} \reb[4.1]{This is motivated by the fact that an MPC, in general, aims to optimize the same underlying task objective and aims to satisfy the same constraints as the planning problem. 
We observe that (i) the geometric planning around obstacles is predominantly a long-horizon problem with negligible influence on the short control horizon, and (ii) a point-mass planner can be solved very fast within each closed-loop iteration.} \reb[4.1]{Therefore, we formulate the planning problem as finding the optimal tail trajectory on the second horizon, given that we can effectively convert high-fidelity states into meaningful low-fidelity states and map constraints from the high- to the low-fidelity model, which we show to be possible for quadrotors.} The two phases are coupled by transition constraints that align (i) position and velocity, (ii) thrust-induced acceleration, and (iii) jerk/body-rate relations. Cost terms are aligned across horizons, allowing the controller to optimize a consistent objective. To improve numerical robustness over long horizons with sharp geometry, we adapt progressive smoothing~\cite{reiter_2024a} of norm-based obstacle models for 3D shapes, morphing cube-like sets to ellipsoids with increasing prediction time, \rev{which improves \ac{RTI} convergence}~\cite{diehl_2005}.

\reb[4.1]{The two-phase formulation alone converges well around local minima, but planning problems often involve severe nonconvexities. Therefore, we propose a parallel computation strategy in which we compare the tail trajectory of the main MPC with alternative solutions that may achieve a lower predicted cost.} \reb[2.1,4.1]{Randomly initialized point-mass solvers run in parallel and, as we show, finish faster than the two-phase MPC, enabling synchronous cost comparison after each \ac{RTI} step.} \reb[2.3,4.8]{For harder combinatorial problems, we additionally deploy a potentially slower global planner, in our case a mixed-integer planner as used by many state-of-the-art planning algorithms~\cite{marcucci2024fast}, which first initializes a guaranteed fast parallel point-mass solver. Only the latter's converged cost is compared to the main MPC, ensuring synchronous evaluation.} Once the cost of one parallel MPC is lower than the cost along the second horizon of the multi-phase MPC, the solver variables are initialized with the lower-cost solution of the corresponding parallel MPC.
In summary, our contributions are as follows:
\begin{itemize}
	\item \reb[4.1]{Temporal planning paradigm. We propose to formulate the planning problem not hierarchically but as the tail horizon of an MPC, yielding a sequential controller-planner structure within a single optimization.}
	\item \reb[4.9]{Two-phase MPC architecture for quadrotors.} A single optimization that couples a high-fidelity quadrotor model with a long-horizon point-mass model via equality constraints on position/velocity, thrust-acceleration, and jerk-body-rate mappings, with cost alignment across horizons. \reb[4.9]{We show that high-fidelity states can be effectively converted into meaningful low-fidelity states for quadrotors.}
	\item Feasibility-preserving low-fidelity constraints. Derivation and tractable inner approximations of thrust and body-rate safe sets for the point-mass model that ensure compatibility with high-fidelity actuator and rate limits.
	\item \reb[2.1,4.9]{Parallel tail-horizon re-planning.} \reb[2.1,4.9]{A parallel computation strategy that compares the tail trajectory of the main MPC against alternative solutions from randomly initialized point-mass solvers.}
	\item \reb[4.9]{Pareto-superior control performance. Under equal computational budgets, the two-phase formulation achieves better closed-loop performance than extending the high-fidelity MPC horizon.}
	\item Comprehensive evaluation. Simulations and real flights show substantial closed-loop gains over standard MPC and hierarchical baselines under equal compute budgets, including Pareto analyses across horizon variations and feasibility approximations, as well as demonstrations of long-horizon planning with iteration times below $\unit[5]{\mathrm{ms}}$.
\end{itemize}

\reb[4.6]{The \textsc{Unique} framework lowers the planning-control performance gap by treating planning as a tail-horizon problem of MPC rather than a separate hierarchical stage.} In obstacle-rich tasks, we observe up to $75\%$ lower closed-loop cost compared to standard MPC and hierarchical designs, while reducing online computation compared to long-horizon, high-fidelity MPC.

\subsection{Outline}
Following \rev{the related work} in Sec.~\ref{sec:related}, Sec.~\ref{sec:nominal_controller} formalizes the nominal high-fidelity MPC. Section~\ref{sec:low_level_model} introduces the point-mass planning model and derives feasibility sets. Section~\ref{sec:unique} presents the unified formulation, including cost/constraint alignment, transition constraints, progressive smoothing, the parallel MPC framework, and \rev{the} numerical approximations for real-time efficiency. Section~\ref{sec:experimental_setup} describes the experimental setup, Section~\ref{sec:evaluation} reports real-world experiments, and Section~\ref{sec:simulated} presents simulated ablations. We conclude with limitations and future directions in Section~\ref{sec:conclusion}.

\subsection{Notation}
For the quaternion rotation of a vector~$x\in\R^3$ and a quaternion~$q\in\R^4$, we use~$q\odot x$, and for the quaternion multiplication with a second quaternion~$p\in\R^4$, we use $q \cdot p$. A power~$p$ of a vector $x\in\R^n$ is taken element wise, with $x^p=[x_1^p,\ldots,x_n^p]^\top$. When using different models along an MPC horizon, we refer to the trajectory parts as \emph{phases}.

\section{Related Work}
\label{sec:related}

\textbf{Classical hierarchical architectures.}
In traditional software stacks, a high-level planner solves a simplified problem, often in discrete or low-dimensional state/control spaces, and provides a reference path or trajectory to a lower-level controller~\cite{hehn_2015}. Graph search~\cite{cieslewski2017rapid,naazare2019application,liu2024integrated} and sampling-based methods such as rapidly exploring random trees (RRTs)~\cite{achtelik2013path} are common choices. For quadrotors, the authors in~\cite{romero_2022} employ a point-mass model for planning, while Model Predictive Contouring Control (MPCC)~\cite{romero_2022a} and its extensions~\cite{krinner_2024} apply the decoupled approach to racing by optimizing along a precomputed trajectory. 
While these planners are computationally efficient and capture high-level objectives, the reduced dimensionality leads to suboptimal trajectories, and the hierarchical decomposition introduces coupling issues between the planner and the controller.

\textbf{Motion planning with surrogate models.}
Approximating the quadrotor as a low-fidelity integrator chain~\cite{romero_2022} and allowing small deviations from its true dynamics~\cite{meyer2023top, teissing2024real} can significantly simplify the motion planning problem. In particular, Hehn and D’Andrea~\cite{hehn_2015} proposed a third-order point-mass model for online planning under actuation constraints, while higher-order (snap) models~\cite{tal_2021} further improve trajectory smoothness and accuracy at the expense of increased computational complexity. Moreover, exploiting the differential flatness of simple quadrotor models enables long-horizon motion planning for complex maneuvers~\cite{faessler_2018, wang2025unlocking} without significantly compromising model fidelity.

\reb[4.3]{
Beyond time-optimal navigation, a broader line of work has studied planning and control for quadrotors via dynamically feasible point-mass trajectory generation, with a strong focus on the combinatorial aspects of the problem. Safe-flight-corridor methods generate polynomial trajectories inside convex free-space decompositions for real-time flight in cluttered environments~\cite{liu2017Planning}, while FASTER uses a similar \ac{MIQP} formulation to this work to navigate specifically unknown environments, including safe backup trajectories~\cite{tordesillas2020faster}. The authors in~\cite{zhepei2022geometrically} propose a very fast planner by eliminating geometric constraints through smooth maps. 
Classical exact cell decompositions~\cite{latombe1991exact} and their modern convex-set variants represent free space as unions or graphs of convex regions, including safe boxes and graphs of convex sets~\cite{marcucci2024shortest,marcucci2024fast,marcucci2023motion} and propose~\ac{MIQP} solvers to provide globally optimal trajectories. As long as obstacles and constraints can be represented as cuboids and quadratic costs are considered, the planners can be used as a backend for our framework. 
Remarkably, these works demonstrate the strength of planning with simplified or geometry-aware models, but they are still primarily planner-focused and are typically deployed upstream of a separate tracking controller, rather than as a single optimization that jointly allocates high-fidelity near-horizon control and low-fidelity long-horizon planning.}
One key limitation arises from their inherently smooth, lower-fidelity plans, which cannot account for more complex aerodynamic effects, necessitating conservatism. In addition, they often require the current jerk state as an initial condition, which is difficult to obtain in practice since it depends on the rate of change of thrust~\cite{tal_2021}.

Nevertheless, these simplified models form the foundation of the long-horizon component of our unified MPC framework, where we address the limitations mentioned above by introducing a multi-phase structure that precedes the point-mass phase, enforcing feasibility-preserving constraints, and ensuring consistency with high-fidelity quadrotor dynamics.
\reb[4.3]{
Instead of proposing a competing planning algorithm to the above approaches, we propose a fundamentally different planning paradigm: not hierarchically, but along the second horizon of \ac{MPC}.
}

\textbf{Long-horizon MPC extensions.}
Several approaches aim to embed long-horizon reasoning directly into MPC. One line of work uses terminal costs learned from data. For instance, reinforcement learning is used to approximate value functions~\cite{reiter_ac4mpc_2025,seel_2024} or supervised learning for convex quadratic surrogates~\cite{abdufattokhov_2024}. If a policy is available, long-term behavior can be approximated by closed-loop costing~\cite{nicolao_1998,diehl_2024}, including recent neural-policy rollouts~\cite{ghezzi_2025}. However, these methods rely on pretrained policies or value functions, which may be environment-specific. Other research has studied recursive feasibility under multiple phases~\cite{behrunani_2024} and developed multi-phase MPC formulations within modern solvers such as \texttt{acados}~\cite{frey_2025}. Li et al.~\cite{li_2021} introduced a dual-model MPC for legged and aerial robots, although it was only tested in offline simulations and \rev{did not provide a recursive-feasibility guarantee}.
\reb[3.1]{The same authors extended the multi-model formulation for quadrupeds~\cite{li_2025}, addressing the switching dynamics of the legged robot's gait pattern and providing a tailored iLQR numerical solution scheme for the \ac{OCP}. Instead, we adapt the state-of-the-art \ac{SQP} algorithm shown in~\cite{frey_2025}, which provides the necessary efficiency for the state constraints inherent in the targeted motion planning subproblem.}

\reb[4.5]{
\textbf{Nonsmooth Obstacles in MPC.}
Exact collision avoidance can be imposed through smooth dual reformulations without conservative geometric approximations~\cite{zhang2021optimization}, while corridor-based methods use discrete polyhedral corridors~\cite{liu2017Planning,tordesillas2020faster}, smooth differentiable parametric corridors based on ellipses~\cite{arrizabalaga2024differentiable}, or differentiable collision models for convex primitives~\cite{tracy2023differentiable}. In contrast, the proposed progressive smoothing, derived from~\cite{reiter_2024a}, is not intended to replace these formulations but to provide a lightweight strategy for norm-based obstacle constraints within a single receding-horizon NLP, improving numerical convergence in cluttered 3D environments while preserving the single-optimization-problem structure. The proposed framework can be adapted to fit specific application domains where other obstacle approximations, such as~\cite{arrizabalaga2024differentiable}, may be preferred.}

\section{Nominal Controller}
\label{sec:nominal_controller}
In the following, we describe the nominal nonlinear MPC that serves as a basis for \textsc{Unique}. First, the high-fidelity model is introduced, followed by the essential system constraints, obstacle constraints, and the MPC formulation.

\subsection{High-Fidelity Control Model}
The high-fidelity quadrotor is modeled following~\cite{kaufmann_2023} as a 6-degree-of-freedom rigid body with mass~$m$ and a diagonal moment of inertia matrix
\(
{J} = \mathrm{diag}(\begin{bmatrix}J_x & J_y & J_z\end{bmatrix}).
\)

The model includes the positions~$\xPos\in\R^3$, velocities in the world coordinate frame~$\xVel\in\R^3$, orientation formulated as quaternions~$\xQuat\in S^3 \subset \R^4$, body rates~$\xBr\in\R^3$, single rotor thrusts~$f_i^\top=\begin{bmatrix}
0&0&f_{z,i}
\end{bmatrix}$ with $i\in\{1,2,3,4\}$ and the collective z-components of the single rotor thrusts~$f_{z}^\T=\begin{bmatrix}
f_{z,1} & f_{z,2} & f_{z,3} & f_{z,4}
\end{bmatrix}\in\R^4$. Thrusts are generated by the rotors with the rotor speed~$\xRot_i$ and the geometric location~$r_{\mathrm{p},i}\in\R^3$ of the $i$-th rotor in the body frame via $f_{z,i}=\xRot_i^2 c_l$ using the thrust coefficient~$c_{\mathrm{l}}\in\R$. 

The total thrust is the sum of all single rotor thrusts~$f_i$, with
\(
T(\xF)=\sum_{i=1}^4
f_{i},
\)
and controlled by the corresponding derivatives \(\uHf:=\uDf\in\R^4\).
When now computing the total force~$\rev{f:\R^4\times\R^3\rightarrow\R^3}$ acting on the drone in the body frame, we add residual force~$\rev{f_\mathrm{res}:\R^3\times\R^4\rightarrow\R^3}$ to the total thrust, with
\begin{align*}
\begin{split}
&f(\xF,v_\bfr):=\\
&T(\xF)+f_\mathrm{res}(v_\bfr,\Omega^2)=T(\xF)+f_\mathrm{res}(v_\bfr,\frac{1}{c_l}f_z),
\end{split}
\end{align*}
where the velocity~$v_\bfr\in\R^3$ in the body frame~$\bfr$ can be obtained via~$\rev{v_\bfr=q^{-1}\odot\xVel}$.

The residual force~$f_\mathrm{res}(v_\bfr,\Omega^2)$ models aerodynamic effects and is obtained by fitting selected features of~$\xRot^2$ and velocities~$\xVel_\bfr$ to measured data. The literature proposes either higher-order polynomials~\cite{kaufmann_2023} or linear features of only the velocity~$\xVel_\bfr$~\cite{faessler_2018}. We use a higher-order polynomial for a high-fidelity residual model\restored{~\cite{kaufmann_2023}}.

In addition to translation forces, we compute rotational counter-torques of the rotors via
\[\tau_{\mathrm{c},i}(f_{z,i})=\kappa_i \begin{bmatrix}
0 & 0 & f_{z,i}
\end{bmatrix}^\top.\]
In symmetric setups, the coefficients~$\kappa_i$ are equal in absolute value and have pairwise the same sign, i.e., $\kappa_1=-\kappa_2=\kappa_3=-\kappa_4$. By adding the torque originating from the single rotor thrusts to the counter torques, the collective torque is
\begin{equation*}
    \tau(f_z)=\rev{\sum_{i=1}^4 \left(\tau_{\mathrm{c},i}(f_{z,i})+r_{\mathrm{p},i}\times \begin{bmatrix}
    0 & 0 & f_{z,i}
    \end{bmatrix}^\top\right)}.
\end{equation*}

With the gravity vector~$g^\T=\begin{bmatrix}
0 & 0 & \unit[9.81]{m/s^2}
\end{bmatrix}$ and the high-fidelity state
\[
{x}=\begin{bmatrix}
\xPos^\T & \xQuat^\T & \xVel^\T & \xBr^\T&f_{z}^\T 
\end{bmatrix}^\T\in\R^{17},
\]
the quadrotor dynamics are finally written as
\begin{align}
\label{eq:3d_quad_dynamics}
\hf{\dot{x}} = 
\begin{bmatrix}
\dot{\xPos} \\  
\dot{\xQuat} \\
\dot{\xVel} \\
\dot{\xBr} \\
\dot{\xF}
\end{bmatrix} = 
\begin{bmatrix}
{\xVel}\\  
{\xQuat} \cdot \begin{bmatrix}0 & \frac{\xBr}{2}\end{bmatrix}^\top \\
\frac{1}{m} \Big({\xQuat} \odot f\big(\xQuat^{-1}\odot\xVel,\,\xF\big)\Big)  - g \\
{J}^{-1}\big( \tau(f_z)-\xBr \times {J}\xBr \big)\\
u
\end{bmatrix}
=: \hf{f}(\hf{x},\hf{u}).
\end{align}
A simplified version of~\eqref{eq:3d_quad_dynamics} uses linear residual forces, which were shown to be \rev{differentially flat}~\cite{faessler_2018}, cf. the supplementary material.

\subsection{Quadrotor System Constraints}
Quadrotors are typically constrained by maximum and minimum rotor speeds, resulting in maximum and minimum single rotor thrusts~$\ub{\xF}$ and $\lb{\xF}$. 
Most quadrotors are equipped with a low-level tracking controller that takes the body rates and collective thrust as inputs. 
Therefore, constraints are expressed as constraints~$\ub{\xBr}_{xy}\in\R$ on the body rate for roll and pitch, and the total collective thrust~$\ub{f}_\mathrm{th}$ and~$\lb{f}_\mathrm{th}$, written as
\begin{subequations}
	\label{eq:con_br_thrust}
	\begin{align}
	\label{eq:con_br}
	\hf{\mathbb{S}}_{\xBr}:=&\big\{\xBr\in\R^{3} \mid
	 |\xBr_x|\leq\ub{\xBr}_{xy}\wedge|\xBr_y|\leq\ub{\xBr}_{xy}\big\}, \\
	\label{eq:con_thrust}
	\hf{\mathbb{S}}_\mathrm{f}:=&\big\{\xF\in\R^{4} \mid 
	\lb{f}_\mathrm{th}\leq T(\xF)\leq\ub{f}_\mathrm{th}\big\}.
	\end{align}
\end{subequations}
We constrain the quadrotor to an operating region with a maximum vertical speed of~$\ub{v}_z\in\R$ and horizontal speed $\ub{v}_{xy}\in\R$, with
\[\mathbb{S}_{v}=\left\{v\in\R^3 \middle| \|v_z\|\leq\ub{v}_z\wedge\|[v_x, v_y]\|\leq\ub{v}_{xy} \right\}.\]
\begin{assumption}
\label{ass:bounded_disturbance}
We assume that the residual force~$f_\mathrm{res}(v_\bfr,\Omega^2)$ is bounded by~$\rev{\ub{f}_\mathrm{res}\in\R_{\geq0}}$ with~$\rev{\|f_\mathrm{res}(v_\bfr,\Omega^2)\|\leq \ub{f}_\mathrm{res}}$ for all $v_\bfr\in\mathbb{S}_v$ and $\rev{c_l \Omega^2\in\hf{\mathbb{S}}_f}$.
\end{assumption}

\subsection{Obstacle Avoidance Constraints}

\begin{figure*}
	\centering
	\includegraphics[width=0.95\linewidth, trim=0.1cm 5.2cm 0.5cm 4.8cm, clip]{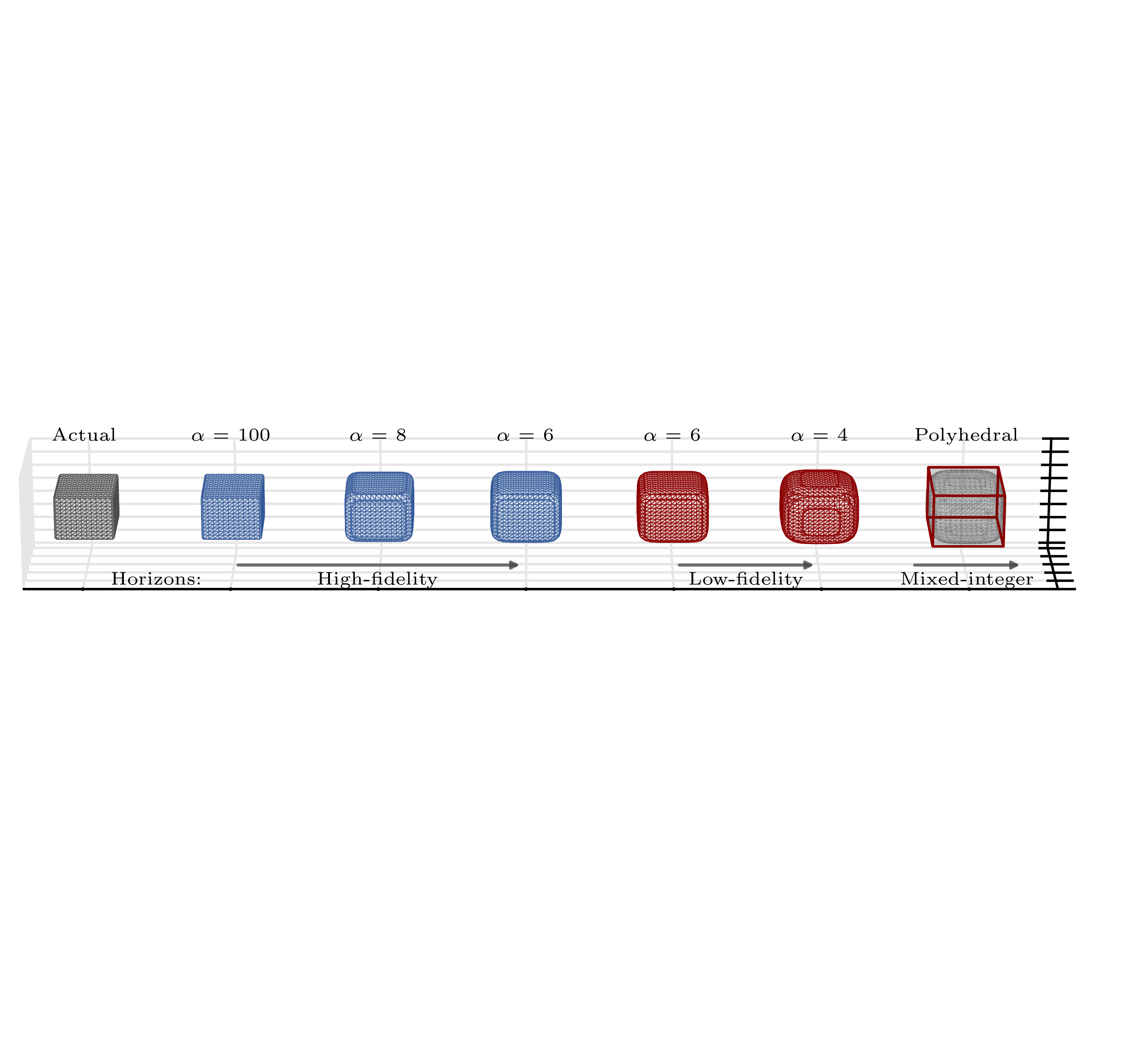}
	\caption{\reb[4.5]{Obstacle representations. Visualization of a cubic obstacle smoothed along the high-fidelity horizon (blue) for $\alpha = 100$, $8$, and $6$. Along the low-fidelity horizon, cubic obstacles can be further smoothed, as shown here for $\alpha = 6$ to $4$. The mixed-integer planner requires a polyhedral over-approximation due to the convex decomposition of the free planning space.}}
	\label{fig:obstacles_shapes}
\end{figure*}

We consider convex obstacle shapes that can be described by \rev{$p$-norms}, including ellipsoids and cubes. An obstacle is described by the tuple~$\theta=(p_o,R_o,d_o,\alpha_0)$ with the obstacle center~$p_o\in\R^3$, the orientation normal vectors~$u_o,v_o,w_o\in\R^3$, with the rotation matrix~$R_o=\begin{bmatrix}
u_o&v_o&w_o
\end{bmatrix}$, the obstacle \rev{scale parameters}~$d_o^\top=\begin{bmatrix}
d_x&d_y&d_z
\end{bmatrix}$, and the norm or shape parameter~$\alpha_o\geq2$. 
The transformation 
\begin{equation*}
\eta(p) = \begin{bmatrix}
\eta_1(p)  & \eta_2(p)  & \eta_3(p) 
\end{bmatrix}^\top: = \rev{\mathrm{diag}(d_o)^{-1}R_o^\top(p-p_o)}
\end{equation*}
is used to obtain normalized obstacle coordinates~$\eta(p)$ and describe
the occupied space via the set
\begin{align}
\label{eq:obstacle}
\mathcal{O}:=
\left\{ \rev{p}\in\R^3 \middle| \left(\frac{1}{3}\sum_{i=1}^3\left|\eta_i(p) \right|^{\alpha}\right)^\frac{1}{\alpha} \leq 1 \right\},
\end{align}
where a parameter~$\alpha\geq2$ defines the shape of the obstacle. \reb[2.6]{The smoothness parameter $\alpha=2$ corresponds to an ellipsoid and $\alpha\rightarrow\infty$ to a cuboid, see Fig.~\ref{fig:obstacles_shapes}.} Note that~\eqref{eq:obstacle} is a scaled norm, where lower values of \rev{$\alpha$} are always over-approximations of larger \rev{$\alpha$}, i.e., $2\leq\alpha_1\leq\alpha_2 \Rightarrow \mathcal{O}(\alpha_2)\subseteq\mathcal{O}(\alpha_1)$, cf.~\cite{reiter_2024a}. \rev{Accordingly, $d_o$ denotes the scale before the factor $3^{1/\alpha}$ induced by this normalized norm.}

\subsection{\reb{Convex Configuration Space Decomposition}}
\reb[4.5]{The collision constraints are most naturally expressed after a decomposition of the collision-free workspace, generally referred to as \emph{exact cell decomposition}~\cite{latombe1991exact,lavalle2006planning}. The general procedure is as follows. The boundary primitives of the inflated obstacles represent a partition of the workspace. For polyhedral obstacles, these primitives are supporting hyperplanes of obstacle facets. For curved obstacles, a conservative polyhedral approximation is used. Intersecting this arrangement with the workspace yields a set of cells whose interiors are collision-free. Each cell is convex whenever it is defined as the intersection of half-spaces from one arrangement region, and adjacent cells that share the same active inequalities can be merged to reduce the overall number of convex cells. The resulting free space~$\mathcal{F}$ is represented as
	\[
	\mathcal{F} = \bigcup_{j=1}^{n_c} C_j,
	\qquad
	C_j = \{ p \in \mathbb{R}^3 \mid H_j p \le h_j \},
	\]
	where $C_j$ is a convex polytope, $n_c$ are the number of convex sets, and $H_j\in\R^{m_j\times 3}$ and $h_j\in\R^{m_j}$ describe $m_j$ halfspace constraints.}

\reb[4.5]{Without loss of generality, we consider axis-aligned box obstacles. After obstacle inflation and clipping to the workspace, all obstacle faces and workspace faces define coordinate-wise cut locations to rectangular voxels. Each elementary box induced by these cuts is classified as free or occupied according to whether its midpoint lies in obstacle-free space; contiguous free boxes are first grouped into maximal connected segments along one coordinate direction and are then merged across shared faces in the remaining directions until no further convex merging is possible. Because all cuts are axis-aligned, each final cell is an axis-aligned box
	\[
	C_j = \{ p \in \mathbb{R}^3 \mid \ell_j \le p \le u_j \},
	\]
	which is a particularly simple polyhedral representation of the exact cell decomposition induced by the inflated obstacles. Additionally, a cell adjacency graph is constructed: two cells $C_i$ and $C_j$ are adjacent if their axis-aligned bounding boxes touch or overlap on every coordinate axis.}

\reb[2.3,4.8]{\paragraph{Mixed-integer encoding.}
	Let $\beta_{k_b,j} \in \{0,1\}$ denote whether the trajectory at binary step~$k_b$ is assigned to cell~$C_j$. To reduce the number of binary variables, cell-assignment binaries are introduced only at a subset of nodes $\mathcal{K}_b = \{s,\, 2s,\, \ldots\} \cup \{N\} \subseteq \{1,\ldots,N\}$ with stride~$s \ge 1$. Each node~$k$ inherits the assignment from the nearest preceding binary step $\kappa(k) = \max\{k_b \in \mathcal{K}_b : k_b \le k\}$, so that the total number of binary variables is $|\mathcal{K}_b|\, n_c$ instead of $N\, n_c$.
	Exactly one cell is selected at each binary step,
	\begin{equation}
	\label{eq:sum_big_m}
	\sum_{j=1}^{n_c} \beta_{k_b,j} = 1, \qquad k_b \in \mathcal{K}_b,
	\end{equation}
	and the position at every node is forced into the selected cell via indicator constraints,
	\begin{align}
	\label{eq:constriants_miqp}
	\beta_{\kappa(k),j} = 1 \;\Rightarrow\; \ell_j \le p_k \le u_j,
	\; j=1,\dots,n_c,\;\; k=1,\dots,N.
	\end{align}
	To ensure spatial consistency across the horizon, cell transitions between consecutive binary steps are restricted to neighboring cells in the adjacency graph,
	\begin{equation}
	\label{eq:cell_transition}
	\sum_{i \in \mathcal{N}(j)\cup\{j\}} \beta_{k_b',i} \;\ge\; \beta_{k_b,j},
	\qquad j=1,\dots,n_c,
	\end{equation}
	for each pair of successive binary steps $k_b, k_b' \in \mathcal{K}_b$, where $\mathcal{N}(j)$ denotes the set of cells adjacent to $C_j$.}

\reb[2.3,4.8]{\paragraph{Online pruning.}
	Before each solve, conservative per-axis reachable position bounds are propagated from the current state~$x_0$ over the planning horizon. At each binary step~$k_b$, cells whose bounding boxes do not overlap the reachable region are fixed to $\beta_{k_b,j}=0$, which substantially reduces the effective number of active binary variables without altering feasibility.
	For node $p_k$ and binary variables $\beta_{\kappa(k)}^\top=[\beta_{\kappa(k),1},\ldots,\beta_{\kappa(k),n_c}]$ the constraints~\eqref{eq:sum_big_m}--\eqref{eq:cell_transition} and the reachability fixing are summarized by $(\lf{P}_p\lf{x}_k,\beta_{\kappa(k)})\in\mathcal{F}$.}

\subsection{Nominal MPC}
\label{sec:nominal_mpc}
With sufficient computational resources and a sufficiently good initial guess, we would solve a nominal MPC problem using the highest-fidelity model available.
We formulate a multiple-shooting \ac{NLP} to solve the MPC problem with state variables~$\hf{X}=\begin{bmatrix}
\hf{x}_0&\ldots&\hf{x}_M
\end{bmatrix}$ and \rev{control variables} $\hf{U}=\begin{bmatrix}
\hf{u}_0&\ldots&\hf{u}_{M-1}
\end{bmatrix}$, and use numerical integration to obtain the consecutive states~$x_{k+1}$ at discrete times $k\hf{t}_\Delta$ via $x_{k+1}=F(x_k,u_k;\hf{t}_\Delta)$, particularly an RK4 scheme with sampling time~$\hf{t}_\Delta$. 
A terminal value function~$\Phi(\hf{x})$ and a terminal set~$h_M(\hf{x}_M)$ approximate the infinite horizon and \rev{can establish recursive feasibility under the standard terminal-set invariance assumptions}. With state selection matrices~$P_p\in\R^{3\times17}$, $P_f\in\R^{4\times17}$, $P_v\in\R^{3\times17}$ and~$P_\omega\in\R^{3\times 17}$ that select the states~$p$, \rev{$f_z$}, $v$ and~$\omega$ of the full state~$x$, the nominal MPC problem is defined as
\begin{mini!}[1]
	{\hf{X}, \hf{U}}{\sum_{k=0}^{M-1} l(\hf{x}_k, \hf{u}_k) + \Phi(\hf{x}_M)}{\label{ocp: learned vfn}}{}
	\addConstraint{\hf{x}_0}{=\hf{\hat{x}}_0}
	\addConstraint{\hf{x}_{k+1}}{= F(\hf{x}_k, \hf{u}_k;\hf{t}_\Delta),}{\; k \in \Z_{[0, M-1]}}
	\addConstraint{P_\omega\hf{x}_k}{\in\hf{\mathbb{S}}_\omega,\;P_f\hf{x}_k\in\hf{\mathbb{S}}_f,\;P_v\hf{x}_k\in\hf{\mathbb{S}}_v}{\; k \in \Z_{[0, M-1]}}
	\addConstraint{P_p\hf{x}_k}{\not\in\rev{\mathcal{O}},}{\; k \in \Z_{[0, M-1]}}
	\addConstraint{0}{\geq h_M(\hf{x}_M).}{}
\end{mini!}
We define the nonlinear least-squares cost
\begin{equation}
\label{eq:cost_hf}
	l(x,u)=t_\Delta(\rev{y(x,u)}-\tilde{y})^\top \mathrm{diag}(w)(\rev{y(x,u)}-\tilde{y}),
\end{equation}
with the tracking states $\rev{y^\top(x,u)} = \begin{bmatrix}
p & \phi(q,\tilde{q}) & v & \omega & f_z & u
\end{bmatrix}\in\R^{20}$ that includes the quaternion error~$\phi:\R^{4\times4}\rightarrow\R^3$ transformed to Euler angles, the reference~$\tilde{y}\in\R^{20}$ and the weight vector~$w\in\R^{20}$.

\section{Low-Fidelity Model}
\label{sec:low_level_model}
For long-horizon trajectory planning, low-fidelity models are inevitable due to the necessary trade-off between accuracy and computation time per planner iteration.
Due to the differential flatness of the quadrotor motion, a chain of integrators is often used to approximately describe the quadrotor motion~\cite{hehn_2015,tal_2021,li_2021}. In this work, we use a chain of three integrators to model the trajectory of a quadrotor, as proposed in~\cite{hehn_2015} for rapid planning of feasible trajectories. Three integrators, combined with specific constraints, are necessary to generate feasible trajectories for quadrotors~\cite{hehn_2015}. \rev{For readability}, we overload the notation for positions~$p$ and velocities~$v$ as for the quadrotor model, and add acceleration states~$a\in\R^3$ and jerk inputs
\(
\lf{u}:=j\in\R^3.
\)
to obtain the low-fidelity model
\begin{equation}
\lf{\dot{x}}=\begin{bmatrix}
\dot{\xPos} &
\dot{\xVel} &
\dot{\xAcc} 
\end{bmatrix}^\top=
\begin{bmatrix}
\xVel &
\xAcc &
\uJerk
\end{bmatrix}^\top:=
\lf{f}(\lf{x},\lf{u}),
\end{equation}
with the low-fidelity state
\(\lf{x}=\begin{bmatrix}
p & v & a
\end{bmatrix}^\T.\)

\subsection{Relation to High-Fidelity Model}
Due to the \rev{differential flatness} of the quadrotor model~\cite{faessler_2018}, all quadrotor states of a simple model with linear residuals~\cite{faessler_2018} can be obtained from derivatives of the position and the orientation of the quadrotor model.
The point-mass model, which is a chain of integrators, can be seen as a lower-fidelity model from which we can derive the most important states for motion planning.
We are particularly interested in the quadrotor position~$p$, velocity~$v$, body rate~$\omega$, and thrust~$T$. These states are typically associated with the optimization objective and constraints related to physical feasibility and safety, as in~\eqref{eq:con_br}, \eqref{eq:con_thrust} and~\eqref{eq:obstacle}.
A point-mass model formulates derivatives of the position, in our case, up to the jerk. These physical quantities are aligned with the quadrotor motion. From the quadrotor model~\eqref{eq:3d_quad_dynamics}, the acceleration can be directly obtained via
\begin{equation*}
	a_\mathrm{quad}(q,v,f_z)=\frac{1}{m} \Big({\xQuat} \odot f\big(\xQuat^{-1}\odot\xVel,\,\xF\big)\Big)  - g 
\end{equation*}
and the jerk via the derivative of the acceleration
\begin{align*}
	j_\mathrm{quad}(q,v,\omega,f_z, \dot{f}_z)=\dot{a}_\mathrm{quad}(q,v,f_z),
\end{align*}
with details in the supplementary material.
Without the consideration of residual forces, i.e., assuming~$f_\mathrm{res}=0$, the quadrotor acceleration simplifies to
\begin{equation*}
	\hat{a}_\mathrm{quad}(q,f_z)=\frac{1}{m} \Big({\xQuat} \odot T(f_z)\Big)  - g ,
\end{equation*}
	with the \rev{scalar collective thrust $T_z(f_z):=e_z^\top T(f_z)=\sum_{i=1}^4 f_{z,i}\geq0$}. Taking its derivative gives \rev{$\dot{T}_z(f_z)=\sum_{i=1}^4\dot f_{z,i}$}, and the jerk follows as
\begin{align*}
	\hat{j}_\mathrm{quad}(q,\omega,f_z, \dot{f}_z)=\xQuat\odot\begin{bmatrix}
		\rev{T_z(f_z)} \omega_y\\ \rev{-T_z(f_z)} \omega_x \\ \rev{\dot T_z(f_z)}
	\end{bmatrix}.
\end{align*}

\subsection{Point-Mass System Constraints}
\label{sec:pm_constraints}
The authors in~\cite{hehn_2015} derived constraints 
\[m\|a+g\|\leq\ub{f}_\mathrm{th}\]
on a third-order point mass model that guarantees feasible trajectories for a quadrotor model with constraints defined in~\eqref{eq:con_br_thrust} but without residual forces. We extend these constraints to include feasibility for the residual model by utilizing the upper bound~$\ub{f}_\mathrm{res}$ from Ass.~\ref{ass:bounded_disturbance}. While~\cite{hehn_2015} required decoupled constraints per axis due to their computationally efficient planning algorithm, we utilize nonlinear programming and successive linearization, which allows for coupled, less restrictive constraints. Particularly, the constraint~$\hf{\mathbb{S}}_\mathrm{f}$ on the collective thrust~\eqref{eq:con_thrust} can be formulated with the individual acceleration components~$a_x,a_y$ and $a_z$ per axis in the world frame for the low-fidelity planning model. The collective constraint for the low-fidelity model is therefore
\begin{equation}
\label{eq:th_constraint_pm}
\mathbb{S}_\mathrm{f}^\mathrm{lf}:=\big\{a\in\R^3\mid \rev{\lb{f}_\mathrm{th}+\ub{f}_\mathrm{res}\leq m\|a+g\|\leq\ub{f}_\mathrm{th}-\ub{f}_\mathrm{res}}\big\}.
\end{equation}

Body rate constraints $\hf{\mathbb{S}}_{\xBr}$ formulated in~\eqref{eq:con_br} for the high-fidelity model can be incorporated in the low-fidelity point mass model via constraints on the time derivative of the unit vector of the gravity-shifted acceleration, i.e.,
\begin{equation}
\label{eq:br_constraint_pm_1}
\bigg\|\frac{d}{dt}\bigg(\frac{a+g}{\|a+g\|}\bigg)\bigg\| \leq \ub{\omega}_{xy}.
\end{equation}
Resolving the time derivative in~\eqref{eq:br_constraint_pm_1} and utilizing the explicit notation of jerk with~$j=\dot{a}+\dot{g}=\dot{a}$, \rev{a conservative sufficient} body rate feasibility constraint for the low-fidelity model can be written as
\begin{equation}
\label{eq:br_constraint_pm_2}
\mathbb{S}_\mathrm{\omega}^\mathrm{lf}:=\left\{\begin{bmatrix}
a \\ j
\end{bmatrix}\in\R^6\middle| \frac{\|j\|}{\|a+g\|}\leq \ub{\omega}_{xy}\right\}.
\end{equation}
The auxiliary constraints on the thrust~\eqref{eq:th_constraint_pm} and the body rate~\eqref{eq:br_constraint_pm_2} are highly nonlinear. However, they can be conservatively reformulated into more optimization-friendly subsets, as shown in the following, with details in~\cite{hehn_2015}.

First, we simplify the lower bound of the thrust constraint~$\mathbb{S}_\mathrm{f}^\mathrm{lf}$ defined in~\eqref{eq:th_constraint_pm}. 
The lower bound of~\eqref{eq:th_constraint_pm} is satisfied if the simpler inequality \[a_z\geq\lb{a}_z\geq\rev{\frac{\lb{f}_\mathrm{th}+\ub{f}_\mathrm{res}}{m}-g},\]
holds, where $\lb{a}_z$ is an auxiliary, more restrictive, bound that is also used in the following to shape the simplified constraints. We refer to~\cite{hehn_2015} for a proof.

Second, the upper bound of~\eqref{eq:th_constraint_pm} can be simplified, where a more restrictive constraint guarantees the feasibility of~\eqref{eq:th_constraint_pm}. The simplified constraint involves two design parameters~$\rev{\alpha_x,\alpha_z}\in(0,1)$ that shape and trade off the more optimization-friendly constraints. \rev{Define the available thrust margin as $\ub{f}_\mathrm{av}:=\ub{f}_\mathrm{th}-\ub{f}_\mathrm{res}$.}
By utilizing the design parameters, box-constraints for $\ub{a}^\top=\begin{bmatrix}
\ub{a}_x & \ub{a}_y & \ub{a}_z
\end{bmatrix}$ can be computed iteratively by
\begin{align*}
\ub{a}_z &:= \alpha_z \left(\frac{1}{m} \rev{\ub{f}_\mathrm{av}}-g\right), \\
\ub{a}_x &:= \alpha_x \sqrt{\left(\frac{1}{m} \rev{\ub{f}_\mathrm{av}}\right)^2-\left(\ub{a}_z+g\right)^2}, \\
\ub{a}_y &:= \sqrt{\left(\frac{1}{m} \rev{\ub{f}_\mathrm{av}}\right)^2-\ub{a}_x^2-\left(\ub{a}_z+g\right)^2}.
\end{align*}

The simplified box constraints, expressed as a set constraining the acceleration states of the low-fidelity model, can finally be written as
\begin{equation}
\label{eq:f_constraint_2}
\hat{\mathbb{S}}_\mathrm{f}^\mathrm{lf}:=\big\{a\in\R^3\mid \begin{bmatrix}
-\ub{a}_x & -\ub{a}_y & \lb{a}_z
\end{bmatrix} \leq \rev{a}\leq \ub{a} \big\}\subset\mathbb{S}_\mathrm{f}^\mathrm{lf},
\end{equation}
which is \rev{a subset} of the thrust constraints~\eqref{eq:th_constraint_pm}.

Next, we simplify the constraints of~$\lf{\mathbb{S}}_\omega$, defined in~\eqref{eq:br_constraint_pm_2}.
A more optimization-friendly subset of~\eqref{eq:br_constraint_pm_2} can be found by exploiting the lower bound on the acceleration
\[\|a+g\|\geq \rev{\lb{a}_z+g}\] to obtain simplified convex constraints on the jerk~$j$
\begin{equation}
\label{eq:br_constraint_pm_3}
\hat{\mathbb{S}}_\mathrm{\omega}^\mathrm{lf}:=\left\{ 
 j
\in\R^3 \middle| 
\|j\|\leq(\lb{a}_z+g)\ub{\omega}_{xy}
\right\}\subseteq \lf{\mathbb{S}}_\mathrm{\omega}.
\end{equation}
Arbitrary box constraints for each individual component of the jerk can be computed, forming a subset of~\eqref{eq:br_constraint_pm_3} and, therefore, decoupling the constraints. A valid choice would be~$\ub{j}_{xyz}=\frac{\lb{a}_z+g}{\sqrt{3}}\ub{\omega}_{xy}$ to obtain the convex decoupled box constraints
\begin{equation}
\label{eq:br_constraint_pm_4}
\widehat{\mathbb{S}}_\mathrm{\omega}^\mathrm{lf}:=\left\{ 
j
\in\R^3 \middle| 
|j_{\{x,y,z\}}|\leq \ub{j}_{xyz}
\right\}\subseteq\lf{\hat{\mathbb{S}}}_\mathrm{\omega}.
\end{equation}

\section{\textsc{Unique}: The Temporal Cascading MPC Formulation}
\label{sec:unique}
\reb[4.1]{In the following, we propose \textsc{Unique}, a multi-phase MPC framework that embeds the planning problem as the tail horizon of the controller.} \reb[4.1]{Rather than solving the planning problem hierarchically on a separate model and tracking the result, we append the low-fidelity point-mass model from Sect.~\ref{sec:low_level_model} as a second phase after the high-fidelity MPC from Sect.~\ref{sec:nominal_mpc}, forming a single optimization that jointly plans and controls.}
The merging of two different models along the MPC horizons requires
(i) the alignment of the cost function,
(ii) recursive feasible constraint formulations,
(iii) \rev{a transition function} between the models.
Very long prediction horizons result in MPC optimization problems with ubiquitous local minima, which may degrade the performance of the multi-phase MPC, particularly in cluttered environments with nonsmooth obstacle shapes.
Two proposed strategies effectively improve performance as part of the \textsc{Unique} framework, i.e., progressive smoothing and parallel low-fidelity optimization from random initial seeds. These strategies and the requirements for merging the two models over the horizon are detailed below.

\subsection{Cost Alignment}
The high fidelity weights~\[w^\top=[w_p^\top, w_\phi^\top, w_v^\top, w_\omega^\top, w_f^\top, w_u^\top]\] of the nonlinear-least square cost~\eqref{eq:cost_hf} are split into groups for physical states, with~$w_p\in\R^3$ for positions, $w_v\in\R^3$ for velocities, $w_\phi\in\R^3$ for Euler angles, $w_\omega\in\R^3$ for body rates, $w_f\in\R^4$ for single rotor thrusts and $w_u\in\R^4$ for its derivatives.
The cost for the low-fidelity model is formulated as a linear least-squares cost for the full low-fidelity state via
\begin{equation}
	\lf{l}(\lf{x},\lf{u}) = \lf{t}_\Delta
	(\lf{y}-\lf{\bar{y}})^\top \mathrm{diag}(\lf{w}) (\lf{y}-\lf{\bar{y}}),
\end{equation}
with $\lf{y}=[\lf{x},\lf{u}]^\top$ and the weights
\[\lf{w}=\begin{bmatrix}
w_p^\top&w_v^\top&w_a^\top&w_j^\top
\end{bmatrix}^\top,\]
that are partially aligned with the high-fidelity weights. In particular, the weights~$w_p$ and $w_v$ are associated with the same physical states and are therefore chosen to be identical.

In the high-fidelity cost~\eqref{eq:cost_hf}, single-rotor vertical thrusts are penalized by a weighted sum of squares. \rev{For equal rotor-thrust component weights $w_{f,0}$ and equal sharing of a collective-thrust variation, $\sum_{i=1}^4(\Delta f_{z,i})^2=m^2\|\Delta a\|^2/4$. We therefore use the approximate acceleration-weight alignment $w_a:=m^2w_{f,0}/4$; attitude, gravity, and residual forces prevent an exact global equivalence.}

The jerk weight~$w_j$ is related to the body rate weight~$w_\omega$ and the force derivative input weight~$\rev{w_u}$. Similar to the acceleration weight, we take the input weight relation~$w_j\approx m^2 w_u$ as an approximation and increase it in closed-loop experiments to account for the body rate weight.

For the terminal value function~$\lf{\Phi}:\R^9\rightarrow\R$, we define a least-square penalty on the final reference position~$\lf{p}_T\in\R^3$ and the weight~$q_T\in\R^3$, i.e., \[\lf{\Phi}(\lf{x}):=(P_p \lf{x}-\lf{p}_T)^\top\mathrm{diag}(q_T)(P_p \lf{x}-\lf{p}_T)\] and a terminal safe set~$\lf{h}(\lf{x})$.
With the simple assumption of considering hovering as a safe state, we can use the corresponding low-level states~$P_{va}\lf{x}$, with the projection matrix $P_{va}\in\R^{6\times9}$ selecting velocities and accelerations, to establish a simple safe terminal constraint \(\lf{h}(\lf{x})=P_{va}\lf{x}=0.\)

\subsection{Constraint Alignment}
\label{sec:constraint_alignment}
As shown in Sect.~\ref{sec:pm_constraints}, we can find constraint sets~$\lf{\mathbb{S}}_\omega$ and $\lf{\mathbb{S}}_f$ for the low-fidelity model that guarantee a feasible trajectory of the high-fidelity model that is constrained via the feasible sets~$\hf{\mathbb{S}}_\omega$ and $\hf{\mathbb{S}}_f$. We propose two subset variants of the low-fidelity set~$\lf{\mathbb{S}}_\omega$ that are more tightly constrained yet numerically easier to solve. The set~$\lf{\hat{\mathbb{S}}}_\omega$ assumes a lower bound~$\lb{a}_z$, leading to a convex quadratic constraint on the jerk~$j$~\eqref{eq:br_constraint_pm_3} and the set~$\widehat{\mathbb{S}}_\mathrm{\omega}^\mathrm{lf}$ decouples the jerk components to box constraints~\eqref{eq:br_constraint_pm_4}. The constraint alignment guarantees that the plan in the second horizon is a feasible plan for the higher-fidelity drone model.

\subsection{Transition Function}
The conditions on the low-fidelity model of the previous Sect.~\ref{sec:constraint_alignment} provide necessary conditions on the position trajectory and its derivatives of the high-fidelity model for feasibility. In this section, we develop necessary conditions for the final state of the high-fidelity model~$\hf{x}_M$, such that the low-fidelity plan for future times~$t\geq M t_\Delta$ can be tracked. 
Our approach explicitly couples all derivatives of the low-fidelity trajectory, up to and including jerk, to the corresponding high-fidelity states.
First, we require the state and velocities of both fidelity models to be aligned via
\begin{equation}
\label{eq:trans_1}
\rev{\pi_{pv}}\left(x,\lf{x}\right):=
\rev{\begin{bmatrix}
P_p \hf{x}\\
P_v \hf{x}
\end{bmatrix}-
\begin{bmatrix}
\lf{P}_p \lf{x}\\
\lf{P}_v \lf{x}
\end{bmatrix}}=0.
\end{equation}
Next, we require the acceleration obtained from the rotor thrusts to be aligned with the acceleration of the low-fidelity model, via
\begin{equation}
\label{eq:trans_2}
\pi_{a}\left(x,\lf{x}\right):=
\rev{a_\mathrm{quad}\!\left(P_q\hf{x},P_v\hf{x},P_f\hf{x}\right)}-P_a \lf{x} = 0.
\end{equation}
Finally, the body rates need to be aligned with the jerk and the acceleration. According to the derivation in~\cite{hehn_2015}, the body rates of the quadrotor model can be obtained from the \rev{gravity-shifted acceleration $s(a):=a+g$} and the jerk~$j=u^\mathrm{lf}$ via
\begin{equation}
	\label{eq:trans_3_pre}
	\small
	\begin{bmatrix}\omega_x&-\omega_y&0\end{bmatrix}^\top
	=q\odot\rev{\left(\frac{j}{\|s(a)\|}-\frac{s(a)s(a)^\top j}{\|s(a)\|^3}\right)}.
\end{equation}

With \rev{$s_\mathrm{lf}:=P_a\lf{x}+g$}, this leads to the constraint
\begin{equation}
\label{eq:trans_3}
\small
\begin{aligned}
\pi_\omega\left(x,\lf{x},\lf{u}\right):={}&
\mathrm{diag}\left([1,-1,0]^\top\right)P_\omega x-P_qx\odot\\[-1mm]
&\rev{\left(\frac{\lf{u}}{\|s_\mathrm{lf}\|}-
\frac{s_\mathrm{lf}s_\mathrm{lf}^\top\lf{u}}{\|s_\mathrm{lf}\|^3}\right)}=0.
\end{aligned}
\end{equation}
We require \rev{$\|a+g\|>0$} in~\eqref{eq:trans_3_pre} and~\eqref{eq:trans_3}. \rev{This follows from the positive lower collective-thrust bound in~\eqref{eq:th_constraint_pm}.}
Additionally, the jerk-related constraint~\eqref{eq:trans_3} can be interpreted as a constraint to the preceding jerk control before the initial transition state~$\lf{x}_0$, thus it is not required for the feasibility of the quadrotor model.
Finally, we summarize the relevant coupling conditions of~\eqref{eq:trans_1} and \eqref{eq:trans_2} by
\begin{equation*}
	\pi\left(\hf{x},\lf{x}\right):=\begin{bmatrix}
	\pi_{pv} \left(\hf{x},\lf{x}\right)&
	\pi_{a} \left(\hf{x},\lf{x}\right)
	\end{bmatrix}^\top=0.
\end{equation*}

\subsection{Progressive Smoothing}
Within~\textsc{Unique}, we aim to plan long-horizon trajectories via numerical optimization, which is inherently prone to getting stuck in local minima, particularly when encountering non-smooth obstacles, such as cubes. To improve convergence, we adapt the progressive smoothing method proposed by~\cite{reiter_2024a} for autonomous driving to the three-dimensional planning domain. In progressive smoothing, potentially non-smooth obstacle shapes~$\mathcal{O}(\alpha)$ from \eqref{eq:obstacle} are progressively \rev{smoothed} along the MPC planning horizon by scheduling the smoothing parameter~$\alpha$ towards a smooth 2-norm. Particularly, a \rev{monotonically} decreasing scheduling function~$\lambda(t)$ with~$\lambda:\R^+\rightarrow\R^+$ depending on the MPC prediction time~$t$ is used, with $\lambda(0)=\alpha_0$ and $\lambda(t_f)=2$, where $t_f$ is the total prediction horizon and~$\alpha_0$ describes the actual obstacle shape.
\rev{The smoothing improves numerical behavior of the \ac{RTI} scheme~\cite{diehl_2005}. Recursive feasibility additionally requires the usual shift, terminal-set, and hard-constraint assumptions.}

\subsection{Multi-Phase MPC formulation}
The multi-phase formulation optimizes the short-horizon high-fidelity states $(\hf{X},\hf{U})$ together with the long-horizon low-fidelity states $(\lf{X},\lf{U})$ in the single NLP:
\begin{algbox}{Main Multi-Phase MPC}
\begin{mini!}[3]
	{\substack{\hf{X}, \hf{U},\\ \lf{X}, \lf{U}}}{\sum_{k=0}^{M-1} l(\hf{x}_k, \hf{u}_k) +\sum_{k=0}^{N-1} \lf{l}(\lf{x}_k, \lf{u}_k)+ \lf{\Phi}(\lf{x}_N)}{\label{eq:ocp_unified}}{}
	\addConstraint{\hf{x}_0}{=\hf{\hat{x}}_0}
	\addConstraint{0}{=\pi\left(\hf{x}_M,\lf{x}_0\right)}
	\addConstraint{\hf{x}_{k+1}}{=F\left(\hf{x}_k, \hf{u}_k;\hf{t}_\Delta\right),}{\; k \in \Z_{[0, M-1]}}
	\addConstraint{\lf{x}_{k+1}}{= \lf{F}\left(\lf{x}_k, \lf{u}_k;\lf{t}_\Delta\right),}{\; k \in \Z_{[0, N-1]}}
	\addConstraint{P_{f}\hf{x}_k}{\in\hf{\mathbb{S}}_f,P_\omega\hf{x}_k\in\hf{\mathbb{S}}_\omega,P_v\hf{x}_k\in\hf{\mathbb{S}}_v}{\;k \in \Z_{[0, M-1]}}
	\addConstraint{\rev{P_a\lf{x}_k}}{\in\lf{\mathbb{S}}_f,\;\rev{[P_a\lf{x}_k;\lf{u}_k]}\in\lf{\mathbb{S}}_\omega,}{\; k \in \Z_{[0, N-1]}}
	\addConstraint{P_p\hf{x}_k}{\not\in\rev{\mathcal{O}}\left(\lambda (k\hf{t}_\Delta)\right),}{\; k \in \Z_{[0, M-1]}}
	\addConstraint{\lf{P}_p\lf{x}_k}{\not\in\rev{\mathcal{O}}\left(\lambda(M\hf{t}_\Delta+k\lf{t}_\Delta)\right),}{\; k \in \Z_{[0, N]}}
	\addConstraint{0}{= \lf{h}\left(\lf{x}_N\right).}{}
\end{mini!}
\end{algbox}
The cost function accumulates stage costs from both phases and a terminal cost on the long horizon. System dynamics are enforced by the respective discrete-time integration functions $F$ and $\lf{F}$, and the transition constraint $\pi(\hf{x}_M,\lf{x}_0)=0$ guarantees consistency of position, velocity, and thrust-induced acceleration. Actuator and body-rate feasibility is maintained through the sets $\hf{\mathbb{S}}_f,\hf{\mathbb{S}}_\omega$ for the high-fidelity model and their inner-approximated counterparts $\lf{\mathbb{S}}_f,\lf{\mathbb{S}}_\omega$ for the low-fidelity model. Both horizons are kept collision-free by enforcing position constraints against the obstacle set $\mathbb{O}(\lambda(\cdot))$, whose recursive feasibility is guaranteed~\cite{reiter_2024a}. Finally, the terminal equality $\lf{h}(\lf{x}_N)=0$ enforces a simple safe set at the goal. The influence of the terminal safe set would be much larger when applied directly after the high-fidelity horizon. 

\subsection{Parallel Low-Fidelity MPCs}
Above, we introduced the multi-phase formulation with progressive smoothing to plan for long horizons. In cluttered environments, solving the nonlinear program~\eqref{eq:ocp_unified} is still prone to getting stuck in local minima due to the long planning horizon and the gradient-based local optimization of MPC solvers~\cite{verschueren_acadosmodular_2022}. \reb[4.1]{Dealing with such nonconvexities is the pivotal point of many planning algorithms discussed in Section~\ref{sec:related}. Since the planning-relevant decision variables are tied dominantly to the second (tail) phase of our formulation, we can exploit this structure.} \reb[4.9]{A key design goal is to keep the main controller fast: the high-fidelity first phase must not be delayed by the combinatorial complexity of planning. Therefore, we offload the search for better tail trajectories to parallel asynchronous solvers that run concurrently with the main MPC.} Formulating this subproblem as a separate \ac{NLP} allows us to optimize parallel instances from different initial guesses and escape local minima of the dominant second planning horizon while sacrificing only minor computational resources.
\reb[2.1]{As an initialization strategy, we roll out $M_{\mathrm{p,ini}}=64$ trajectories from $\lf{\hat{x}}_0$ by simulating the low-fidelity model with an RK4 integrator and applying random jerks sampled from a uniform distribution~$j\sim \mathcal{U}(\lb{j},\ub{j})$. We then select $N_p$ candidates, prioritizing feasible trajectories and otherwise using low-cost infeasible ones.}

\reb[2.2]{To guarantee continuity, we start from the same initial point-mass state~$\lf{\hat{x}}_0$ obtained from the multi-phase MPC~\eqref{eq:ocp_unified}; the parallel MPC subproblems are}
\begin{algbox}{{\reb[3.4]{Parallel Point-Mass MPC}}}
\begin{mini}[1]
	{\substack{\lf{X}, \lf{U}}}{\sum_{k=0}^{N-1} \lf{l}(\lf{x}_k, \lf{u}_k)+ \lf{\Phi}(\lf{x}_N)}{\label{eq:ocp_parallel}}{J_\mathrm{lf}(\reb[3.4]{\lf{x}_0})=}
	\addConstraint{\lf{x}_0}{=\lf{\hat{x}}_0}
	\addConstraint{\lf{x}_{k+1}}{= \lf{F}\left(\lf{x}_k, \lf{u}_k;\lf{t}_\Delta\right),}{\; k \in \Z_{[0, N-1]}}
		\addConstraint{\rev{P_a\lf{x}_k}}{\in\lf{\mathbb{S}}_f,\;\rev{[P_a\lf{x}_k;\lf{u}_k]}\in\lf{\mathbb{S}}_\omega,}{\; k \in \Z_{[0, N-1]}}
		\addConstraint{\lf{P}_p\lf{x}_k}{\not\in\rev{\mathcal{O}}\left(\lambda(M\hf{t}_\Delta+k\lf{t}_\Delta)\right),}{\; k \in \Z_{[0, N]}}
	\addConstraint{0}{= \lf{h}\left(\lf{x}_N\right).}{}
\end{mini}
\end{algbox}

\reb[2.3]{Crucially for planning problems with severe nonconvexities, we add a particular parallel point-mass MPC variant that explicitly considers the nonconvexities in the free planning space by a decomposition into convex cells linked via binary indicator variables. The resulting MIQP formulation requires a dedicated solver.}
\begin{algbox}{{\reb[2.3]{Parallel Mixed-Integer MPC}}}
	\begin{mini}[1]
		{\substack{\lf{X}, \lf{U}, B}}{\sum_{k=0}^{N-1} \lf{l}(\lf{x}_k, \lf{u}_k)+ \lf{\Phi}(\lf{x}_N)}{\label{eq:ocp_miqp}}{J_\mathrm{mi}(\lf{x}_0)=}
		\addConstraint{\lf{x}_0}{=\lf{\hat{x}}_0}
				\addConstraint{\lf{x}_{k+1}}{= \lf{F}\left(\lf{x}_k, \lf{u}_k;\lf{t}_\Delta\right),}{\; \rev{0\leq k<N}}
					\addConstraint{\rev{P_a\lf{x}_k}}{\in\lf{\mathbb{S}}_f,}{\; \rev{0\leq k<N}}
					\addConstraint{\rev{[P_a\lf{x}_k;\lf{u}_k]}}{\in\lf{\mathbb{S}}_\omega,}{\; \rev{0\leq k<N}}
					\addConstraint{(\lf{P}_p\lf{x}_k,\rev{\beta_{\kappa(k)}})}{\in\mathcal{F}}{\; \rev{0\leq k\leq N}}
		\addConstraint{0}{= \lf{h}\left(\lf{x}_N\right).}{}
	\end{mini}
\end{algbox}

\rev{Due to the feasibility of the initial state from~\eqref{eq:ocp_unified}, all MPC subproblems provide recursive feasibility when combined with the high-fidelity horizon of~\eqref{eq:ocp_unified}. A proof under the standard shift and terminal-set assumptions is provided in the supplementary material.}
Once any of the MPC subproblems initialized with random primal variables~$({}^{0}\lf{X}, {}^{0}\lf{U})^i$ exhibits a feasible lower cost~$J_\mathrm{lf}\big(({}^{0}\lf{X}, {}^{0}\lf{U})^i,\lf{\hat{x}}_0\big)$ than the second horizon of~\eqref{eq:ocp_unified}, the second horizon of~\eqref{eq:ocp_unified} is reinitialized with the optimal lower cost decision variables of $i$-th parallel MPC subproblem.
Therefore, \rev{the accepted initialization has a lower evaluated tail cost before the next RTI step}. \reb[4.9]{\rev{Since the first horizon is unchanged during reinitialization, this reduces the evaluated objective at that instant; subsequent RTI steps and closed-loop shifts need not preserve a monotonic decrease.}} See~\cite{reiter_ac4mpc_2025} for details on parallel initialization and evaluation strategies.

\subsection{Algorithm and Numerical Efficiency}
\begin{figure}
	\centering
	\includegraphics[width=1.67in, trim=0.3cm 0.3cm 0.2cm 0.7cm, clip]{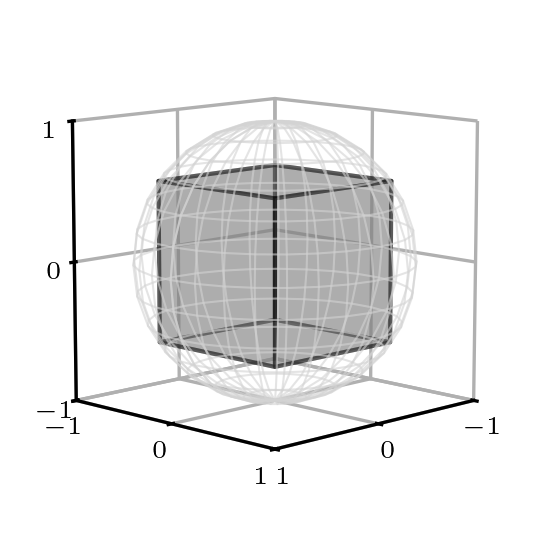}
	\includegraphics[width=1.67in, trim=0.3cm 0.3cm 0.2cm 0.7cm, clip]{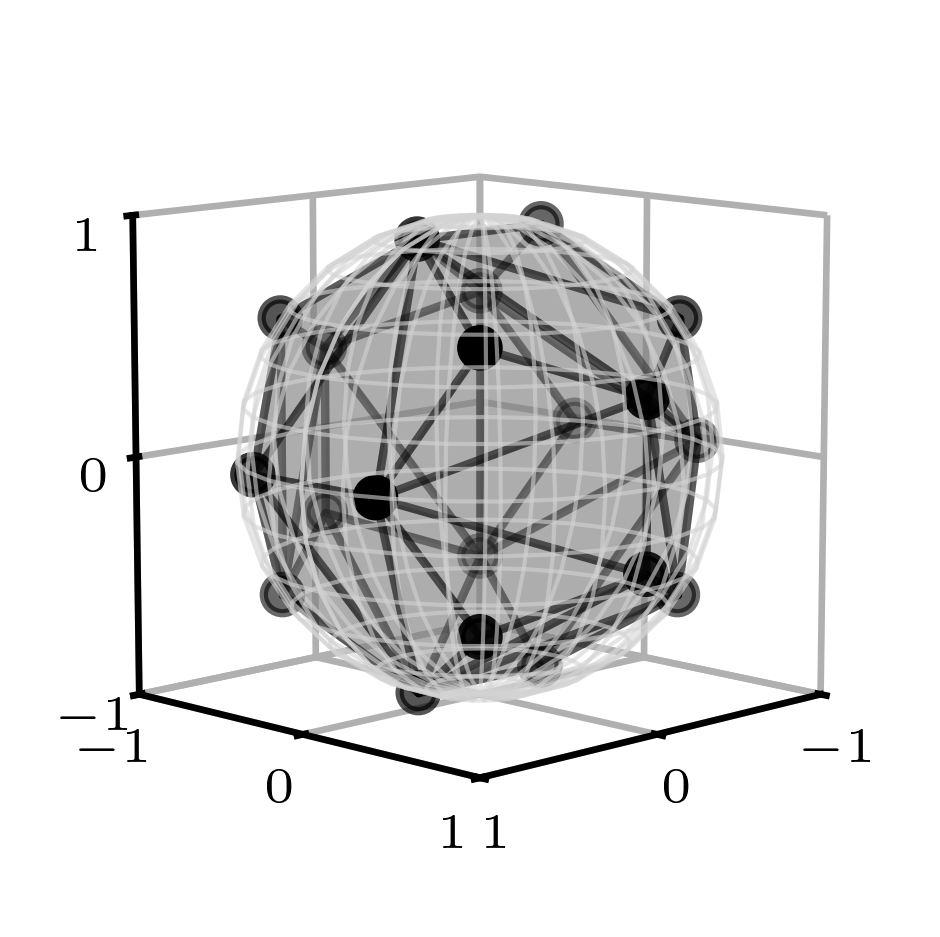}
	\caption{Visualization of inner-norm approximation. An inscribed cube covers less volume than the dodecahedron with 12 flat faces.}
	\label{fig:norm}
\end{figure}

The complete \textsc{Unique} algorithm is stated in Alg.~\ref{alg:unique}. We use the notation $I_k={k,k+1,\ldots,N+M+1}$ for the prediction indexes of the MPC.
\begin{algorithm}[t]
	\caption{\textsc{Unique}: Unified Multi-Fidelity MPC}
	\label{alg:unique}
	\begin{algorithmic}[1]
		\STATE \textbf{Inputs:} HF state $\hat{x}_0$, reference $\tilde{y}(t)$, obstacles $\{\mathcal{O}_i\}$, smoothing schedule $\lambda(t)$, number of parallel MPC restarts $S$, number of RTI~$P$ before reinitializing
		
		\STATE  \rev{Set $\alpha_i = \lambda(t_i)$ at each prediction time $t_i$, $i\in I_0$}
\reb{		\STATE  Decompose obstacles $\{\mathcal{O}_i\}$ to free space $\mathcal{F}$}
		\STATE \textbf{Repeat at each control step $k$:}
		\STATE \hspace{1em} \textbf{1. Obstacles update:}
		\STATE \hspace{2em} Set obstacle param. $\theta_i$ for prediction steps~$i\in I_k$
\reb{
		\STATE \hspace{1em} \textbf{2. Main multi-phase MPC:}
		\STATE \hspace{2em} Solve QP of multi-phase MPC at $x_k=\hat{x}_k$
		\STATE \hspace{2em} Evaluate second horizon cost: $(J_{\mathrm{lf},k})_0$
		\STATE \hspace{2em} Get initial l.f. state $\lf{\hat{x}}_0\gets\lf{x}_0$
}\reb{		
		\STATE \hspace{1em} \textbf{3. Parallel Mixed-Integer Planner:}
		\STATE \hspace{2em} Preprocess binary variables
		\STATE \hspace{2em} Solve MIQP of \rev{point-mass planner} at $(\lf{x}_0)_\mathrm{mi}:=\lf{\hat{x}}_0$
		\STATE \hspace{2em} Evaluate second horizon cost: $(J_{\mathrm{lf},k})_\mathrm{mi}$
}\reb{			
		\STATE \hspace{1em} \textbf{4. Parallel Point-Mass MPCs:}
		\STATE \hspace{2em} Solve $j=1:S$ point-mass MPCs at $(\lf{x}_0)_j:=\lf{\hat{x}}_0$
		\STATE \hspace{2em} Evaluate parallel point-mass MPC costs: $(J_{\mathrm{lf},k})_j$
}\reb{			
		\STATE \hspace{1em} \textbf{5. Evaluate \& Reinitialize:}
		\STATE \hspace{2em} Get lowest cost $\lambda^\star=\arg\min_{\lambda\in{0,\ldots,S,\mathrm{mi}}}(J_{\mathrm{lf},k})_{\lambda}$
}\reb{			
		\STATE \hspace{3em} Reinitialize multi-phase MPC\\ 
		\STATE \hspace{4em}$(\lf{x}_i)_0 \gets (\lf{x}_i)_{\lambda^\star}\;\forall i\in\{0,\ldots,N\}$
		\STATE \hspace{4em}\rev{$(\lf{u}_i)_0 \gets (\lf{u}_i)_{\lambda^\star}\;\forall i\in\{0,\ldots,N-1\}$}
}\reb{			
		\STATE \hspace{2em} \textbf{if} $k\mod P = 0$ \textbf{then}
		\STATE \hspace{3em} Randomly reini. parallel p.m. MPCs $j=1:S-1$
		\STATE \hspace{3em} Reini. $S$-th parallel p.m. MPC with MIQP solution
}
		\STATE \hspace{1em} \textbf{6. Apply and shift:}
		\STATE \hspace{2em} Execute first main multi-phase control~$u_0$
		\STATE \hspace{2em} Shift primal variables of all MPCs
	\end{algorithmic}
\end{algorithm}
In order to solve~\eqref{eq:ocp_unified} numerically efficiently, we use the solver~\textsc{acados}~\cite{verschueren_acadosmodular_2022} with the interior-point \ac{QP} solver \textsc{HPIPM}~\cite{frison_hpipm_2020} without condensing, and deploy the \ac{RTI} scheme~\cite{diehl_2005}. \reb[3.2]{We do not use condensing, since for the considered problem size we did not observe an improvement.} We use slack variables with exact \rev{$L_1$ and quadratic $L_2$ penalties}~\cite{nocecal_numerical_2006} for obstacle avoidance constraints and approximate convex quadratic constraints of the sets~$\hat{\mathbb{S}}_\mathrm{f}^\mathrm{lf}$ and $\hat{\mathbb{S}}_\mathrm{\omega}^\mathrm{lf}$ via a polyhedron~$\{x\in\R^3\mid b_\mathrm{p}\leq A_\mathrm{p}x\}$, with $b_\mathrm{p}\in\R^{12}$ and $A_\mathrm{p}\in\R^{12\times 3}$. In particular, we use a dodecahedron with 12 surfaces instead of the six surfaces of a cube approximation as in~$\widehat{\mathbb{S}}_\mathrm{\omega}^\mathrm{lf}$ and $\hat{\mathbb{S}}_\mathrm{f}^\mathrm{lf}$ to reduce the conservatism of the inner approximation by adding more constraints, as shown in Fig.~\ref{fig:norm}. The approximation via a polyhedron allows the QP solver to safely handle the norm constraints, which is not possible when linearizing the constraints within a single QP subroutine of \ac{SQP}. Notably, the velocity constraint~$\mathbb{S}_v$ spans the whole practical operating region of the quadrotor and can therefore be removed in~\eqref{eq:ocp_unified}.
\reb[2.5]{Remarkably, we normalize quaternions before using them to initialize the MPC problem, but do not explicitly constrain them to the unit sphere within the MPC, as we observed convergence problems and no performance benefit for converged solutions.}

\section{Experimental Setup}
\label{sec:experimental_setup}
The proposed algorithm is evaluated in both real-world and simulation environments using a realistic model that accounts for aerodynamic effects, as detailed in~\cite{foehn_agilicious_2022} and the supplementary material.
For both real-world experiments and simulations, we utilize the high-end \emph{Offboard} drone. 
The specifications are summarized in Tab.~\ref{tab:offboard_specs}. The drone is controlled from a ground station via the low-latency Crossfire protocol (CRSF). The overall latency of the pipeline, from state estimate to control command, is typically around \unit[20]{ms}, including all processing and transmission delays with CRSF.
The drone is equipped with the \texttt{Betaflight}~\cite{betaflight} lowest-level onboard controller, which tracks input body rates and total thrust. Therefore, the total thrust obtained from the \rev{rotor-thrust states} is used as the reference for the lowest-level controller.
The flight experiments are conducted in a large industrial hall equipped with a motion-capture system providing state estimation with millimeter accuracy at \unit[400]{Hz} across a volume spanning $\unit[25 \times 12 \times 4]{m}$.
\begin{table}[t]
\caption{Specifications of the ``Offboard'' Quadrotor}
\label{tab:offboard_specs}
\begin{minipage}[c]{0.65\linewidth}
    \begin{tabularx}{1\linewidth}{X>$l<$|r>$l<$}
        \toprule
        Parameter & & Value & \text{Unit} \\ \midrule \grayrow
        Geometry &[r_{p,i}]_x & $\pm$ 7.5 & \unit{cm} \\  \grayrow
         &[r_{p,i}]_y & $\pm$ 10.0 & \unit{cm} \\  
        Mass & m & 0.6 & \unit{kg} \\  \grayrow
        Inertia & J_{x} & 2.4 & \unit{g m^2} \\  \grayrow
        & J_{y} & 1.8 & \unit{g m^2} \\ \grayrow
        & J_{z} & 3.8 & \unit{g m^2} \\ 
        Motor Const. & c_l & 1.6 & \unit{\mu N s^2} \\  
	     & \kappa & 11.0 & \rev{\unit{mm}} \\  \grayrow
        Quad. Diag. & l & 38.2 & \unit{cm} \\
        Max Thrust & \ub{f}_{\mathrm{th}} & 34.0 & \unit{N} \\  \grayrow
        Max Rate & \ub{\omega}_{xy} & 10.0 & \unit{s^{-1}} \\   \grayrow
         & \ub{\omega}_{z} & 6.0 & \unit{s^{-1}} \\ 
        \bottomrule
    \end{tabularx}
\end{minipage}
\begin{minipage}[c]{0.3\linewidth}
    \flushright
    	\includegraphics[width=\linewidth]{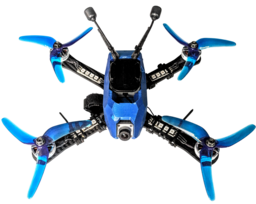}
\end{minipage}
\end{table}
In the final evaluations, we employ \rev{three} distinct experimental settings: the \emph{constant velocity environment}, the \emph{tracking environment}, \rev{and the motion-planning maze environment}, which are described below. In \rev{these experiments}, the obstacles can be smooth ellipses or nonsmooth cuboids.

\textbf{Constant Velocity Environment. }
In this simulation environment, we evaluate the control performance, including eight randomly moving dynamic obstacles, when aiming to fly at a constant speed of \unit[15]{$\frac{\mathrm{m}}{\mathrm{s}}$} at a specific height. 
The environment evaluates how early adaptation towards long-horizon prediction and short-horizon adaptation due to moving obstacles affect the closed-loop performance of maintaining a reference speed.
More details on the environment are given in the supplementary material.

\textbf{Reference Tracking Environment. }
In contrast to the previous environment, where the path is of minor importance, the reference-tracking environment requires the drone to track periodic trajectories that are occluded by varying numbers of static objects, with obstacle counts ranging from 3 to 34. The trajectory resembles shapes such as a triangle, a rounded rectangle, a figure-eight, a butterfly, or a sinusoidal wave. 
The tracking environment cost emphasizes the position tracking of the reference and, due to its periodic flight structure, can be readily evaluated in real-world experiments.
More details on the environment are given in the supplementary material.

\reb[4.8]{
\textbf{Motion Planning Maze Environment. }
The motion planning environments require the consideration of major nonconvexities that cannot be handled by smoothing alone. The goal is to traverse a $6\times6$~m tunnel at a target speed of $1~\frac{\mathrm{m}}{\mathrm{s}}$ with stage costs defined by~$x_k^\top Q x_k$ and $u_k^\top R u_k$. The tunnel has a length of $17$~m for real-world experiments and $40$~m for simulated experiments. We place up to $14$ random axis-aligned walls, see Fig.~\ref{fig:maze_top_view}, thereby forming a maze-like configuration space.
}

All environments are evaluated using the quadratic closed-loop cost with the diagonal matrix weight~$Q^\mathrm{eval}\in\R^{n_x\times n_x}$ and weights~$R^\mathrm{eval}\in\R^{n_u\times n_u}$ to promote smooth flight. \rev{For the fixed-duration comparisons,} the total simulation time is $t_\mathrm{sim}=\unit[24]{s}$ with a sampling time of $t_{\Delta,\mathrm{sim}}=\unit[40]{ms}$. We evaluate the closed-loop cost for $N_\mathrm{sim}$~steps with the simulated states~$\hat{X}=[\hat{x}_0,\ldots,\hat{x}_{N_\mathrm{sim}-1}]$ and controls~$\hat{U}=[\hat{u}_0,\ldots,\hat{u}_{N_\mathrm{sim}-1}]$ via
\begin{align}
\begin{split}
J&^\mathrm{eval}(\hat{X},\hat{U}):=\\
&t_{\Delta,\mathrm{sim}}\sum_{k=0}^{N_\mathrm{sim}-1}\rev{\left(\|\hat{x}_k-\tilde{x}_k\|_{2,Q^\mathrm{eval}} +
\|\hat{u}_k-\tilde{u}_k\|_{2,R^\mathrm{eval}}\right)}.
\end{split}
\end{align}
For the benchmark comparison, we normalize the accumulated cost via $\bar{J}^\mathrm{eval}(\hat{X},\hat{U})=J^\mathrm{eval}(\hat{X},\hat{U})/ J^\mathrm{eval}(\hat{X}^{\mathrm{UNIQUE}},\hat{U}^{\mathrm{UNIQUE}})$ by the cost from \textsc{Unique}, and similarly we normalize the computation time~$t_\mathrm{comp}$ taken by the MPC solver.

\section{\reb{Real-World Experiments}}
\label{sec:evaluation}
We evaluate and ablate our approach by comparing it against two benchmarks: the \emph{hierarchical approach}~\cite{hehn_2015,romero_2022} and the high-fidelity \emph{standard MPC} formulation, which uses a longer horizon without utilizing the point-mass model. 
The hierarchical formulation first plans a trajectory using only a point-mass model, formulating it as an MPC problem, and then employs a high-fidelity MPC tracking controller. The planner updates the trajectory every tenth lower-level tracking iteration. Remarkably, all parameters of the point-mass planning MPC, the tracking MPC, and the extended-horizon MPC are set to the corresponding phases of \textsc{Unique} for a fair comparison, except for the position-tracking weight, which is adapted for the hierarchical lower-level tracking MPC. 

The comparison in Sect.~\ref{sec:eval_realworld} involves real-world experiments in the reference-tracking environment. We analyze performance on four different tracks with varying numbers of obstacles. Additionally, we evaluate progressive smoothing for cubic obstacles and parallel MPCs within the proposed initialization strategy to avoid local minima. \reb[4.8]{In the first comparison, we consider a use case with many obstacles and few local minima, so random initial guesses can escape them reliably.}

\reb[4.8]{In Sect.~\ref{sec:eval_realworld_plan}, we compare the three variants in the maze environment with randomly placed horizontal and vertical walls.}

\subsection{\reb{Real-World Tracking Comparison}}
\label{sec:eval_realworld}
\begin{figure}
	\centering
	\includegraphics[width=\linewidth, trim=0cm 0.0cm 0cm 0cm, clip]{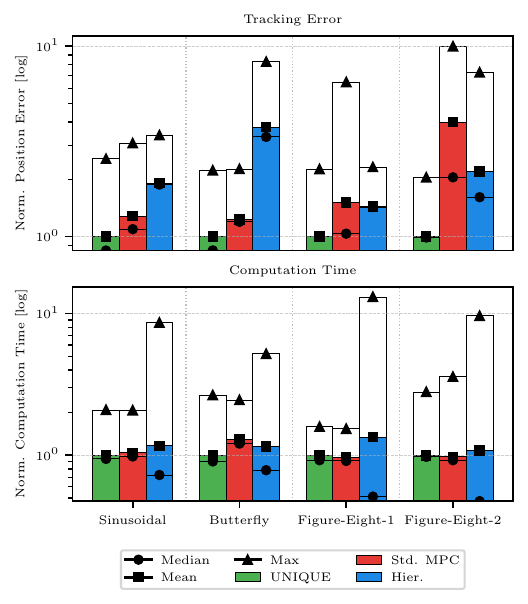}
	\caption{Comparison between \textsc{Unique}, the standard MPC formulation, and the \rev{hierarchical} MPC formulation regarding the online computation time and the tracking error (position). The tracking error is significantly reduced by \textsc{Unique}, with online computation time either unchanged or slightly reduced.}
	\label{fig:pareto_sim_1}
\end{figure}

\begin{figure*}
	\centering
	\includegraphics[width=\linewidth, trim=0.4cm 0cm 0.5cm 0.5cm, clip]{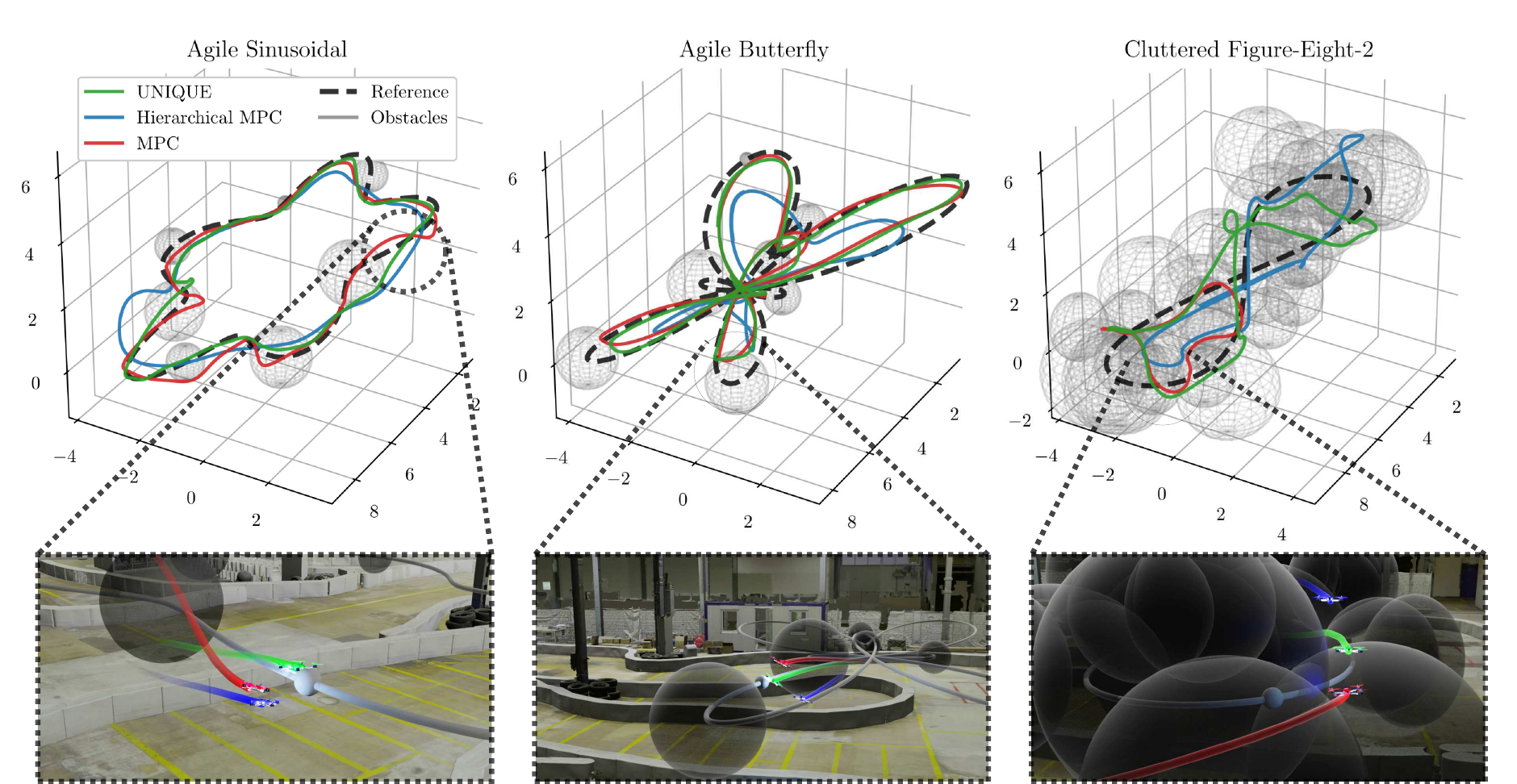}
	\caption{Visualization and rendered images of trajectories from different controllers in real-world experiments in a flying arena. We compare the proposed \textsc{Unique} framework with the hierarchical and MPC settings. The top plots show agile flights with seven smooth obstacles, therefore, less susceptible to local minima. Yet, \textsc{Unique} still outperforms the two baselines due to the increased long-horizon prediction. Standard MPC struggles once an obstacle appears on the horizon, attempting to evade it with more costly maneuvers. The hierarchical MPC planner and controller setting is limited by the coarse path of the higher-level planner (point mass model), which is apparent in the agile flight of the Butterfly track. The lower plots illustrate the cluttered environment, which exhibits numerous local minima. The standard MPC is not even capable of following the reference coarsely, as it gets stuck in the Figure-Eight-2 track on the left. The hierarchical framework escapes some local minima due to the long-horizon planning, but still exhibits a large tracking error. Only the \textsc{Unique} framework is able to track the cluttered Figure-Eight tracks due to its reinitialization strategy, which utilizes parallel and long-horizon planning. The axes are labeled in SI units.}
	\label{fig:tracks_rwe}
\end{figure*}

The real-world flights are performed across three representative trajectory shapes, with varying numbers of obstacles and different track periods, resulting in both agile and slower flight speeds. 
The evaluations involve (i) the flat $\unit[15\times 8]{\mathrm{m}}$ Agile Sinusoidal track with seven obstacles, a period of~$\unit[10]{\mathrm{s}}$ and accelerations of $\unit[20]{\frac{\mathrm{m}}{\mathrm{s}^2}}$, (ii) the flat $\unit[15\times 8]{\mathrm{m}}$ Agile Butterfly track with seven obstacles and accelerations of $\unit[35]{\frac{\mathrm{m}}{\mathrm{s}^2}}$, (iii) the flat $\unit[15\times 8]{\mathrm{m}}$ Cluttered Figure-Eight-Track-1 with $30$ obstacles and a period of $\unit[40]{\mathrm{s}}$ and (iv) a second similar Cluttered Figure-Eight-2 track with $34$ obstacles.
The evaluations shown in Tab.~\ref{tab:mpc-comparison} and Fig.~\ref{fig:pareto_sim_1} demonstrate that \textsc{Unique} consistently achieves a lower tracking error while requiring comparable or even reduced online computation time. 
The hierarchical approach increases the tracking error by up to $274.4\%$ and standard MPC by up to~$301.2\%$, corresponding to a reduction of the evaluation cost by $75\%$ using \textsc{Unique}.
The significant increase in tracking error of the standard MPC formulation in cluttered environments stems from the abundance of local minima, which necessitate long-horizon planning.
The hierarchical approach particularly struggles in the most agile Butterfly scenario, as the point-mass model can only approximate agile flight due to its conservative bounds.
\begin{table}[t]
	\centering
	\small
	\caption{\rev{Tracking Error} in Real-World Experiments.}
	\label{tab:mpc-comparison}
	\resizebox{\linewidth}{!}{
	\begin{tabular}{lll}
		\toprule
		Method & c.time (ms) & tracking error (m) \\
		\midrule
		& mean/med/max & mean/med/max \\
		\midrule
        \multicolumn{3}{c}{Sinusoidal} \\
UNIQUE        & 2.12 / 2.00 / 4.43  &\textbf{0.32} / \textbf{0.27} / \textbf{0.82} \\
Standard      & 2.21 / 2.08 / 4.40  &{\color{darkred}0.41} / 0.35 / 0.99 ({\color{darkred}+28.1\%}) \\
Hierarchical  & 2.47 / 1.54 / 18.27 &{\color{darkred}0.61} / 0.60 / 1.09 ({\color{darkred}+90.6\%}) \\
\midrule
\multicolumn{3}{c}{Butterfly} \\
UNIQUE        & 1.91 / 1.73 / 5.07  &\textbf{0.39} / \textbf{0.33} / \textbf{0.87} \\
Standard      & 2.47 / 2.31 / 4.69  &{\color{darkred}0.48} / 0.47 / 0.88 ({\color{darkred}+23.1\%}) \\
Hierarchical  & 2.21 / 1.50 / 9.92  &{\color{darkred}1.46} / 1.30 / 3.23 ({\color{darkred}+274.4\%}) \\
\midrule
\multicolumn{3}{c}{Figure-Eight-1} \\
UNIQUE        & 7.02 / 6.48 / 11.16 &\textbf{0.79} / \textbf{0.79} / \textbf{1.79} \\
Standard      & 6.77 / 6.41 / 10.79 &{\color{darkred}1.20} / 0.82 / 5.14 ({\color{darkred}+51.9\%}) \\
Hierarchical  & 9.44 / 3.59 / 91.84 &{\color{darkred}1.14} / 1.13 / 1.84 ({\color{darkred}+44.3\%}) \\
\midrule
\multicolumn{3}{c}{Figure-Eight-2} \\
UNIQUE        & 7.83 / 6.98 / 21.90   &\textbf{0.83} / \textbf{0.82} / \textbf{1.71} \\
Standard      & 7.64 / 7.23 / 28.05   &{\color{darkred}3.33} / 1.71 / 8.32 ({\color{darkred}+301.2\%}) \\
Hierarchical  & 8.48 / 3.72 / 75.34   &{\color{darkred}1.83} / 1.34 / 6.08 ({\color{darkred}+120.5\%}) \\
		\bottomrule
	\end{tabular}
}
\end{table}
The trajectory visualizations in Fig.~\ref{fig:tracks_rwe} show that \textsc{Unique} tracks tighter paths with fewer deviations, especially in high-curvature segments where the hierarchical approach struggles. These results confirm that cascading models within a single optimization not only improve accuracy but also avoid the overhead of long-horizon high-fidelity MPC, leading to superior closed-loop performance in practice.

\begin{figure}
	\centering
	\includegraphics[width=\linewidth, trim=0cm 0cm 0cm 0cm, clip]{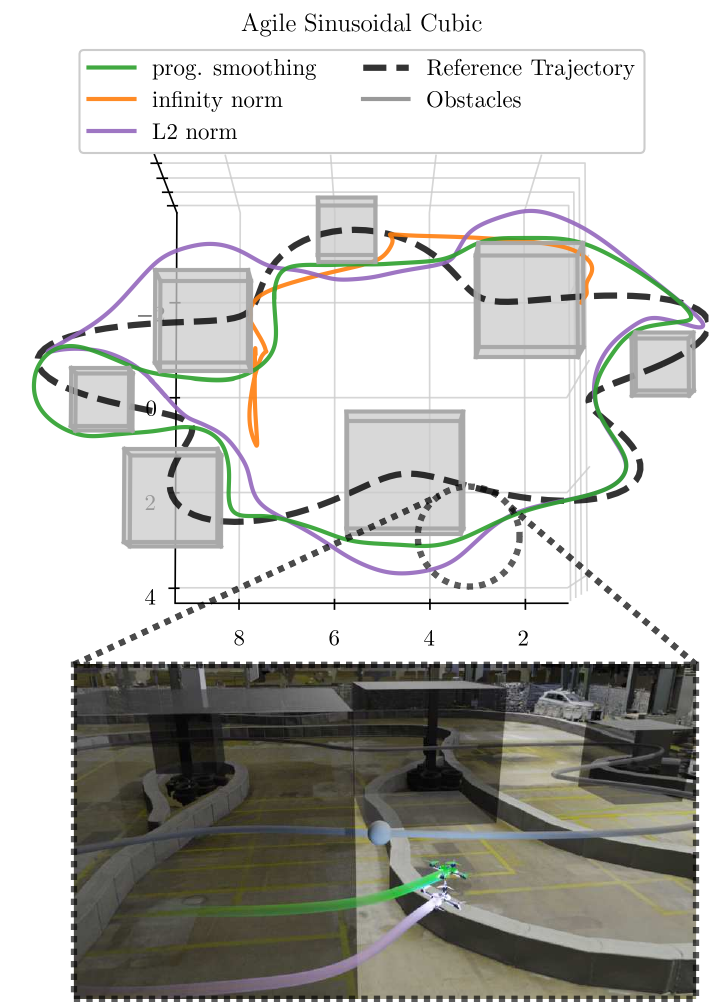}
	\caption{Real-world comparison of \textsc{Unique} with different formulations of cubic virtual obstacles. We compare the proposed progressive smoothing against a standard L2 norm and a tight infinity norm. The L2 norm is conservative, while the infinity norm leads to numerically unstable behavior. Progressive smoothing allows for a numerically stable, tight cubic obstacle representation. The lower rendering visualizes the L2 norm and progressive smoothing variant in a particular \rev{snapshot}.}
	\label{fig:prog_smooth_rwe}
\end{figure}

\textbf{Progressive Smoothing. } 
The influence of progressive smoothing and its significance for cubic obstacles is evaluated on the Sinusoidal track. We compare variants of \textsc{Unique} with fixed L2-norm, infinity norm, or progressive smoothing. The trajectories are shown in Fig.~\ref{fig:prog_smooth_rwe} and reveal a superior performance of \textsc{Unique} with progressive smoothing (mean/max tracking error of $\unit[0.6/1.4]{\mathrm{m}}$), compared to the conservative L2-norm approximation (mean/max tracking error of $\unit[0.8/2.0]{\mathrm{m}}$) and the numerically unstable but tight infinity norm. 
Remarkably, the computation time is not increased by progressive smoothing, since the norms and related parameters along the horizon are fixed.
More details on progressive smoothing with different schedules and the plotted MPC predictions are shown in the supplementary material.

\begin{figure}
	\centering
	\includegraphics[width=\linewidth, trim=0.2cm 0.3cm 0.2cm 0.2cm, clip]{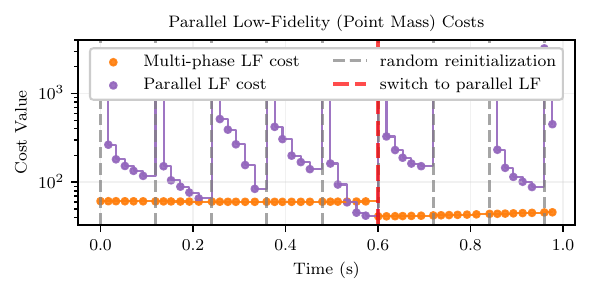}
	\caption{Comparison of the parallel point-mass MPC initialization approach. One parallel point-mass MPC is structurally identical to the second phase of the main multi-phase MPC and is randomly initialized every seventh step. Once the parallel MPC cost is lowest, the second phase of the main MPC is initialized with the corresponding lower-cost decision variables (at \unit[0.6]{s}). Due to the random initialization and RTI, the parallel MPC cost decreases significantly over the seven iterations, which are clearly visible as ``stairs''.}
	\label{fig:parallel_initializations}
\end{figure}

\textbf{Parallel Point-Mass MPC. }
We evaluate the influence of a parallel point-mass MPC in the Cluttered-Figure-Eight-2 track, where we activate or deactivate the parallel point-mass MPC. We perform 7 RTI steps on the parallel point-mass MPC before we randomly reinitialize. An active re-initialization and the related costs are plotted in Fig.~\ref{fig:parallel_initializations}.
Notably, the parallel point-mass MPC computation time is evaluated assuming parallel cores, i.e., the total \textsc{Unique} computation time is~$\max(t_\mathrm{comp,0},\ldots,t_\mathrm{comp,\rev{S}})$, where $t_\mathrm{comp,0}$ is the computation time of the multiphase MPC and $t_\mathrm{comp,1},\ldots,t_\mathrm{comp,\rev{S}}$ are the computation times of the parallel point-mass MPCs.
The mean and median computation times do not increase in our experiments compared to the multi-phase MPC without parallelization. However, the maximum computation time is increased from $\unit[10.9]{\mathrm{ms}}$ to $\unit[21.9]{\mathrm{ms}}$, which is still below the benchmark approaches compared, cf. Fig.~\ref{fig:tracks_rwe} and Tab.~\ref{tab:mpc-comparison}.
Considering the \rev{tracking error}, the difference when using parallel point-mass MPCs is remarkable in this obstacle-rich environment, as the mean \rev{tracking error} decreases by $77.1\%$ from $3.5/2.0/8.3$ (mean/median/max, \rev{in meters}) when using \textsc{Unique} without parallel roll-outs to $0.8/0.8/1.7$ when using the parallel roll-outs. This is due to the multi-phase MPC getting trapped in the local minima, similarly to the MPC in Fig.~\ref{fig:tracks_rwe}.
In the Agile Sinusoidal or Agile Butterfly tracks, the influence of the parallel point-mass MPCs was evaluated to be negligible due to the absence of dominant local minima. 

\subsection{\reb{Real-World Planning Comparison}}
\label{sec:eval_realworld_plan}
\begin{figure}
	\centering
	\includegraphics[width=\linewidth, trim=0.0cm 0.25cm 0.0cm 0.2cm, clip]{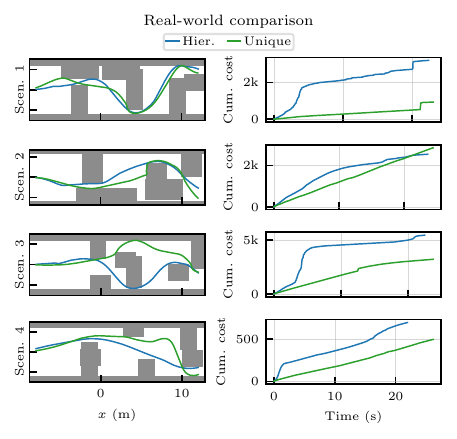}
	\caption{\reb[4.8]{Real-world evaluation showing trajectories and cumulative closed-loop costs on four randomly generated scenes for the hierarchical controller and \textsc{Unique}.}}
	\label{fig:rw_grid}
\end{figure}

\begin{figure}
	\centering
	\includegraphics[width=0.8\linewidth, trim=0.0cm 0.25cm 0.0cm 0.25cm, clip]{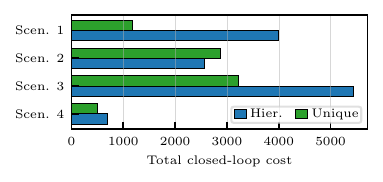}
	\caption{\reb[4.8]{Real-world evaluation on four randomly generated scenes. \textsc{Unique} achieves a $70.8\%$, $40.9\%$, and $28.7\%$ lower accumulated closed-loop cost than the hierarchical approach in three scenarios, and a $10.9\%$ higher cost in one scenario, \rev{corresponding to a $34.2\%$ reduction after aggregating the costs across all four scenarios}.}}
	\label{fig:rw_bars}
\end{figure}

\reb[4.8]{We evaluate the three navigation algorithms (std., hier., \textsc{Unique}) in a maze environment built from virtual obstacles in a flying arena equipped with a Vicon motion-capture system. As before, the hierarchical framework uses the same hyperparameters as the second horizon of \textsc{Unique} to provide a reference for a tracking MPC, whose hyperparameters match the first horizon of \textsc{Unique} except for a stronger position-tracking cost. We start at the position $p_0=[-9,0,1]$~m and aim to traverse the tunnel in the positive $x$ direction until $x_T\geq12$~m. Various obstacles block the tunnel, requiring planners with an 8-second horizon to find low-cost detours, as shown in Fig.~\ref{fig:rw_grid}. The standard MPC formulation is excluded from the comparison because it does not finish. Figure~\ref{fig:rw_grid} shows rapid cost increases for the hierarchical planner when a parallel point-mass planner finds a lower-cost trajectory that deviates from the previous plan and the tracking controller attempts to follow it. This replanning is much smoother when it is integrated into the second horizon of \textsc{Unique}. Scene~2 reveals a failure case for \textsc{Unique}: the parallel planners found a lower-cost trajectory around the blocking obstacle, but the slack variables were violated, preventing switching. As shown in Fig.~\ref{fig:rw_bars}, this increased the accumulated cost over the baseline. In the other scenarios, however, the sequential planning framework reduced the cost.}

\section{\reb{Simulated Experiments}}
\label{sec:simulated}
In the following simulation experiments, we ablate different hyperparameters in the constant-velocity \reb{and maze} environments to assess the robustness and scaling properties of \textsc{Unique}.
\reb[2.4,4.8]{First, we focus on control-oriented evaluations, emphasizing the closed-loop performance gains, feasibility, and computation time achieved via the multi-phase formulation. 
In Sect.~\ref{sec:eval_constraint}, we evaluate the different constraint approximations of the second horizon and provide some qualitative insights into progressive smoothing and the parallel initialization strategy.	
Secondly, we focus on the performance gain achieved via the parallel planning framework using the point-mass and MIQP planners.}
\reb[4.8]{Specifically, in Sect.~\ref{sec:planning_eval}, we evaluate how the computational burden scales with the number of obstacles in the mixed-integer planner, how different combinations of parallel solvers influence performance, and how \textsc{Unique} compares against the hierarchical formulation when both use equal planners.}
\subsection{\reb{Control-Focused Evaluation of Hyperparameters}}
\label{sec:eval_hyper}
The proposed \textsc{Unique} formulation comprises numerous hyperparameters, which we compare in simulated experiments within a constant velocity environment by evaluating their closed-loop cost. The most critical hyperparameters are the particular horizon lengths, number of shooting nodes of the short and long horizon, and the choice of the low-fidelity force constraint approximation~$\mathbb{S}^\mathrm{lf}_f$ or $\hat{\mathbb{S}}^\mathrm{lf}_f$ and the low-fidelity body rate constraint approximation~$\mathbb{S}^\mathrm{lf}_{\omega},\hat{\mathbb{S}}^\mathrm{lf}_{\omega}$ or $\widehat{\mathbb{S}}^\mathrm{lf}_{\omega}$. Additionally, the discretization time can be spaced by a factor~$c_\mathrm{space}$ for future horizons, for example by~$t_{\Delta,i+1}=c_\mathrm{space}t_{\Delta,i}$. The following section aims to answer crucial hyperparameter evaluation questions.

\begin{figure}
	\centering
	\includegraphics[width=\linewidth, trim=0cm 0.6cm 0cm 0cm, clip]{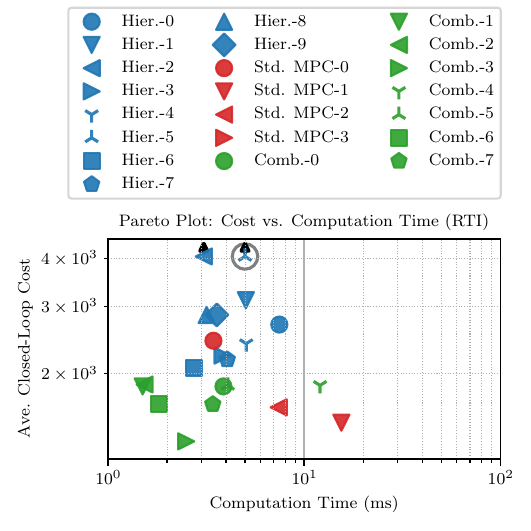}
	\includegraphics[width=\linewidth, trim=0cm 0cm 0cm 3.3cm, clip]{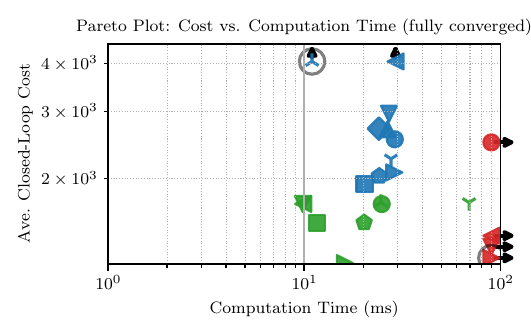}
	\caption{Pareto comparison of closed-loop cost and average per-iteration computation time in the constant-velocity environment. We compare different hyperparameter settings for the hierarchical, the standard MPC, and the unified formulation. The top plot shows a comparison for the real-time iteration scheme, while the bottom plot shows a comparison for the fully converged solver, which is nearly an order of magnitude slower. The numerically challenging long-horizon standard MPC formulations require comparably much higher computation times to fully converge than \textsc{Unique}, which is also Pareto optimal in both plots.}
	\label{fig:pareto_sim_2}
\end{figure}

\textit{\reb[2.4]{How do hyperparameters affect the closed-loop performance gains?}}
Fig.~\ref{fig:pareto_sim_2} shows a comparison of the Pareto front of closed-loop performance and online computation time for the \ac{RTI} scheme and the fully converged solvers, with parameters shown in Tab.~\ref{tab:single_summary_colored}.
The comparison includes a standard MPC with increasing spacing of the high-fidelity horizon, which constitutes an alternative method for planning more coarsely in the distant future. However, due to the nonlinearity of the high-fidelity model, the spacing leads to convergence problems.
The three different horizons of the standard MPC hint a barrier in the Pareto plot in Fig.~\ref{fig:pareto_sim_2} and also, the hierarchical approach clusters in a specific region. Both benchmarks are clearly outperformed by \textsc{Unique}.

\textit{How do hyperparameters affect convergence?}
Although the RTI scheme is the only applicable approach in real-world evaluations, the fully converged evaluation shown in Fig.~\ref{fig:pareto_sim_2} and Tab.~\ref{tab:single_summary_colored} contrasts the different approaches in terms of their numerical complexity. \rev{Across the reported configurations, full convergence increases computation time by factors of approximately 6--7 for \textsc{Unique} and 29--137 for standard MPC relative to RTI. Thus, \textsc{Unique} requires substantially less additional time for full convergence than the long-horizon standard MPC configurations.}
\begin{table*}
	\centering
	\caption{Parameter and evaluation of \textsc{Unique}, the hierarchical setting, and the standard MPC controller in the randomized constant velocity simulation experiment.}
	\resizebox{\linewidth}{!}{
		\begin{tabular}{lcccccccccc}
			\toprule
			Experiment & Hor. / $t_\Delta$ & Nodes & Spacing &Hor. / $t^\mathrm{lf}_\Delta$ & Nodes & Spacing & Interval & Constr. & Cost & Comp. Time \\
			Unit & (s) &  &  &(s) &  &  &  &  & Mean Val. & Mean Val (ms) \\
			\midrule
			& \multicolumn{3}{c|}{High-Fid. Part} & \multicolumn{5}{c|}{Low-Fidelity Part} &  \multicolumn{2}{c}{RTI / Converged}  \\
			\midrule
            \grayrow
			\tikz[baseline=-0.6ex] \node[circle, draw={rgb,1:red,0.1725; green,0.6275; blue,0.1725}, fill={rgb,1:red,0.1725; green,0.6275; blue,0.1725}, line width=0.8pt, inner sep=0pt, minimum size=8pt] {}; \textsc{Unique}-0 & 0.52 / 0.04 & 13 &  & 17.60 / 0.80 & 22 &0.10 &  & $\mathbb{S}_\mathrm{f}^\mathrm{lf}$ , $\hat{\mathbb{S}}_\omega^\mathrm{lf}$ & 1853.74 / 1717.60 & 3.88 / 24.85 \\
			\tikz[baseline=-0.6ex] \node[regular polygon, regular polygon sides=3, rotate=180, draw={rgb,1:red,0.1725; green,0.6275; blue,0.1725}, fill={rgb,1:red,0.1725; green,0.6275; blue,0.1725}, line width=0.8pt, inner sep=0pt, minimum size=8pt] {}; \textsc{Unique}-1 & 0.52 / 0.04 & 13 &  & 17.60 / 0.80 & 22 &0.10 &  & $\hat{\mathbb{S}}_\mathrm{f}^\mathrm{lf}$, $\widehat{\mathbb{S}}_\omega^\mathrm{lf}$ & 1853.74 / 1717.60 & \textbf{1.50} / \textbf{9.84} \\
            \grayrow
			\tikz[baseline=-0.6ex] \node[regular polygon, regular polygon sides=3, rotate=90, draw={rgb,1:red,0.1725; green,0.6275; blue,0.1725}, fill={rgb,1:red,0.1725; green,0.6275; blue,0.1725}, line width=0.8pt, inner sep=0pt, minimum size=8pt] {}; \textsc{Unique}-2 & 0.52 / 0.04 & 13 &  & 16.00 / 0.80 & 20 &0.20 &  & $\hat{\mathbb{S}}_\mathrm{f}^\mathrm{lf}$, $\widehat{\mathbb{S}}_\omega^\mathrm{lf}$ & 1875.22 / 1720.68 & 1.54 / 9.88 \\
			\tikz[baseline=-0.6ex] \node[regular polygon, regular polygon sides=3, rotate=-90, draw={rgb,1:red,0.1725; green,0.6275; blue,0.1725}, fill={rgb,1:red,0.1725; green,0.6275; blue,0.1725}, line width=0.8pt, inner sep=0pt, minimum size=8pt] {}; \textsc{Unique}-3 & 0.80 / 0.04 & 20 &  & 7.20 / 0.80 & 9 &0.00 &  & $\mathbb{S}_\mathrm{f}^\mathrm{lf}$ , $\hat{\mathbb{S}}_\omega^\mathrm{lf}$ & \textbf{1331.95} / \textbf{1203.88} & 2.50 / 16.15 \\
            \grayrow
			\tikz[baseline=-0.6ex] \node[regular polygon, regular polygon sides=3, rotate=180, draw={rgb,1:red,0.1725; green,0.6275; blue,0.1725}, fill={rgb,1:red,0.1725; green,0.6275; blue,0.1725}, line width=0.8pt, inner sep=0pt, minimum size=6.800pt] {}; \textsc{Unique}-4 & 0.80 / 0.04 & 20 &  & 15.20 / 0.20 & 76 &0.00 &  & $\mathbb{S}_\mathrm{f}^\mathrm{lf}$ , $\hat{\mathbb{S}}_\omega^\mathrm{lf}$ & 1865.09 / 1734.07 & 12.05 / 69.26 \\
			\tikz[baseline=-0.6ex] \node[regular polygon, regular polygon sides=3, rotate=0, draw={rgb,1:red,0.1725; green,0.6275; blue,0.1725}, fill={rgb,1:red,0.1725; green,0.6275; blue,0.1725}, line width=0.8pt, inner sep=0pt, minimum size=6.800pt] {}; \textsc{Unique}-5 & 0.80 / 0.04 & 20 &  & 15.20 / 0.20 & 76 &0.00 &  & $\hat{\mathbb{S}}_\mathrm{f}^\mathrm{lf}$, $\widehat{\mathbb{S}}_\omega^\mathrm{lf}$ & 1865.12 / 1734.09 & 4.08 / 24.59 \\
            \grayrow
			\tikz[baseline=-0.6ex] \node[rectangle, draw={rgb,1:red,0.1725; green,0.6275; blue,0.1725}, fill={rgb,1:red,0.1725; green,0.6275; blue,0.1725}, line width=0.8pt, inner sep=0pt, minimum width=8pt, minimum height=8pt] {}; \textsc{Unique}-6 & 0.80 / 0.04 & 20 &  & 3.20 / 0.20 & 16 &0.00 &  & $\hat{\mathbb{S}}_\mathrm{f}^\mathrm{lf}$, $\widehat{\mathbb{S}}_\omega^\mathrm{lf}$ & 1664.14 / 1535.86 & 1.81 / 11.64 \\
			\tikz[baseline=-0.6ex] \node[regular polygon, regular polygon sides=5, draw={rgb,1:red,0.1725; green,0.6275; blue,0.1725}, fill={rgb,1:red,0.1725; green,0.6275; blue,0.1725}, line width=0.8pt, inner sep=0pt, minimum size=8pt] {}; \textsc{Unique}-7 & 0.80 / 0.04 & 20 &  & 3.20 / 0.20 & 16 &0.00 &  & $\mathbb{S}_\mathrm{f}^\mathrm{lf}$ , $\hat{\mathbb{S}}_\omega^\mathrm{lf}$ & 1665.23 / 1536.51 & 3.42 / 20.21 \\
			\midrule
			& \multicolumn{3}{c|}{Controller} & \multicolumn{5}{c|}{Planner} &  \\
			\midrule
            \grayrow
			\tikz[baseline=-0.6ex] \node[circle, draw={rgb,1:red,0.1216; green,0.4667; blue,0.7059}, fill={rgb,1:red,0.1216; green,0.4667; blue,0.7059}, line width=0.8pt, inner sep=0pt, minimum size=8pt] {}; Hier.-0 & 2.00 / 0.04 & 50 &  & 24.00 / 0.20 & 120 & & 10 & $\mathbb{S}_\mathrm{f}^\mathrm{lf}$ , $\hat{\mathbb{S}}_\omega^\mathrm{lf}$ & 2688.66 / 2535.47 & 7.47 / 28.98 \\
			\tikz[baseline=-0.6ex] \node[regular polygon, regular polygon sides=3, rotate=180, draw={rgb,1:red,0.1216; green,0.4667; blue,0.7059}, fill={rgb,1:red,0.1216; green,0.4667; blue,0.7059}, line width=0.8pt, inner sep=0pt, minimum size=8pt] {}; Hier.-1 & 2.00 / 0.04 & 50 &  & 24.00 / 0.40 & 60 & & 10 & $\mathbb{S}_\mathrm{f}^\mathrm{lf}$ , $\hat{\mathbb{S}}_\omega^\mathrm{lf}$ & 3112.53 / 2957.01 & 5.04 / 27.06 \\
            \grayrow
			\tikz[baseline=-0.6ex] \node[regular polygon, regular polygon sides=3, rotate=90, draw={rgb,1:red,0.1216; green,0.4667; blue,0.7059}, fill={rgb,1:red,0.1216; green,0.4667; blue,0.7059}, line width=0.8pt, inner sep=0pt, minimum size=8pt] {}; Hier.-2 & 2.00 / 0.04 & 50 &  & 8.00 / 0.20 & 40 & & 10 & $\hat{\mathbb{S}}_\mathrm{f}^\mathrm{lf}$, $\widehat{\mathbb{S}}_\omega^\mathrm{lf}$ & 7170.31 / 7019.67 & 3.08 / 29.21 \\
			\tikz[baseline=-0.6ex] \node[regular polygon, regular polygon sides=3, rotate=-90, draw={rgb,1:red,0.1216; green,0.4667; blue,0.7059}, fill={rgb,1:red,0.1216; green,0.4667; blue,0.7059}, line width=0.8pt, inner sep=0pt, minimum size=8pt] {}; Hier.-3 & 2.00 / 0.04 & 50 &  & 8.00 / 0.20 & 40 & & 10 & $\mathbb{S}_\mathrm{f}^\mathrm{lf}$ , $\hat{\mathbb{S}}_\omega^\mathrm{lf}$ & 2231.18 / 2077.67 & 3.84 / 28.81 \\
            \grayrow
			\tikz[baseline=-0.6ex] \node[regular polygon, regular polygon sides=3, rotate=180, draw={rgb,1:red,0.1216; green,0.4667; blue,0.7059}, fill={rgb,1:red,0.1216; green,0.4667; blue,0.7059}, line width=0.8pt, inner sep=0pt, minimum size=6.800pt] {}; Hier.-4 & 2.00 / 0.04 & 50 &  & 8.00 / 0.20 & 40 & & 4 & $\mathbb{S}_\mathrm{f}^\mathrm{lf}$ , $\hat{\mathbb{S}}_\omega^\mathrm{lf}$ & 2397.97 / 2255.47 & 5.07 / 27.72 \\
			\tikz[baseline=-0.6ex] \node[regular polygon, regular polygon sides=3, rotate=0, draw={rgb,1:red,0.1216; green,0.4667; blue,0.7059}, fill={rgb,1:red,0.1216; green,0.4667; blue,0.7059}, line width=0.8pt, inner sep=0pt, minimum size=6.800pt] {}; Hier.-5 & 0.40 / 0.04 & 10 &  & 8.00 / 0.20 & 40 & & 4 & $\mathbb{S}_\mathrm{f}^\mathrm{lf}$ , $\hat{\mathbb{S}}_\omega^\mathrm{lf}$ & 22189.62 / 30800.65 & 4.99 / 10.97 \\
            \grayrow
			\tikz[baseline=-0.6ex] \node[rectangle, draw={rgb,1:red,0.1216; green,0.4667; blue,0.7059}, fill={rgb,1:red,0.1216; green,0.4667; blue,0.7059}, line width=0.8pt, inner sep=0pt, minimum width=8pt, minimum height=8pt] {}; Hier.-6 & 1.20 / 0.04 & 30 &  & 8.00 / 0.20 & 40 & & 10 & $\mathbb{S}_\mathrm{f}^\mathrm{lf}$ , $\hat{\mathbb{S}}_\omega^\mathrm{lf}$ & 2072.90 / 1934.08 & 2.73 / 20.31 \\
			\tikz[baseline=-0.6ex] \node[regular polygon, regular polygon sides=5, draw={rgb,1:red,0.1216; green,0.4667; blue,0.7059}, fill={rgb,1:red,0.1216; green,0.4667; blue,0.7059}, line width=0.8pt, inner sep=0pt, minimum size=8pt] {}; Hier.-7 & 1.20 / 0.04 & 30 &  & 8.00 / 0.20 & 40 & & 4 & $\mathbb{S}_\mathrm{f}^\mathrm{lf}$ , $\hat{\mathbb{S}}_\omega^\mathrm{lf}$ & 2178.72 / 2033.14 & 4.06 / 24.13 \\
            \grayrow
			\tikz[baseline=-0.6ex] \node[regular polygon, regular polygon sides=3, rotate=0, draw={rgb,1:red,0.1216; green,0.4667; blue,0.7059}, fill={rgb,1:red,0.1216; green,0.4667; blue,0.7059}, line width=0.8pt, inner sep=0pt, minimum size=8pt] {}; Hier.-8 & 2.00 / 0.04 & 50 &  & 4.00 / 0.20 & 20 & & 10 & $\mathbb{S}_\mathrm{f}^\mathrm{lf}$ , $\hat{\mathbb{S}}_\omega^\mathrm{lf}$ & 2848.82 / 2695.63 & 3.18 / 26.60 \\
			\tikz[baseline=-0.6ex] \node[diamond, draw={rgb,1:red,0.1216; green,0.4667; blue,0.7059}, fill={rgb,1:red,0.1216; green,0.4667; blue,0.7059}, line width=0.8pt, inner sep=0pt, minimum size=8pt] {}; Hier.-9 & 2.00 / 0.04 & 50 &  & 4.00 / 0.20 & 20 & & 4 & $\mathbb{S}_\mathrm{f}^\mathrm{lf}$ , $\hat{\mathbb{S}}_\omega^\mathrm{lf}$ & 2848.93 / 2703.57 & 3.59 / 24.04 \\
			\midrule
			& \multicolumn{3}{c|}{Controller} & \multicolumn{5}{c|}{} &  &  \\
			\midrule
            \grayrow
			\tikz[baseline=-0.6ex] \node[circle, draw={rgb,1:red,0.8392; green,0.1529; blue,0.1569}, fill={rgb,1:red,0.8392; green,0.1529; blue,0.1569}, line width=0.8pt, inner sep=0pt, minimum size=8pt] {}; Std. MPC-0 & 2.00 / 0.04 & 50 & 0.00 &  &  & &  &  & 2439.76 / 2489.68 & 3.45 / 474.95 \\
			\tikz[baseline=-0.6ex] \node[regular polygon, regular polygon sides=3, rotate=180, draw={rgb,1:red,0.8392; green,0.1529; blue,0.1569}, fill={rgb,1:red,0.8392; green,0.1529; blue,0.1569}, line width=0.8pt, inner sep=0pt, minimum size=8pt] {}; Std. MPC-1 & 8.00 / 0.04 & 200 & 0.00 &  &  & &  &  & 1490.52 / 1328.99 & 15.47 / 2102.14 \\
            \grayrow
			\tikz[baseline=-0.6ex] \node[regular polygon, regular polygon sides=3, rotate=90, draw={rgb,1:red,0.8392; green,0.1529; blue,0.1569}, fill={rgb,1:red,0.8392; green,0.1529; blue,0.1569}, line width=0.8pt, inner sep=0pt, minimum size=8pt] {}; Std. MPC-2 & 4.00 / 0.04 & 100 & 0.00 &  &  & &  &  & 1634.34 / 1420.70 & 7.41 / 1013.03 \\
			\tikz[baseline=-0.6ex] \node[regular polygon, regular polygon sides=3, rotate=-90, draw={rgb,1:red,0.8392; green,0.1529; blue,0.1569}, fill={rgb,1:red,0.8392; green,0.1529; blue,0.1569}, line width=0.8pt, inner sep=0pt, minimum size=8pt] {}; Std. MPC-3 & 7.92 / 0.04 & 55 & 0.10 &  &  & &  &  & \rev{n.c.} / 1242.62 & 5.23 / 152.10 \\
			\bottomrule
		\end{tabular}
	}
	\label{tab:single_summary_colored}
\end{table*}

\begin{figure}
	\centering
	\includegraphics[width=\linewidth, trim=0cm 0.95cm 0cm 0cm, clip]{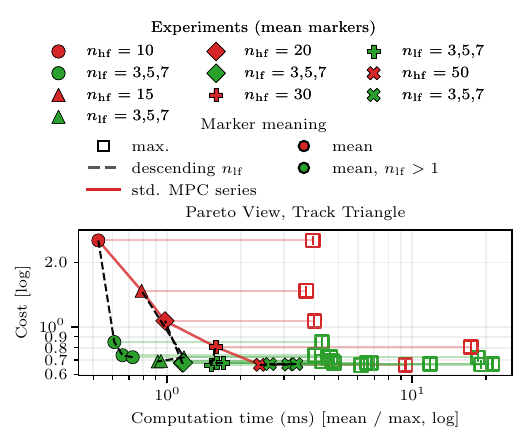}
	\includegraphics[width=\linewidth, trim=0cm 0.95cm 0cm 3.4cm, clip]{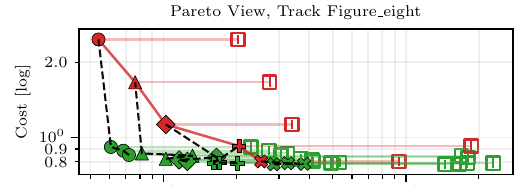}
	\includegraphics[width=\linewidth, trim=0cm 0.95cm 0cm 3.4cm, clip]{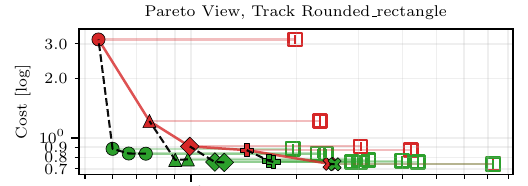}
	\includegraphics[width=\linewidth, trim=0cm 0cm 0cm 3.4cm, clip]{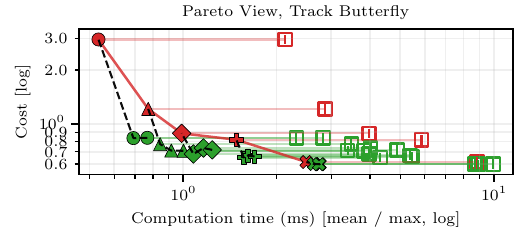}
	\caption{Pareto comparison between increasing the horizon of the standard MPC formulation (red line), as opposed to increasing the horizon with the planner formulation (green markers and dashed gray line) on different test tracks. Remarkably, adding the second low-fidelity horizon outperforms MPC with increased horizons.}
	\label{fig:tracking_results}
\end{figure}

\textit{How does the second phase influence \reb{control} performance?}
In Fig.~\ref{fig:tracking_results}, we investigate how adding steps of the second phase influences the Pareto performance in comparison to how adding standard MPC steps influences the performance. We take standard MPC with~$n_\mathrm{hf}=\{10,15,20,30,50\}$ shooting nodes and $t_{\Delta}^\mathrm{hf}=\unit[40]{\mathrm{ms}}$ (red line in Fig.~\ref{fig:tracking_results}) and append to each standard high-fidelity MPC \rev{the} low-fidelity part of~$n_\mathrm{lf}=\{3,5,7\}$ with $t_{\Delta}^\mathrm{lf}=\unit[400]{\mathrm{ms}}$ (black dotted line).
The reference tracking involves three obstacles and shapes of a Triangle, a rounded Rectangle, a Figure Eight, and a Butterfly. Additionally, we add a small sinusoidal z-component.
The evaluation on the Pareto front reveals that it is more efficient to add the low-fidelity horizon of \textsc{Unique} than to increase the MPC horizon. \reb[4.9]{This confirms that, even from a pure control perspective, the two-phase structure yields superior closed-loop performance under the same computational budget.}

\subsection{Evaluation of Constraint Approximation}
\label{sec:eval_constraint}

\begin{figure*}
	\centering
	\includegraphics[width=\textwidth, trim=0cm 0cm 0cm 0cm, clip]{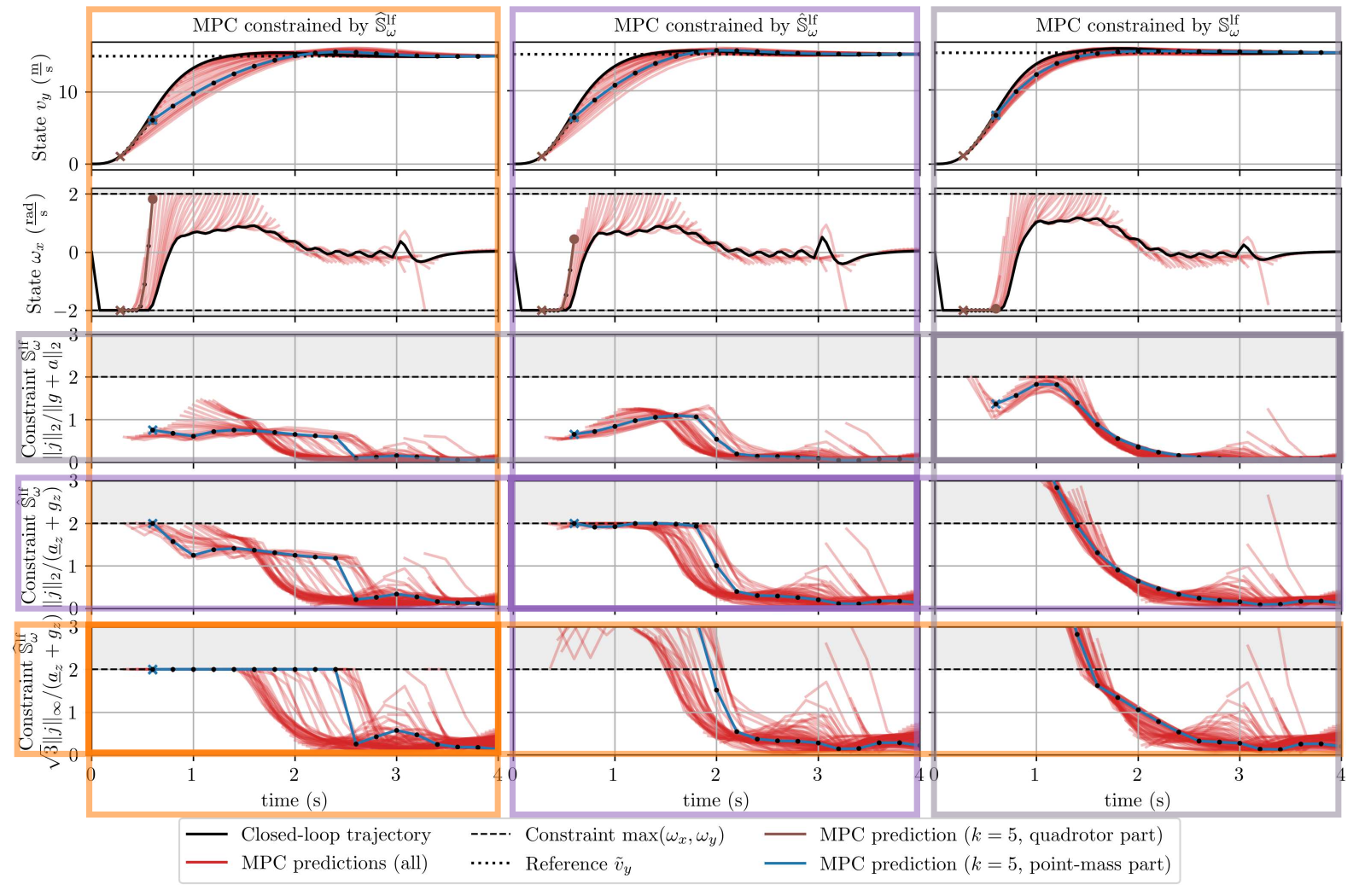}
	\caption{Visualization of different safe set approximations for body rates (three columns: \colorbox{taborange}{box constraints $\widehat{\mathbb{S}}_\omega^{\mathrm{lf}}$}, \colorbox{tabpurple}{polyhedral $\hat{\mathbb{S}}_\omega^{\mathrm{lf}}$}, \colorbox{tabgrey}{nonlinear $\mathbb{S}_\omega^{\mathrm{lf}}$}) for a step response to a horizontal set speed ($14\frac{\mathrm{m}}{\mathrm{s}}$, first row). The second row shows the body rates and the respective constraints for the high-fidelity model. The lower three rows show the different constraint function evaluations on the lower-fidelity part, with the lower white region being feasible. The third row shows the most accurate \colorbox{tabgrey}{nonlinear constraints}. It reveals that these are satisfied for all formulations, however, more conservatively, for the \colorbox{taborange}{box constraints} and the \colorbox{tabpurple}{polyhedral constraints}. The lower off-diagonal plots show the respective active constraint evaluation.}
	\label{fig:constraint_sets}
\end{figure*}

In this Section, we evaluate the influence of the different constraint formulations on the low-fidelity model phase. Particularly, we assess the rise time, i.e., the time it takes to reach $90\%$ of the set velocity of $\unit[14]{\frac{\mathrm{m}}{\mathrm{s}}}$ from hovering using the different
low-fidelity force constraint approximation~$\mathbb{S}^\mathrm{lf}_f$ (polyhedral) or $\hat{\mathbb{S}}^\mathrm{lf}_f$ (box) and the low-fidelity body rate constraint approximation {$\mathbb{S}^\mathrm{lf}_{\omega}$ (nonlinear)}, {$\hat{\mathbb{S}}^\mathrm{lf}_{\omega}$ (polyhedral)} or {$\widehat{\mathbb{S}}^\mathrm{lf}_{\omega}$ (box)}\footnote{We use a color coding for body rate constraint types to simplify the reading\rev{.}}.
Fig.~\ref{fig:constraint_sets} shows the corresponding velocity state~$v_y$, the related body rate~$\omega_x$, and the evaluation of the three different constraint functions that approximate the body rates. In addition, all MPC predictions of both phases are plotted, as well as one particular prediction at step $k=5$ equal to $\unit[0.2]{\mathrm{s}}$. As anticipated, the more conservative constraints lead to an increased rise time. Conservatism can be verified in the lower three rows, which show their constraint function evaluations. Yet, despite the visible larger conservatism in the second horizon, the re-optimization along the first high-fidelity horizon can improve the closed-loop performance. 

\begin{figure}
	\centering
	\includegraphics[width=\linewidth, trim=0cm 2cm 0cm 10cm, clip]{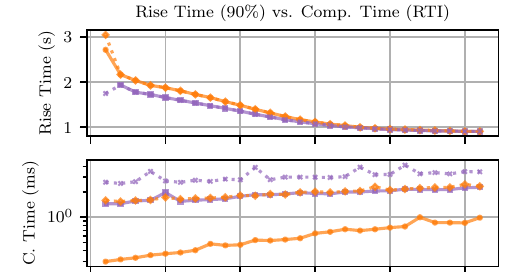}
	\begin{tikzpicture}
		\node[inner sep=0] (stepconv) {\includegraphics[width=\linewidth, trim=0cm 0cm 0cm 9.95cm, clip]{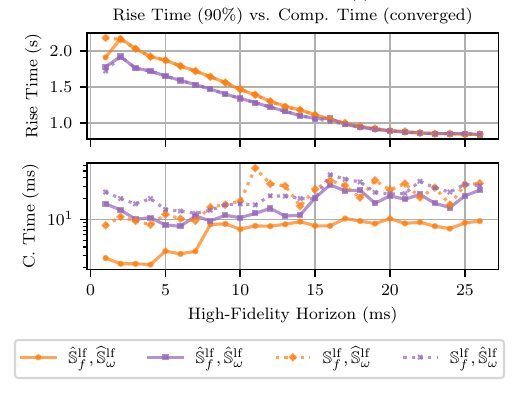}};
		\node[fill=white,inner xsep=12pt,inner ysep=2pt,font=\scriptsize] at ([yshift=1.36cm]stepconv.south) {High-Fidelity Horizon (shooting nodes)};
	\end{tikzpicture}
	\caption{Comparison of the rise time (90\% of the set value) and computation time for a horizontal velocity step regarding different high-fidelity horizons using a large low-fidelity horizon. \rev{The horizontal axis reports the number of high-fidelity shooting nodes.} The experiments reveal the diminishing influence of the different safe set formulations on the closed-loop performance when the first high-fidelity model horizon is increased. More accurate safe set formulations require longer computation times but improve performance, particularly for the shorter high-fidelity model phase.}
	\label{fig:step}
\end{figure}

Trivially, if the first horizon is increased, the second horizon and the related constraints become less relevant. We evaluated this dependency in Fig.~\ref{fig:step}, where we increased the first high-fidelity horizon and evaluated the rise time and computation time for RTIs and the fully converged solution. If the first horizon is increased \rev{beyond 20 shooting nodes ($\unit[0.8]{s}$)}, the constraint formulation on the second horizon becomes irrelevant for the particular maneuver.
Both the converged and the RTI evaluations show an increasing difference in the constraint formulation for shorter horizons of the first phase. Using {box constraints on body rates with~$\rev{\widehat{\mathbb{S}}^\mathrm{lf}_{\omega}}$} and thrust~$\rev{\hat{\mathbb{S}}^\mathrm{lf}_{f}}$ shows \rev{the lowest computation time, particularly for RTI}.
The constraint formulation for the thrust shows no difference in the evaluation. 
\reb{\subsection{Planning evaluations in the maze environment}
\label{sec:planning_eval}
In the following, we evaluate the planning capabilities of \textsc{Unique} in the simulated randomized maze environment. We examine how the computation time of the potentially slow MIQP planner scales with the number of obstacles, how the parallel synchronization can be satisfied, and how the typical hierarchical decomposition compares in different configurations against \textsc{Unique}.}

\begin{figure*}
	\centering
	\includegraphics[width=\linewidth, trim=0cm 0cm 0cm 0cm, clip]{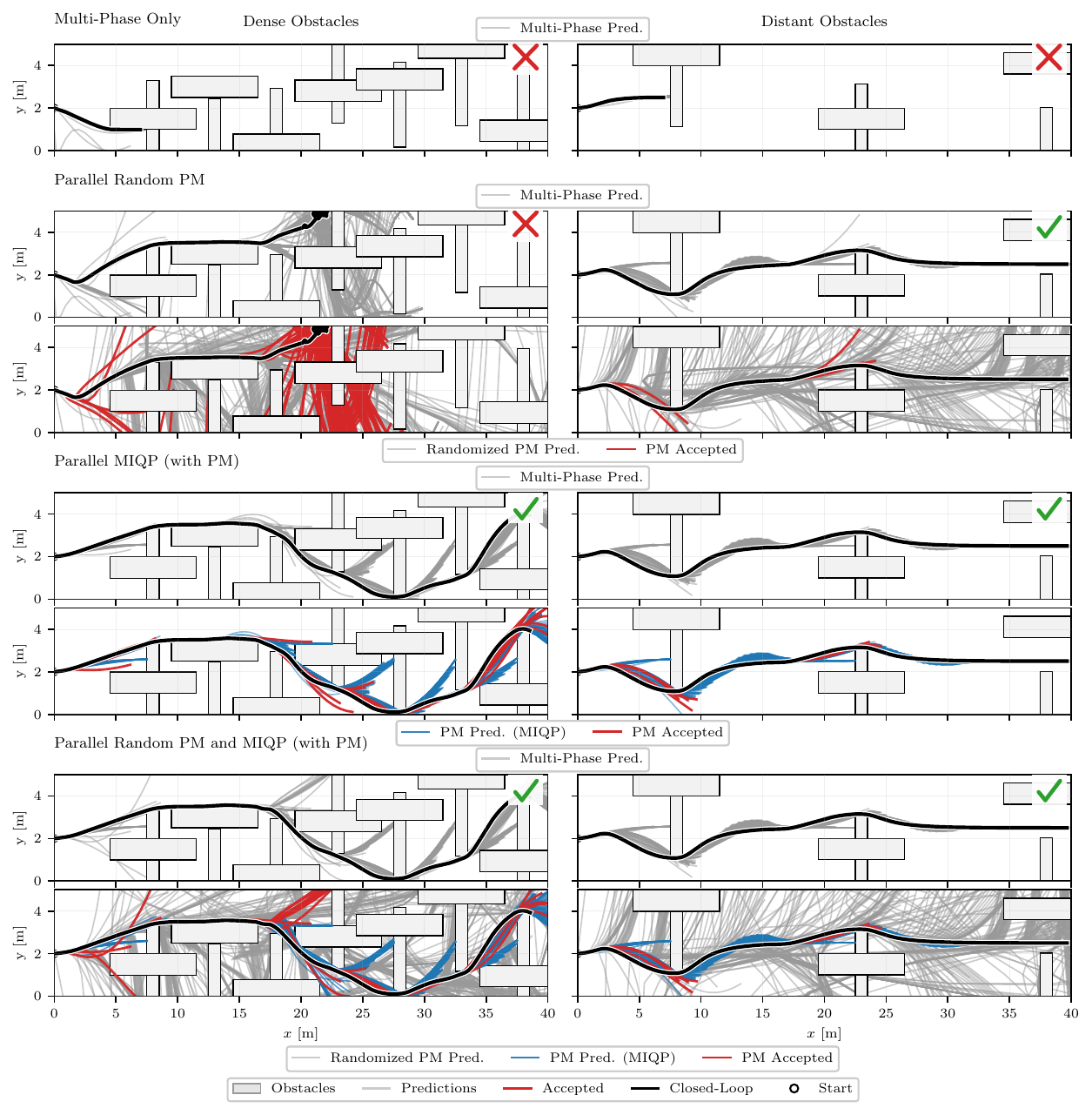}
	\caption{\reb[4.8]{Top view of simulation comparisons of \textsc{Unique} with different hyperparameters. The left column shows a denser maze and the right column a sparser maze. The first row shows that \textsc{Unique} without parallel planning gets easily stuck in local minima. The second row shows successful planning in the simpler right-hand scenario with parallel randomly initialized point-mass planners. The plots show how the free space is explored by alternating \acp{RTI} and reinitialization with initial guesses. The third row shows successful planning with the MIQP parallel planner, and the last row shows successful planning for all scenarios with additional point-mass planners.}}
	\label{fig:maze_top_view}
\end{figure*}

\begin{figure}
	\centering
	\includegraphics[width=\linewidth, trim=0.2cm 0.2cm 0cm 0.2cm, clip]{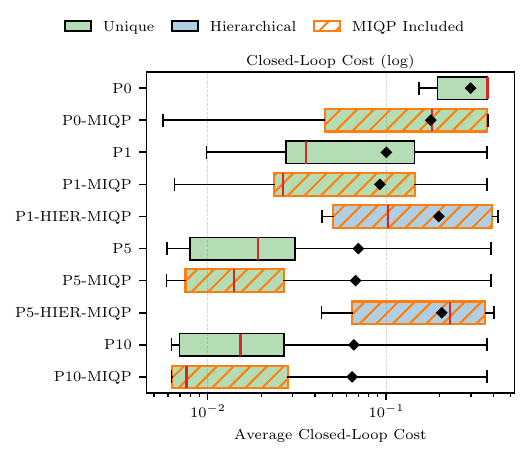}
	\caption{\reb[4.8]{Closed-loop cost comparison on randomized instances of the maze environment. Including the mixed-integer planner yields the lowest closed-loop cost, provided sufficient computational resources. The hierarchical planner/controller architecture has a significantly larger closed-loop cost, despite also being able to navigate the mazes toward the goal. 
	}}
	\label{fig:maze_progress}
\end{figure}

\reb[4.8]{
\textit{How do planning hyperparameters affect the closed-loop performance?}
The main planning hyperparameters are the use of the MIQP solver as an initializer and the number of parallel, randomly initialized point-mass solvers.
We vary the hyperparameters in the maze environment and compare the closed-loop cost, which is mainly the progress in the maze-tunnel, with a constant velocity.
Fig.~\ref{fig:maze_top_view} shows a top view of two maze environments (denser on the left and sparser on the right) for different configurations. The plots qualitatively show the benefit of the MIQP solver, which finds gaps more reliably than randomly initialized point-mass solvers. In the simpler, sparser environment, the parallel point-mass solvers can find an equally good solution. Figure~\ref{fig:maze_progress} shows a quantitative comparison across 40 randomized mazes and different configurations. Remarkably, with 10 parallel randomly initialized solvers, a solution of similar quality to that of the MIQP solver was found.}

\reb[4.8]{
\textit{How does \textsc{Unique} compare against the hierarchical configuration with the same hyperparameters?}
In Fig.~\ref{fig:maze_progress}, we evaluate, for some configurations, the hierarchical setting, i.e., a planner with equal parameters and parallel evaluations together with a tracking MPC controller. The figure shows that, in terms of closed-loop cost, \textsc{Unique} clearly outperforms the hierarchical baseline. Remarkably, if only progress is considered, the hierarchical configuration also achieves comparable performance, yet in a suboptimal way.
}

\reb[4.8]{
	\textit{Can real-time control be achieved?}
	In Fig.~\ref{fig:maze_timings} the computation times of the parallel point-mass solvers, the main multi-phase MPC, and the MIQP are compared statistically. The time-critical solvers finish in time, while the MIQP has to run asynchronously because it exceeds the time budget.
	\begin{figure}
		\centering
		\includegraphics[width=\linewidth, trim=0cm 0.25cm 0cm 0.2cm, clip]{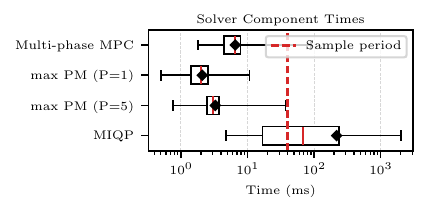}
		\caption{\reb[4.8]{Timing comparison of the different solvers in the maze environment. Remarkably, the parallel, randomly initialized point-mass solvers can be solved much faster than the main multi-phase MPC and can therefore be synchronized after each iteration. However, the MIQP solver must be executed asynchronously due to its longer computation time.}}
		\label{fig:maze_timings}
	\end{figure}
}

\reb[3.5]{
	\textit{Can the parallel evaluations finish on time?}
	Since our algorithm assumes the parallel point-mass solver completes after each iteration of the main multi-phase MPC, we evaluate the share of iterations in which the parallel solver finishes before the main multi-phase MPC in the maze environment over 200 iterations. We report a success rate of 99.5\%. Moreover, we evaluate the same metric for a reinitialized full multi-phase MPC with randomized initial guesses and find that only 12\% of the runs finish in time, highlighting the advantage of the considerably cheaper point-mass evaluations.}

\reb[4.8]{
	\textit{How does MIQP computation time scale with the number of obstacles?}
	Most critically, the MIQP planner uses binary indicator constraints for obstacle-free convex cells, the number of which grows with the number of obstacles. In principle, this scales an NP-hard problem and could lead to a large increase in computation time. However, we show that, even with many obstacles, our pruning, preprocessing, and Gurobi's internal optimization strategies keep computation times below one second, even for 50 obstacles, as detailed in the supplementary material.
}

\section{Conclusion and Discussion}
\label{sec:conclusion}

\reb[4.1]{The proposed \textsc{Unique} framework introduces a temporal cascading paradigm that formulates the planning problem along the tail horizon of an MPC, replacing the conventional hierarchical planner-controller decomposition with a sequential controller-planner structure within a single optimization.} A long-horizon, low-fidelity point-mass model provides prospective planning, while a short-horizon, high-fidelity model ensures precise control, achieving real-time performance without hierarchical decomposition.
\reb[4.1]{This is enabled by the observation that the planning subproblem, primarily geometric obstacle avoidance, is a long-horizon concern that can be solved very efficiently with a point-mass model at each closed-loop iteration.} By coupling both models through feasibility-preserving transition constraints, \textsc{Unique} maintains recursive feasibility across phases. This integration avoids the redundancy and limitations of the point-mass model typical of planner-tracker pipelines and mitigates the limited long-horizon planning of single-model MPCs.

The key mechanisms of \emph{progressive obstacle smoothing} and \emph{parallel low-fidelity MPCs} enhance convergence in cluttered environments, preventing convergence to highly suboptimal local minima while preserving computational speed. Experiments show a reduction of up to 75\% in closed-loop cost compared to hierarchical and standard MPC baselines.
Both mechanisms still cannot guarantee finding the global optimal solution, as is common in NMPC. However, the experiments reveal the practical low-cost closed-loop performance. Notably, evaluating the multiple shooting cost of an RTI trajectory is nontrivial due to infeasible gaps in the dynamics. We simply use the cost of the potentially infeasible trajectory as it resembles the closed-loop cost sufficiently in our experiments. More advanced cost evaluation strategies are presented in~\cite{reiter_ac4mpc_2025}.

Another limitation is that tuning the cost of the point-mass model remains task-dependent. Because the transformation between point-mass and quadrotor states is not bijective, generic, task-independent tuning of the costs is impossible. Extending the framework to automatically tune costs for specific tasks is considered for future work.

\reb[4.4]{Notably, motion planning \restored{for even simple models} is NP-hard~\cite{lavalle2006planning}\restored{, which was also shown for \acp{MIQP} in general~\cite{delpia2017Mixed} and can even be undecidable for nonconvex \acp{MINLP}~\cite{koeppe2012Complexity,burer2012nonconvex}}. However, motion planning with linear models, formulated as \acp{MIQP} or similar programs, performs sufficiently well in practice, as shown by various algorithms~\cite{marcucci2024fast,marcucci2023motion,marcucci2024shortest,tordesillas2020faster}, highlighting the focus on planning particularly along the second horizon of the proposed algorithm.}
\reb[4.8]{For motion-planning problems of increasing complexity, such as larger mazes, the proposed approach cannot be applied without modifications. The main limiting factors are that the MPC horizon cannot be made arbitrarily long under a fixed computational budget and that numerical stability decreases as the horizon grows. Moreover, highly nonconvex planning costs such as time-optimal flight remain challenging to include in a numerically stable nonlinear program, although recent advances also aim to include more nonconvex costs~\cite{baumgaertner_2022}. A possible direction for future extensions would be to condense the long-horizon plan into a value function, potentially combined with the two stages.}

\reb[4.9]{The temporal cascading paradigm builds on two observations that hold well for quadrotors but may not transfer directly to all robotic systems: (i) that geometric planning around obstacles is predominantly a long-horizon concern with negligible influence on the short control horizon, and (ii) that meaningful state mappings from the high-fidelity to the low-fidelity model exist. For platforms where these conditions are less clear, e.g., legged robots with hybrid contact dynamics or manipulators with joint-space obstacles, the paradigm may still apply in principle, but the design of the transition function and the feasibility-preserving constraints require platform-specific analysis.
Moreover, the effectiveness of temporal cascading relies on the low-fidelity planning model converging quickly once it reaches a good local minimum during closed-loop iteration. This property, which we empirically confirm for point-mass models solved via RTI, makes the second horizon computationally cheaper than the first. If the local planning model were too expensive to solve, as would be the case for a high-fidelity model on the tail horizon, the computational advantage over simply extending the MPC horizon would vanish.
Importantly, when any parallel solver, including the MIQP, cannot finish within the main cycle, the algorithm degrades non-critically: the main multi-phase MPC retains its own feasible tail trajectory, and only completed, lower-cost solutions trigger reinitialization. This ensures that the real-time control loop is never blocked by the combinatorial planning search.}
\reb[4.9]{A central design principle is to keep the high-fidelity controller fast and delegate the computationally harder planning search to parallel asynchronous tasks. By the principle of optimality, any improvement found on the tail trajectory reduces the overall cost without affecting the first-phase controller iteration time. The Pareto analyses confirm that this decomposition is beneficial not only for planning but also for control: under the same computational budget, the two-phase structure consistently outperforms a longer, high-fidelity MPC horizon.}

\reb[4.1]{\rev{\textsc{Unique} is a real-time MPC that treats planning as a tail-horizon problem rather than a hierarchical stage, including long-horizon planning as an extension to short-horizon control in a single computationally efficient \ac{NLP} through temporal cascading of planning and control.}}

\bibliographystyle{IEEEtran}
\bibliography{library}


\begin{IEEEbiography}[{\includegraphics[width=1in,height=1.25in,clip,keepaspectratio]{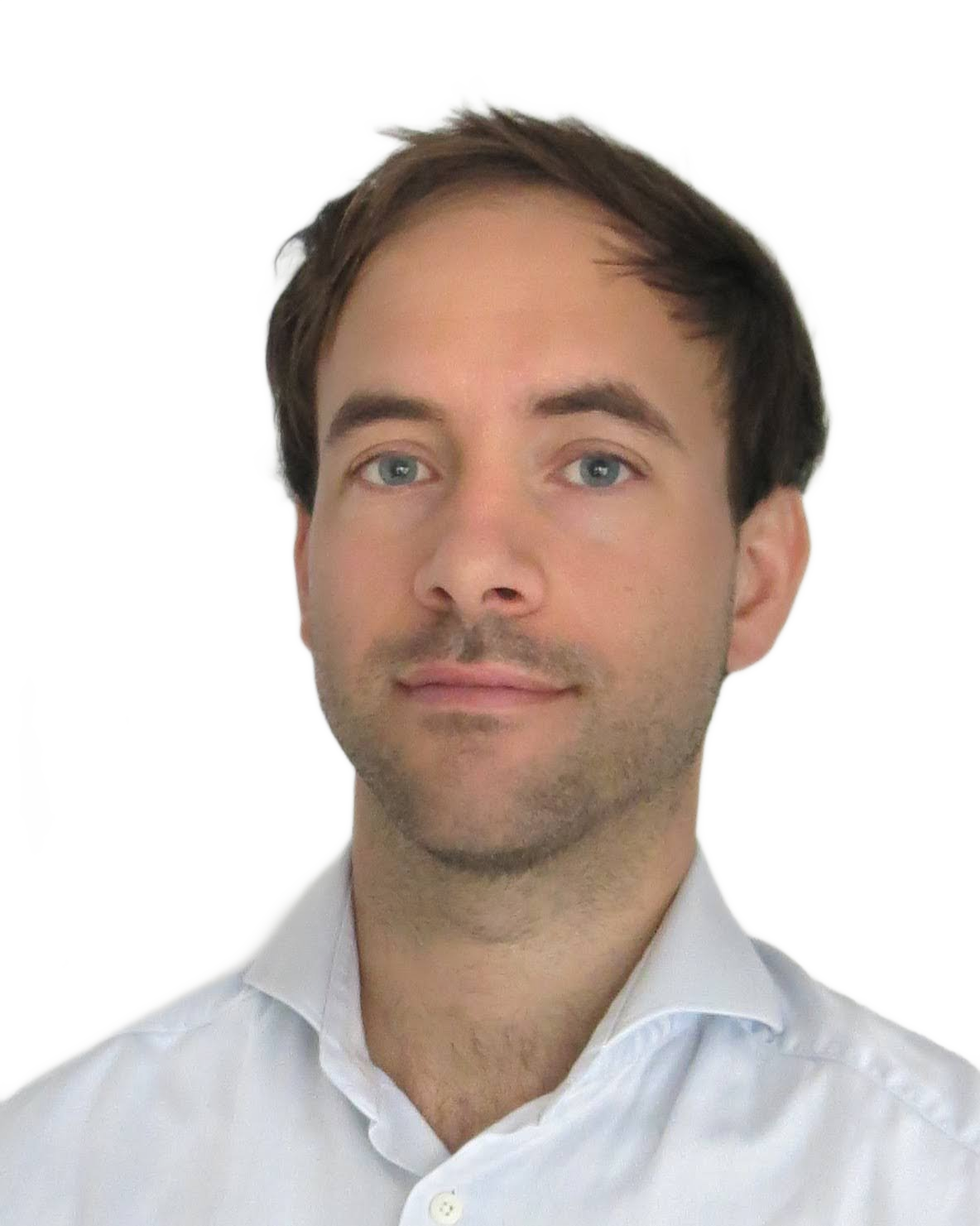}}]{Rudolf Reiter}
is a postdoctoral researcher in the Robotics and Perception Group under Prof. Davide Scaramuzza at the University of Zurich, Zurich, Switzerland. He received his Ph.D. degree from the University of Freiburg, Freiburg, Germany, under Prof. Moritz Diehl, and specializes in control systems, optimization, machine learning, and robotics. His research focuses on learning- and optimization-based motion planning and control for autonomous systems.
\end{IEEEbiography}

\begin{IEEEbiography}[{\includegraphics[width=1in,height=1.25in,clip,keepaspectratio]{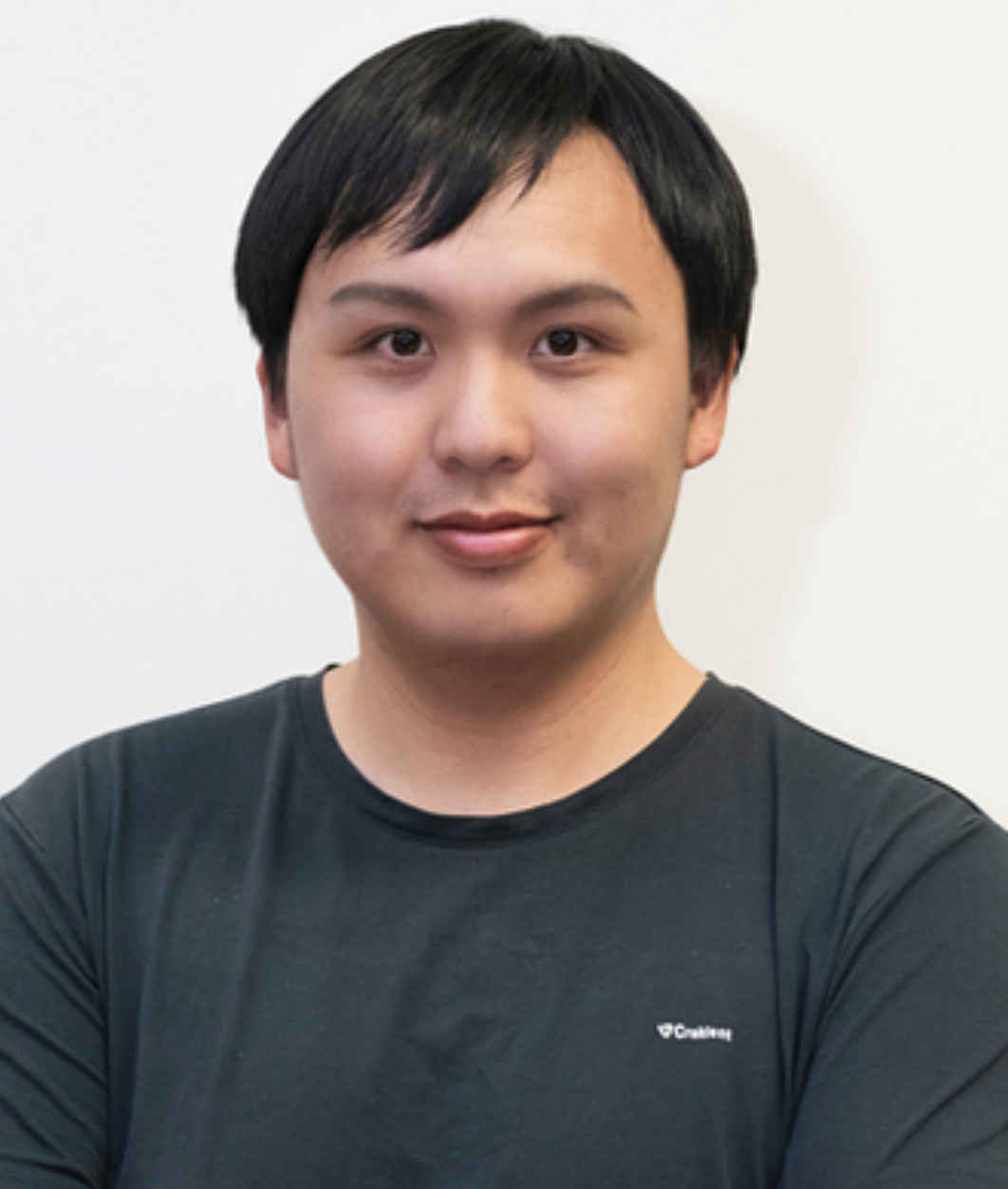}}]{Chao Qin}
received the B.Eng. degree in electrical engineering from Xidian University, Xi'an, China, in 2016, the M.S. degree in aerospace engineering from Shanghai Jiao Tong University, Shanghai, China, in 2019, and the Ph.D. degree in aerospace engineering from the University of Toronto, Toronto, ON, Canada, in 2025.
\end{IEEEbiography}

\begin{IEEEbiography}[{\includegraphics[width=1in,height=1.25in,clip,keepaspectratio]{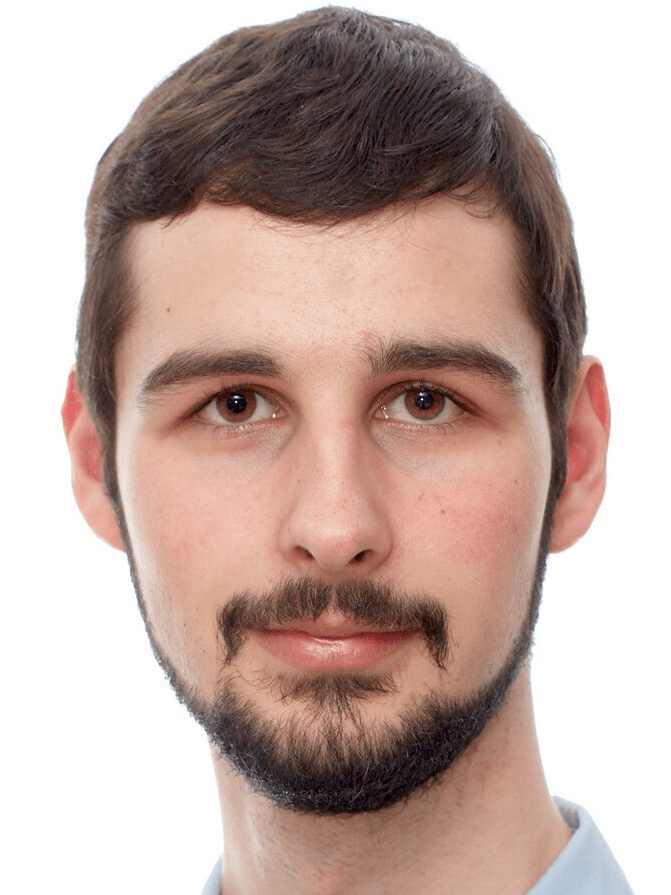}}]{Leonard Bauersfeld}
received the M.Sc. degree in robotics, systems and control from ETH Zurich, Zurich, Switzerland, in 2020. He obtained his Ph.D. degree in the Robotics and Perception Group at the University of Zurich, Zurich, Switzerland, led by Prof. Davide Scaramuzza. His research interests are autonomous vision-based quadrotor flight and quadrotor simulations. He works on combining first-principles methods with modern data-driven models to advance agile quadrotor flight.
\end{IEEEbiography}

\begin{IEEEbiography}[{\includegraphics[width=1in,height=1.25in,clip,keepaspectratio]{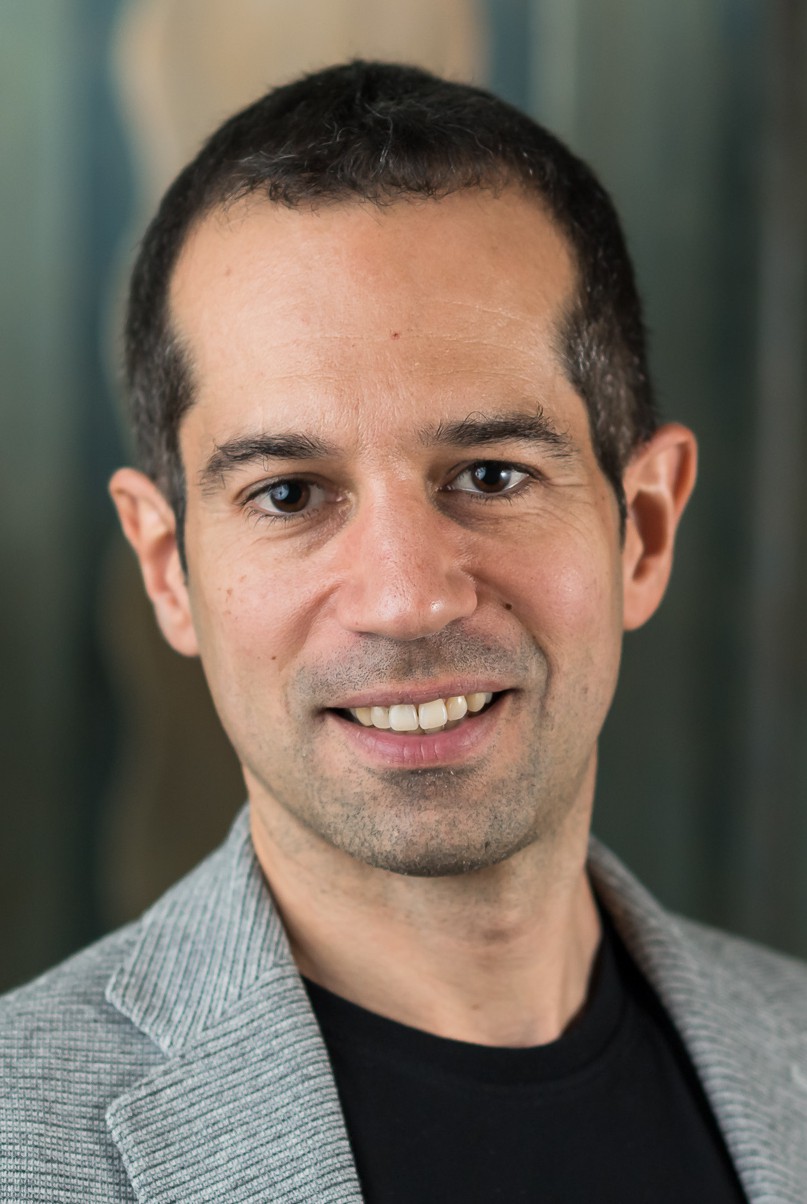}}]{Davide Scaramuzza}
is a Professor of Robotics and Perception at the University of Zurich. He did his Ph.D. at ETH Zurich, a postdoc at the University of Pennsylvania, and was a visiting professor at Stanford University and NASA Jet Propulsion Laboratory. His research focuses on autonomous, agile navigation of mobile robots using standard and event-based cameras. He made fundamental contributions to visual-inertial state estimation, autonomous vision-based agile navigation of micro flying robots, and low-latency perception with event cameras, which were transferred to many products, from drones to automobiles, cameras, AR/VR headsets, and mobile devices. He pioneered autonomous, vision-based navigation of drones, which inspired the algorithm of the NASA Mars helicopter. In 2022, his team demonstrated that an AI-powered drone could outperform the world champions of drone racing. He received several awards, including an IEEE Technical Field Award, the IEEE Fellowship, the IEEE Robotics and Automation Society Early Career Award, a European Research Council Consolidator Grant, a Google Research Award, and many paper awards. In 2015, he co-founded Zurich-Eye, today Meta Zurich, which developed the head-tracking software of the Meta Quest. In 2020, he co-founded SUIND, which builds autonomous drones for precision agriculture. Many aspects of his research have been featured in the media, such as The New York Times, The Guardian, The Economist, and Forbes. He co-authored the book "Introduction to Autonomous Mobile Robots," published by MIT Press, which has sold over 10 thousand copies worldwide and is among the most used textbooks for teaching mobile robotics. He has been consulting the United Nations on disaster response, the Fukushima Action Plan, disarmament, and AI for good.
\end{IEEEbiography}


\twocolumn[{%
\begin{center}
{\LARGE \bf Supplementary Material for:\\
Temporal Cascading of Planning and Control for Quadrotor MPC\par}
\vspace{1.2em}
{\large Rudolf Reiter, Chao Qin, Leonard Bauersfeld, and Davide Scaramuzza\par}
\vspace{2em}
\end{center}}]

This document provides supplementary derivations and experimental parameters for the main paper.

\appendix

\subsection{Residual Models for Aerodynamic Effects}
\label{sec:app_res_model}
Different types of residual forces related to their fidelity are modeled in the literature for quadrotors. We consider a polynomial higher-order model with increased accuracy but sophisticated nonlinearities for motion planning~\cite{kaufmann_2023}. 
Aerodynamic effects are considered by a residual force~$f_\text{res}(v_\bfr,\xRot^2):=\begin{bmatrix}
f_x^\text{res}&f_y^\text{res}&f_z^\text{res}
\end{bmatrix}^\T$.  The aerodynamic effects are assumed to primarily depend on the velocity~$\xVel_\bfr$ in the body frame and the average squared motor speed~$\overline{\Omega^2}=1/4\,\Omega^\top \Omega$. The forces $f_x^\text{res}$ and $f_y^\text{res}$ represent induced drag, and $f_z^\text{res}$ is induced lift. The quantities $v_x$, $v_y$, and $v_z$ denote the three axial velocity components in the body frame, and $v_{xy}$ denotes the velocity in the $(x,y)$ plane of the quadrotor. Based on insights from the underlying physical processes, linear and quadratic combinations of the individual terms are selected. The forces are modeled by
\begin{align*}
f_x^\text{res}&(v_\bfr,\xRot) := m  \begin{bmatrix} 1&
v_x & v_x|v_x| & v_x \,\overline{\Omega^2}
\end{bmatrix} c_x,  \\
f_y^\text{res}&(v_\bfr,\xRot) :=m \begin{bmatrix}1 & v_y & v_y|v_y| & v_y \,\overline{\Omega^2}\end{bmatrix}  c_y, \\ 
f_z^\text{res}&(v_\bfr,\xRot) :=m c_z^\top\\
&\begin{bmatrix} 1&v_z & v_z^3 & v_{xy} & v_{xy}^2 & v_{xy}\,\overline{\Omega^2} & v_{xy}\,v_z\,\overline{\Omega^2} & \overline{\Omega^2} \end{bmatrix}^\top,
\end{align*}
where the weights~$\rev{c_x\in\R^4,\,c_y\in\R^4}$ and $c_z\in\R^8$ are obtained from fitting the model to real-world data.
\begin{table}[h]
	\centering
	\caption{Residual Parameters}
	\label{tab:res_params}
	\renewcommand{\arraystretch}{1.2}
	\setlength{\tabcolsep}{4pt}
	\begin{tabularx}{\linewidth}{lX}
		\hline
		\textbf{Parameter} & \textbf{Value / Description} \\ \hline
		
		\multicolumn{2}{l}{\textit{High-Fidelity Quadrotor Model}} \\ 
		$c_x$ & [1.18e-02, -1.39e-01, -1.59e-03, -8.31e-08]\\ 
		$c_y$ & [-3.21e-02, -9.79e-02, -6.85e-03, -1.01e-07]\\ 
		$c_z$ & [-5.00e-01, 1.16e-01, -4.25e-01, -1.04e-06, 1.76e-07, -1.89e-08, -1.03e-03, \rev{0.00e+00}]\\
		\hline
	\end{tabularx}
\end{table}

\subsection{\rev{Differential flatness} for quadrotors}
\label{sec:app_diff_flat}
\rev{Differential flatness} refers to systems where there exists an output~$y=h(x,u,\dot{u},\ddot{u},\ldots,u^{(r)})$, such that all states~$x$ and controls~$u$ can be described by \rev{a function} of derivatives of~$y$, i.e.,
\begin{align}
	x&=\gamma_1(y,\dot{y},\ldots,y^{(k)}),\\
	u&=\rev{\gamma_2}(y,\dot{y},\ldots,y^{(k)}),
\end{align}
for some functions~$\gamma_1, \gamma_2$ and constants~$r,k\in\Z$.
As shown in~\cite{faessler_2018}, quadrotors with linear disturbance models are \rev{differentially flat} for an output~$[p^\top, \psi]^\top$, where $\psi\in\R$ is the heading angle.
The property of \rev{differential flatness} implies that trajectory planning can be done directly in the space of flat outputs. This means that
trajectories can be designed for the flat outputs, and the full system trajectory recovered.

\subsection{Quadrotor jerk}
\label{sec:app_jerk}
In the following, we derive the jerk of the quadrotor similar to~\cite{faessler_2018}.
\rev{Define the body-frame velocity $v_\bfr:=q^{-1}\odot v$ and let $C\in\R^{3\times3}$ denote the linear body-frame residual-force matrix. Then}
\begin{equation*}
\rev{ma=q\odot\left(T+Cv_\bfr\right)-mg.}
\end{equation*}
\rev{Using $\frac{d}{dt}(q\odot z)=q\odot(\dot z+\omega\times z)$ gives}
\begin{align*}
\rev{\dot v_\bfr}&\rev{=q^{-1}\odot\dot v-\omega\times v_\bfr,}\\
\rev{\frac{d}{dt}(q\odot T)}&\rev{=q\odot(\dot f_\mathrm{th}+\omega\times T),}\\
\rev{\frac{d}{dt}(q\odot Cv_\bfr)}&\rev{=q\odot(C\dot v_\bfr+\omega\times Cv_\bfr).}
\end{align*}
\rev{Consequently, with $\dot v$ as defined in the main paper, Eq.~(1), the jerk is}
\begin{align*}
\rev{j_\mathrm{quad}}&\rev{(q,v,\omega,T,\dot f_\mathrm{th})=\frac{1}{m}q\odot\Big(\dot f_\mathrm{th}+\omega\times T}\\[-1mm]
&\rev{\qquad+C(q^{-1}\odot\dot v-\omega\times v_\bfr)+\omega\times(Cv_\bfr)\Big).}
\end{align*}

\noindent\rev{\textit{Recursive feasibility.} \textbf{Proposition 1.} Under initial feasibility, shift-consistent hard constraints, feasibility-preserving transition and constraint alignment, and an invariant hover terminal set, any feasible parallel tail sharing the transition state preserves recursive feasibility. \emph{Proof.} Concatenating the unchanged high-fidelity horizon with such a tail is feasible because its initial state, dynamics, constraints, and terminal condition agree. After applying the first control, shift the trajectory by one control period; alignment provides a feasible high-fidelity continuation at the phase boundary, and appending the invariant hover state completes a feasible candidate. Induction proves the claim.\hfill$\square$}

\subsection{Parameters for Constant Velocity Experiments}
\label{sec:app_param_const_vel}
Table~\ref{tab:const_vel_params} shows the parameters used in the Constant Velocity experiments. We used equal parameters for~\textsc{Unique}, the standard MPC and the hierarchical setting, if not stated differently in the main paper for the particular hyperparameter comparison. These experiments do not use the parallel point-mass MPCs.
\begin{table}[t]
	\centering
	\caption{Parameters for Constant Velocity Experiments (SI units)}
	\label{tab:const_vel_params}
	\renewcommand{\arraystretch}{1.2}
	\setlength{\tabcolsep}{4pt}
	\begin{tabularx}{\linewidth}{lX}
		\hline
		\textbf{Parameter} & \textbf{Value / Description} \\ \hline

		\multicolumn{2}{l}{\textit{High-Fidelity Quadrotor Model}} \\
		$w$ &
		[0, 0.01, 1.0, 30, 30, 30, 10, 0, 0, 10, 10, 10, 3, 3, 3, 3, 3e$^{-5}$, 3e$^{-5}$, 3e$^{-5}$, 3e$^{-5}$] \\
		$\tilde{y}$ &
		[0, 0, 0, 0, 0, 0, 15, 0, 0, 0, 0, 0, 1.47, 1.47, 1.47, 1.47, 0, 0, 0, 0] \\

		\multicolumn{2}{l}{\textit{Low-Fidelity Point-Mass Model}} \\
		$w^\mathrm{lf}$ &
		[0.0, 0.01, 1, 10, 0.0001, 0.0001, 1.08, 1.08, 1.08, 0.1, 0.1, 0.1] \\
		$\tilde{y}^\mathrm{lf}$ &
		[0, 0, 0, 0, 0, 0, 0, 0, 0, 0, 0, 0] \\[2mm]

		\multicolumn{2}{l}{\textit{Point Mass Constraint Parameters}} \\
		$\lb{a}_z$ & $-5.0$ (requirement: $\rev{\lb{a}_z>(\lb{f}_\mathrm{th}+\ub{f}_\mathrm{res})/m-g}$) \\
		$\ub{f}_\mathrm{res}$ & $2.0$ \\
		$\alpha_x$ & $0.5$ \\
		$\alpha_z$ & $0.5$ \\[2mm]
		
		\multicolumn{2}{l}{\textit{Planner Parameters}} \\
		MIQP active & no \\[2mm]

		\multicolumn{2}{l}{\textit{Tracking MPC Weights (different from above)}} \\
		$w[0:2]$ &
		[10, 10, 10]\\[2mm]

		\multicolumn{2}{l}{\textit{Environment Settings}} \\
		Ellipsoid dims. &
		[34, 34, 34], [25, 25, 25], [55, 55, 55], [52, 52, 52] \\
		Ellipsoid pos. &
		[155, 20, 0], [50, -20, 0], [255, 20, 0], [450, -20, 0] \\
		Max speeds & 2 \\
		Movement type & Brownian motion \\[2mm]

		\multicolumn{2}{l}{\textit{Simulation Settings}} \\
		Drone start & [0, 0, 0] \\
		Simulation steps & 10000 \\
		Time step & 0.04 \\[2mm]

		\multicolumn{2}{l}{\textit{Evaluation}} \\
		$Q_{\text{eval}}, R_{\text{eval}}$ & $\mathrm{diag}(Q_{\text{eval}}, R_{\text{eval}})=diag(w)$ (same as MPC costs) \\

		\hline
	\end{tabularx}
\end{table}

\subsection{Parameters for Tracking experiments}
\label{sec:app_param_tracking}
Table~\ref{tab:tracking_params} shows the parameters used in the Tracking experiments. We used equal parameters for~\textsc{Unique}, the standard MPC and the hierarchical setting. These experiments use one parallel point-mass MPC in the Cluttered-Figure-Eight environment. In the cubic obstacle environment, the equal obstacle positions and dimensions are used but with the maximum volume inscribed cube of the ellipsoid. Parameters not stated are equal to Tab.~\ref{tab:const_vel_params}
\begin{table}[t]
	\centering
	\caption{Parameters for Trajectory Tracking Experiments (SI units)}
	\label{tab:tracking_params}
	\renewcommand{\arraystretch}{1.2}
	\setlength{\tabcolsep}{4pt}
	\begin{tabularx}{\linewidth}{lX}
		\hline
		\textbf{Parameter} & \textbf{Value / Description} \\ \hline

		\multicolumn{2}{l}{\textit{High-Fidelity Quadrotor Model}} \\
		$w$ &
		[500, 500, 500, 10, 10, 10, 0, 0, 0, 10, 10, 10, 3, 3, 3, 3, 3e$^{-5}$, 3e$^{-5}$, 3e$^{-5}$, 3e$^{-5}$] \\
		$\tilde{y}$ &
		[0, 0, 0, 0, 0, 0, 0, 0, 0, 0, 0, 0, 1.47, 1.47, 1.47, 1.47, 0, 0, 0, 0] \\

		\multicolumn{2}{l}{\textit{Low-Fidelity Point-Mass Model}} \\
		$w^\mathrm{lf}$ &
		[500, 500, 500, 0, 0, 0, 0.05, 0.05, 0.05, 0.1, 0.1, 0.1] \\
		$\tilde{y}^\mathrm{lf}$ &
		[0, 0, 0, 0, 0, 0, 0, 0, 0, 0, 0, 0] \\[2mm]
		
		\multicolumn{2}{l}{\textit{Planner Parameters}} \\
		MIQP active & no \\[2mm]

		\multicolumn{2}{l}{\textit{Environment Agile Sinusoidal}} \\
		Ellipsoid radii &		.5, .2, .5, .8, .5, .8, .9 \\
		Ellipsoid pos. &		Equal spacing \\
		\textsc{Unique} Hor. $N,M/t_{\Delta},t_{\Delta}^\mathrm{lf}$ & 23, 10 / 0.02, 0.2\\
		Hier. Appr. Hor. $N,M/t_{\Delta},t_{\Delta}^\mathrm{lf}$ & 23, 12 / 0.02, 0.2\\
		Std. MPC Hor. $N/t_{\Delta}$ & 30 / 0.02\\[2mm]

		\multicolumn{2}{l}{\textit{Environment Agile Butterfly}} \\
		Ellipsoid radii &		.5, .2, .5, .8, .5, .8, .9 \\
		Ellipsoid pos. &		Equal spacing \\
		\textsc{Unique} Hor. $N,M/t_{\Delta},t_{\Delta}^\mathrm{lf}$ & 23, 5 / 0.02, 0.1\\
		Hier. Appr. Hor. $N,M/t_{\Delta},t_{\Delta}^\mathrm{lf}$ & 23, 12 / 0.02, 0.2\\
		Std. MPC Hor. $N/t_{\Delta}$ & 35 / 0.02\\[2mm]

		\multicolumn{2}{l}{\textit{Environment Cluttered Figure Eight-1}} \\
		Ellipsoid radii &		1.5, 0.9, 1.5, 0.8, 0.9, 1.5, 0.9, 1.5, 0.9, 0.9, 1.5, 1.5, 1.5, 0.9, 1.5, 0.9, 1.5, 0.9, 1.5, 1.5, 1.5, 0.9, 0.9, 0.9, 1.5, 1.5, 1.5, 0.9, 0.9, 0.9, 1.5 \\
		Ellipsoid pos. &		Equal spacing, random displacement by \unit[1]{m} in each axis \\
		\textsc{Unique} Hor. $N,M/t_{\Delta},t_{\Delta}^\mathrm{lf}$ & 30, 30 / 0.02, 0.15\\
		Hier. Appr. Hor. $N,M/t_{\Delta},t_{\Delta}^\mathrm{lf}$ & 30, 40 / 0.02, 0.15\\
		Std. MPC Hor. $N/t_{\Delta}$ & 50 / 0.02\\[2mm]

		\multicolumn{2}{l}{\textit{Environment Cluttered Figure Eight-2}} \\
		Ellipsoid radii &		1.5, 0.9, 1.5, 0.8, 0.9, 1.5, 0.9, 1.5, 0.9, 0.9, 1.5, 1.5, 1.5, 0.9, 1.5, 0.9, 1.5, 0.9, 1.5, 1.5, 1.5, 0.9, 0.9, 0.9, 1.5, 1.5, 1.5, 0.9, 0.9, 0.9, 1.5, 0.9, 0.9, 0.9 \\
		Ellipsoid pos. &		Equal spacing, random displacement by \unit[1]{m} in each axis \\
		\textsc{Unique} Hor. $N,M/t_{\Delta},t_{\Delta}^\mathrm{lf}$ & 30, 30 / 0.02, 0.3\\
		Hier. Appr. Hor. $N,M/t_{\Delta},t_{\Delta}^\mathrm{lf}$ & 30, 30 / 0.02, 0.3\\
		Std. MPC Hor. $N/t_{\Delta}$ & 50 / 0.02\\[2mm]

		\multicolumn{2}{l}{\textit{Simulation Settings}} \\
		Simulation steps & various (completing one lap) \\
		Time step & 0.02 \\[2mm]

		\multicolumn{2}{l}{\textit{Evaluation}} \\
		$Q_{\text{eval}}, R_{\text{eval}}$ & $\mathrm{diag}([1,1,1,\boldsymbol{0}])$ (positions only) \\

		\hline
	\end{tabularx}
\end{table}

\subsection{Parameters for Maze Experiments}
\label{sec:app_param_const_maze}
Table~\ref{tab:maze_params} shows the parameters used in the \rev{maze experiments}. We used equal parameters for~\textsc{Unique}, the standard MPC and the hierarchical setting, if not stated differently in the main paper for the particular hyperparameter comparison. \rev{These experiments use the parallel point-mass MPCs and, where indicated, the MIQP initializer.}
\begin{table}[t]
	\centering
	\caption{Parameters for Maze Experiments (SI units)}
	\label{tab:maze_params}
	\renewcommand{\arraystretch}{1.2}
	\setlength{\tabcolsep}{4pt}
	\begin{tabularx}{\linewidth}{lX}
		\hline
		\textbf{Parameter} & \textbf{Value / Description} \\ \hline
		
		\multicolumn{2}{l}{\textit{High-Fidelity Quadrotor Model}} \\
		$w$ &
		[0, 0.001, 10.0, 3, 3, 3, 1, 0, 1, 10, 10, 10, 3, 3, 3, 3, 3e$^{-5}$, 3e$^{-5}$, 3e$^{-5}$, 3e$^{-5}$] \\
		$\tilde{y}$ &
		[0, 0, 1, 0, 0, 0, 1, 0, 0, 0, 0, 0, \rev{1.47}, 1.47, 1.47, 1.47, 0, 0, 0, 0] \\
		
		\multicolumn{2}{l}{\textit{Low-Fidelity Point-Mass Model}} \\
		$w^\mathrm{lf}$ &
		[0, 0.001, 10.0, 1, 0, 1, 0.01, 0.01, 0.01, 0.1, 0.1, 0.1] \\
		$\tilde{y}^\mathrm{lf}$ &
		[0, 0, 1, 1, 0, 0, 0, 0, 0, 0, 0, 0] \\[2mm]
		
		\multicolumn{2}{l}{\textit{Point Mass Constraint Parameters}} \\
		$\lb{a}_z$ & $-5.0$ (requirement: $\rev{\lb{a}_z>(\lb{f}_\mathrm{th}+\ub{f}_\mathrm{res})/m-g}$) \\
		$\ub{f}_\mathrm{res}$ & $2.0$ \\
		$\alpha_x$ & $0.5$ \\
		$\alpha_z$ & $0.5$ \\[2mm]
		
		\multicolumn{2}{l}{\textit{Planner Parameters}} \\
		MIQP active & yes \\
		Smoothing $\alpha$ & 20 (constant) \\
		Number $S$ par. PM & 3\\
		Reini. steps $P_\mathrm{PM}$ & 20\\
		Reini. steps $P_\mathrm{MIQP}$ & 40\\[2mm]

		\multicolumn{2}{l}{\textit{Tracking MPC Weights (different from above)}} \\
		$w[0:2]$ &
		[30, 30, 30]\\[2mm]
		
		\multicolumn{2}{l}{\textit{Environment Settings}} \\
		obst. width &  $2-2.5$m\\
		obst. length &  $4-6$m\\
		obst. height &  $10$m\\
		orientation&  $\{0,90\}$ degree around z-axis\\
		$N$ obstacles & 6 to 14 \\[2mm]
		
		\multicolumn{2}{l}{\textit{Simulation Settings}} \\
		Drone start & [-9, 0, 1] \\
		Simulation steps & until finished ($x\geq12$m)\\
		Time step & 0.04 \\[2mm]
		
		\multicolumn{2}{l}{\textit{Evaluation}} \\
		$Q_{\text{eval}}, R_{\text{eval}}$ & $\mathrm{diag}(Q_{\text{eval}}, R_{\text{eval}})=diag(w)$ (same as MPC costs) \\
		\hline
	\end{tabularx}
\end{table}

\subsection{Obstacle avoidance with progressive smoothing}
\label{sec:app_progressive_smoothing}
\begin{figure}
	\centering
	\includegraphics[width=\linewidth, trim=0.45cm 0.2cm 0.4cm 0.2cm, clip]{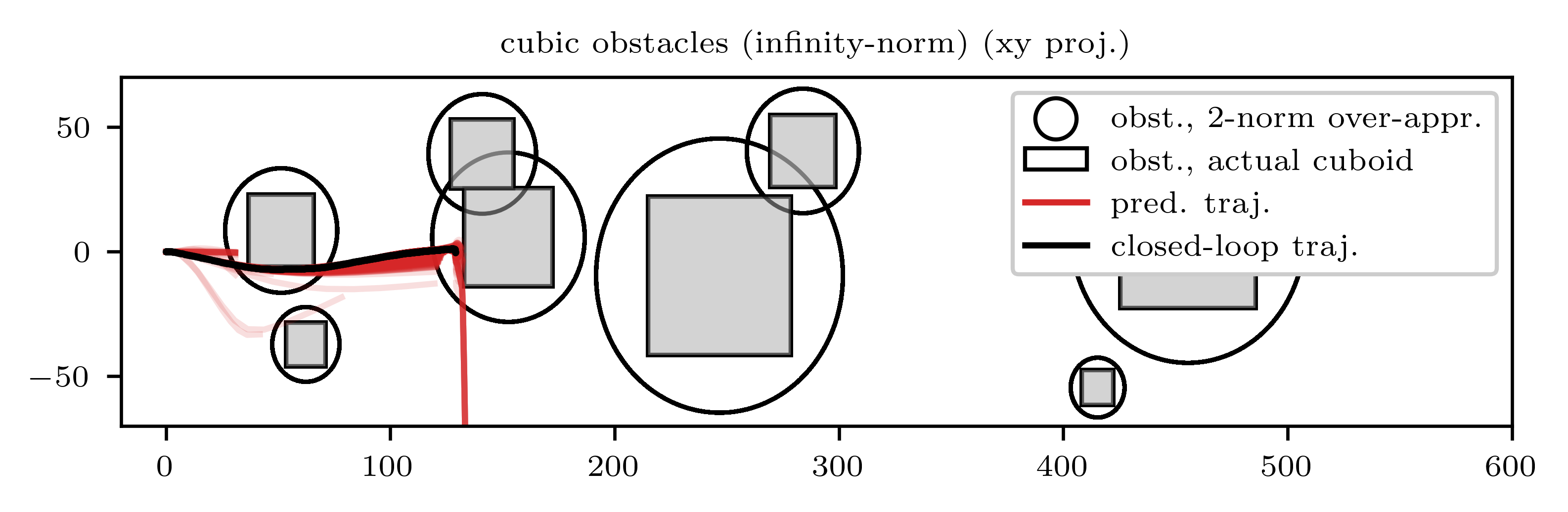}
	\includegraphics[width=\linewidth, trim=0.45cm 0.2cm 0.4cm 0.2cm, clip]{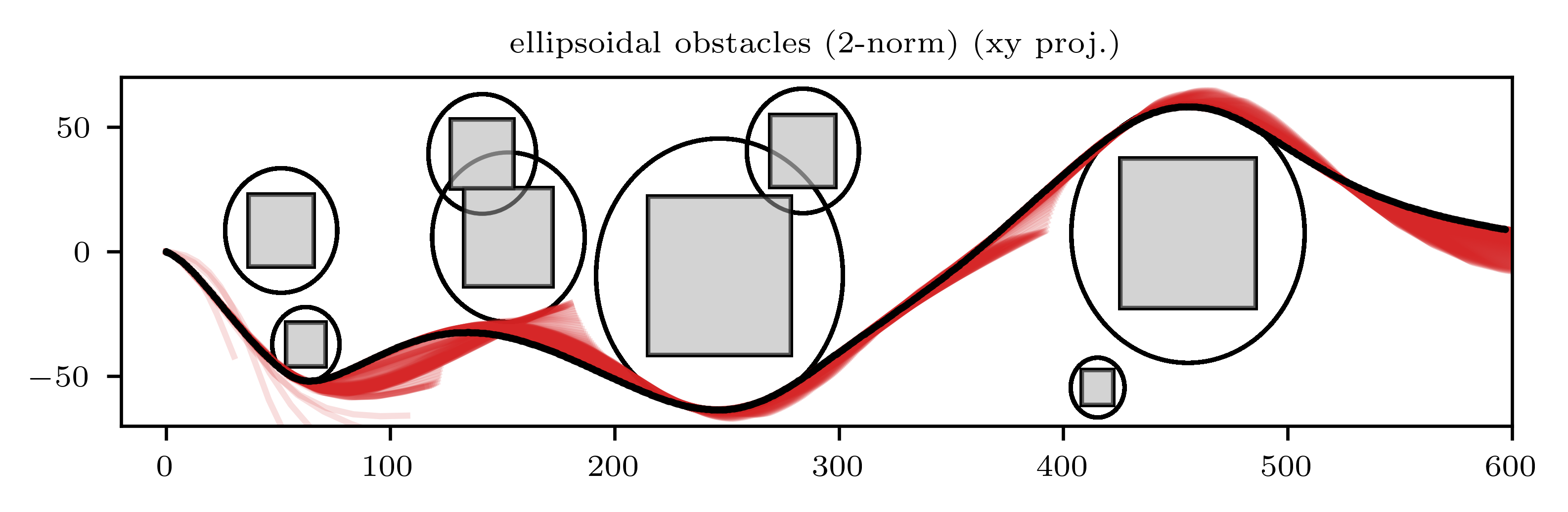}
	\includegraphics[width=\linewidth, trim=0.45cm 0.2cm 0.4cm 0.2cm, clip]{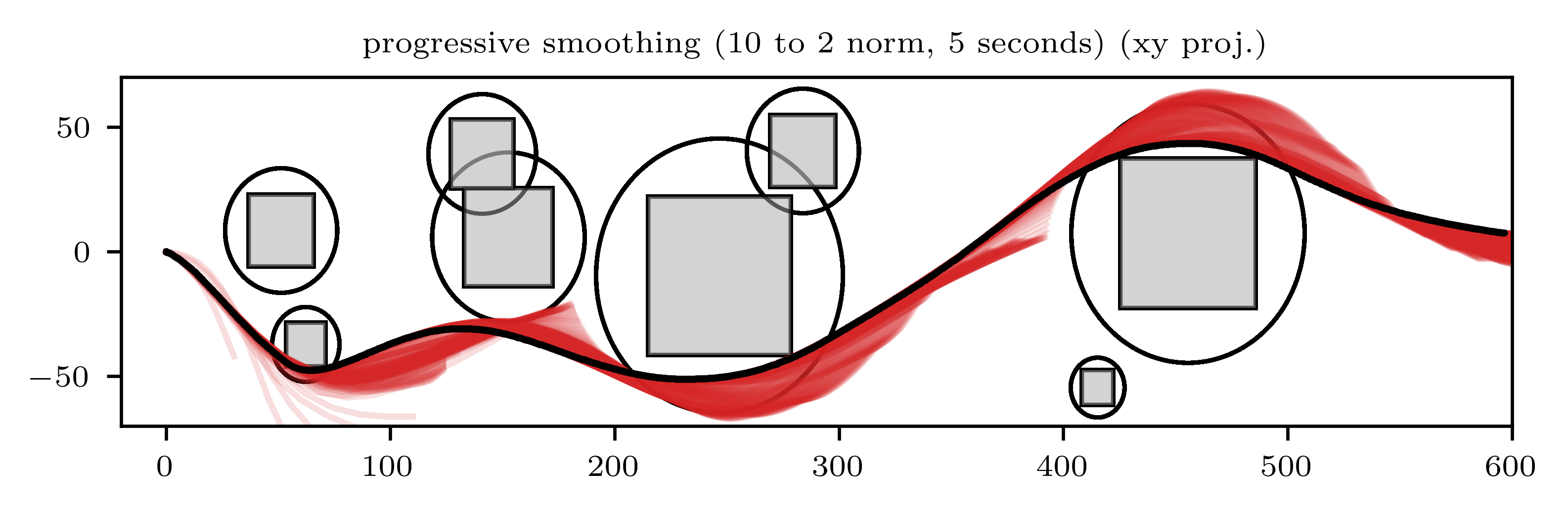}
	\includegraphics[width=\linewidth, trim=0.45cm 0.2cm 0.4cm 0.2cm, clip]{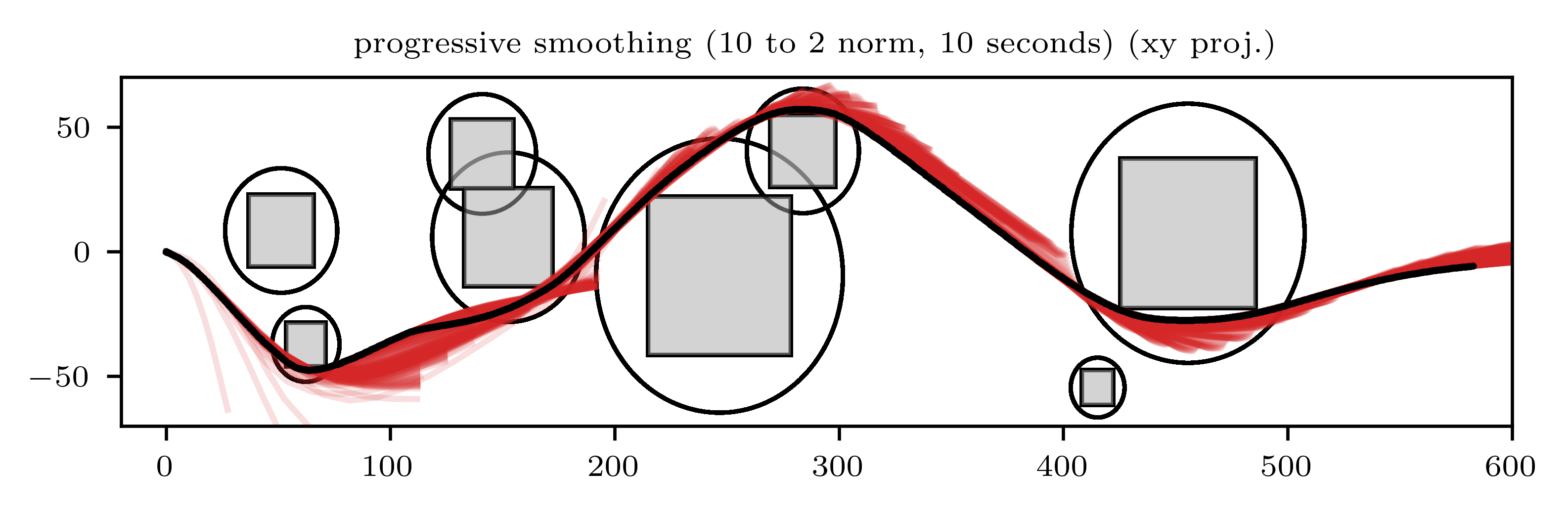}
	\caption{Comparison of the influence of progressive smoothing in simulated experiments with cubic obstacles. In the first plot, the cubic obstacles are formulated via infinity norms, which leads to nonsmooth shapes and the drone getting stuck in local minima. The second plot shows conservative overapproximations with spheres, where the drone successfully navigates around obstacles, yet at a large distance. The final two plots show different parameterizations of progressive smoothing, which allows tight navigation without getting stuck in local minima.}
	\label{fig:obstacles}
\end{figure}

To give more qualitative insights into progressive smoothing, we plot the predicted MPC trajectories in the constant velocity environment with static obstacles in Fig.~\ref{fig:obstacles}. We compare the predictions with 2-norm, infinity norm, and two progressive smoothing schedules, that linearly vary the norm from~$10$ to~$2$ in either~$5$ or $\unit[10]{\mathrm{s}}$. The evaluation reveals a slightly different behavior when using progressive smoothing. However, the closed-loop trajectories are similar and \rev{traverse tightly} along the cubic obstacle shapes.

\subsection{Scaling of the mixed-integer planner}
\label{sec:app_miqp_scaling}
\begin{figure}
	\centering
	\includegraphics[width=\linewidth, trim=0cm 0cm 0cm 0cm, clip]{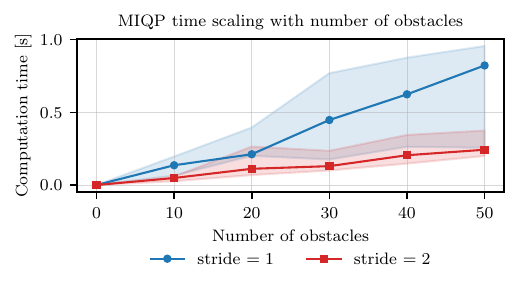}
	\caption{Time scaling of the mixed-integer planner with an increased number of obstacles, using a stride of one or two (skipped nodes in the obstacle-avoidance formulation).}
	\label{fig:time_scaling}
\end{figure}

The mixed-integer planner uses binary indicator constraints for obstacle-free convex cells, whose number grows with the number of obstacles, which in principle scales an NP-hard problem. Nevertheless, as shown in Fig.~\ref{fig:time_scaling}, our pruning, preprocessing, and Gurobi's internal optimization strategies keep the computation time below one second even for 50 obstacles.

\end{document}